\documentclass[10pt]{article}
\usepackage[preprint]{tmlr}

\usepackage{amsmath,amsfonts,bm}

\def\eqref#1{equation~\ref{#1}}

\def\1{\bm{1}}

\newcommand{\valid}{\mathcal{D_{\mathrm{valid}}}}

\DeclareMathAlphabet{\mathsfit}{\encodingdefault}{\sfdefault}{m}{sl}
\SetMathAlphabet{\mathsfit}{bold}{\encodingdefault}{\sfdefault}{bx}{n}

\usepackage{amsmath}
\usepackage{amssymb}
\usepackage{booktabs}
\usepackage{tabularx}
\usepackage{longtable}
\usepackage{float}
\usepackage{placeins}
\usepackage{array}
\usepackage{enumitem}
\usepackage{xcolor}
\usepackage{graphicx}
\usepackage{tikz}
\usepackage[skins,breakable]{tcolorbox}
\usepackage{hyperref}
\usepackage{url}
\hypersetup{
  colorlinks=true,
  linkcolor=blue,
  citecolor=blue,
  urlcolor=blue,
  filecolor=blue
}
\usetikzlibrary{arrows.meta,positioning,fit,shapes.geometric,calc,matrix,decorations.pathreplacing,backgrounds}

\newcolumntype{Y}{>{\raggedright\arraybackslash}X}
\newcolumntype{L}[1]{>{\raggedright\arraybackslash}p{#1}}
\newcolumntype{B}[1]{>{\raggedright\arraybackslash\bfseries}p{#1}}
\newcommand{\SurveyTableSetup}{\footnotesize\setlength{\tabcolsep}{4pt}\renewcommand{\arraystretch}{1.08}}
\newcommand{\SurveyLongTableSetup}{\small\setlength{\tabcolsep}{4pt}\setlength{\LTcapwidth}{\linewidth}\renewcommand{\arraystretch}{1.08}}
\setlist[itemize]{leftmargin=*, itemsep=2pt, topsep=2pt}
\setlist[enumerate]{leftmargin=*, itemsep=2pt, topsep=2pt}
\definecolor{nsblue}{RGB}{38,84,140}
\definecolor{nsgreen}{RGB}{36,117,74}
\definecolor{nsorange}{RGB}{179,102,34}
\definecolor{nsred}{RGB}{155,54,54}
\definecolor{nsgray}{RGB}{88,94,104}
\definecolor{nslight}{RGB}{244,247,250}
\newtcolorbox{takeawayquote}{
  enhanced,
  breakable,
  colback=nslight,
  colframe=nslight,
  borderline west={2.4pt}{0pt}{nsblue},
  boxrule=0pt,
  arc=0pt,
  left=8pt,
  right=8pt,
  top=7pt,
  bottom=7pt,
  boxsep=0pt,
  before skip=9pt,
  after skip=9pt,
  width=0.96\linewidth,
  center,
  fontupper=\small
}
\title{In-Context Reinforcement Learning under Non-Stationarity: A Survey}

\author{\name A Run \\
\addr Independent Researcher
\AND
\name Ziluo Ding\thanks{Corresponding author: \href{mailto:ziluo@pku.edu.cn}{ziluo@pku.edu.cn}} \\
\addr Beijing Innovation Center of Humanoid Robotics}

\def\month{MM}
\def\year{YYYY}
\def\openreview{}

\newcommand{\ICRL}{In-context reinforcement learning}

\newcommand{\ctx}{C}
\newcommand{\env}{\mathcal{M}}
\newcommand{\policy}{\pi_\theta}

\begin{document}

\maketitle

\begin{abstract}
The development of decision-pretrained transformers, algorithm distillation, long-context meta-RL, and retrieval-augmented agents has renewed interest in in-context reinforcement learning (ICRL): the ability of a pretrained or fine-tuned decision model to infer latent task rules and improve future behavior from interaction context, without test-time parameter updates.
This line of work asks when trial-and-error evidence, rewards, transitions, demonstrations, feedback, or retrieved experience can make learning-like computation happen inside the context window.
However, existing surveys of ICRL mainly organize the field around pretraining objectives, architectures, context formats, evaluation protocols, and theoretical mechanisms, while the non-stationary setting remains comparatively underexamined.
In changing environments, accumulated context is not merely more evidence about a fixed task: the reward specification, transition kernel, observation channel, action interface, constraint model, or demonstration and memory distribution can fall out of alignment with the current regime.
Previously useful context can therefore become stale, misleading, or useful again when an old regime returns.
We survey non-stationary ICRL as the problem of adapting through context while deployed policy parameters remain fixed: the policy must infer both the current decision rule and which parts of its accumulated evidence still support that rule.
We define non-stationary ICRL, relate it to meta-RL, decision sequence modeling, retrieval-augmented RL, value- and model-aware ICRL, and reward-feedback agents, and organize the literature along three questions: what changes, how the change unfolds, and how observable the change is to the agent.
We further analyze methods through context-management operations and influence facets, including what is written, retrieved, compressed, trusted, forgotten, or isolated, and argue that average return and held-out task generalization are insufficient to establish adaptation under drift.
The survey identifies a research agenda around lifecycle evaluation, stale-context stress tests, validity-aware retrieval, adaptive forgetting, context poisoning, and finite-context theory for non-stationary ICRL.
\end{abstract}

\section{Introduction}
\label{sec:introduction}

\ICRL{} (ICRL) asks whether a pretrained or fine-tuned sequential decision model can infer task rules and improve future actions from the interaction history available in its context, without test-time parameter updates~\citep{duan2016rl2}.
The context may contain states, actions, rewards, transition outcomes, demonstrations, feedback, retrieved memories, or summaries, but the defining property is learning-like adaptation through that context rather than through an online optimizer~\citep{laskin2023algorithm,zisman2024noise}.
That idea runs through black-box meta-reinforcement learning \citep{duan2016rl2,wang2016learning}, gradient-based meta-learning \citep{finn2017maml}, offline trajectory modeling \citep{janner2021trajectory}, decision transformers \citep{chen2021decision}, algorithm distillation \citep{laskin2023algorithm}, supervised ICRL \citep{lee2023supervised}, and long-context meta-RL \citep{melo2022transformersmetarl,grigsby2023amago}.
It also inherits a broader question from in-context learning: when does a sequence model merely condition on examples \citep{brown2020language,min2022rethinking}, and when does its forward pass implement something close to inference \citep{chan2022datadistribution,garg2022transformers,akyurek2023whatlearning}, optimization \citep{vonoswald2023gradient,kirsch2022gpicl}, or learning \citep{wies2023learnability,akyurek2024contextlanguage}?
This capability is especially appealing in reinforcement learning.
If useful adaptation can happen through the input stream, then a system can respond quickly without running an optimizer, changing deployed weights, or hiding the adaptation process inside an online update.

That promise is much easier to state in stationary settings than in changing ones.
Recent survey work already organizes ICRL broadly around pretraining, context construction, architectures, evaluation, and theory \citep{moeini2025surveyicrl}.
Here we focus on the case where the environment changes during the lifetime over which context is accumulated.
By non-stationarity, we mean temporal changes in the decision process that alter which accumulated evidence remains valid for action.
In a stationary few-shot task, more context is usually more evidence about one latent task.
Under non-stationarity, the same extra trajectory can be useful evidence, obsolete evidence, a misleading demonstration, or a memory that becomes useful again only after a previous regime returns.
The question is no longer simply whether the agent uses context.
It is whether the agent can tell which parts of its context still describe the world it is acting in.

For ICRL, non-stationarity is best viewed as a change in the validity conditions of context: past trajectories, rewards, feedback, demonstrations, or retrieved memories may no longer license the same policy choice.
The underlying source can be a changed reward specification~\citep{besbes2014nonstationarybandit,song2026reward}, transition kernel~\citep{lecarpentier2019nonstationary,chandak2020optimizing}, observation channel~\citep{cobbe2020procgen,wang2025popgymarcade}, action interface~\citep{chandak2020changingaction,sinii2024variable}, preference or safety model~\citep{moeini2025safeicrl,song2026reward}, task distribution~\citep{alshedivat2017continuous,wang2025anymdp}, or behavior policy behind the data~\citep{wang2026suboptimalicrl,schmied2024radt}.
These sources impose different adaptation burdens.
Reward switches require reinterpretation of feedback, transition drift changes action--outcome evidence, action remapping changes the meaning of behavior, and behavior-policy shift changes the trustworthiness of demonstrations or retrieved trajectories.
An abrupt reward switch creates a different adaptation problem~\citep{adams2007changepoint,garivier2011ucb} from gradual transition drift~\citep{lecarpentier2019nonstationary,feng2023nonstationary}; a recurring latent mode creates a different memory problem~\citep{hallak2015contextualmdp,killian2017hipmdp} from one-way degradation~\citep{khetarpal2022continual,abel2023definition}; an observed task descriptor creates a different inference problem~\citep{hallak2015contextualmdp,humplik2019taskinference} from a hidden change point~\citep{adams2007changepoint,keplinger2025nsgym}.
For ICRL, the relevant consequence is that context has a lifetime.
It can be useful when it identifies the current mode, harmful when it anchors the agent to a past mode, and useful again when an old mode recurs.
This framing draws on older Markov decision process \citep{puterman1994markov}, partially observable decision process \citep{kaelbling1998pomdp}, contextual MDP \citep{hallak2015contextualmdp}, hidden-parameter MDP \citep{killian2017hipmdp,doshivelez2013hipmdp}, and epistemic-POMDP \citep{ghosh2021epistemic} formalisms, but shifts the emphasis from solving a known model class to managing evidence inside a fixed context window.

We use \emph{\textbf{Non-Stationary ICRL}} for fixed-parameter, context-conditioned sequential decision making in which the decision process changes during the policy's context-bearing lifetime.
The deployed policy must use recent and stored interaction history to detect regime evidence, revise its latent task belief, and change its action distribution without test-time parameter updates.
The central difficulty is not only selecting the policy appropriate for the current regime, but also deciding which previous rewards, transitions, demonstrations, feedback, or memories still provide valid evidence for that choice.
It covers algorithm distillation \citep{laskin2023algorithm} and decision-pretrained transformers \citep{furuta2022gdt,liu2023agentic,polubarov2026vintixii} when interaction history drives adaptation; supervised ICRL \citep{lee2023supervised,lin2023transformers} and offline-pretrained agents \citep{tarasov2025qlearning,wang2024tdicrl} when task examples, rewards, or trajectories identify the current regime; retrieval-augmented ICRL \citep{schmied2024radt,sridhar2024regent} and retrieval-based RL \citep{goyal2022retrievalrl,humphreys2022largescale} when memory selection decides what evidence the model sees; and value-aware agents \citep{liu2026sicql,berkes2026spice} or model-aware agents \citep{son2025dicp,xu2026icprl} when values or dynamics are inferred from context.
It leaves out standard offline RL policies that do not adapt to new test-time evidence, deployment-time fine-tuning, and domain-specific robotics or multi-agent surveys unless their mechanisms clarify the fixed-parameter context problem.
We treat LLM agents and broader foundation-agent systems as bridge cases when they close the loop between context, action, feedback, and future behavior.
Pure conversational memory, retrieval-augmented question answering, or one-shot prompting is therefore outside the core claim unless the retained context changes decisions in a repeated environment.
The closest bodies of work are therefore not one lineage but several intersecting ones: lifelong and continual ICRL \citep{xu2024psbl,wang2025cicrl}; non-stationary MDP analyses \citep{lecarpentier2019nonstationary,chandak2020optimizing,wei2021nonstationary}, dynamic-regret analyses \citep{luo2024act,pettet2024policysearch,chen2025dynamicregret}, and near-optimal non-stationary control \citep{mao2021nearoptimal}; value- and uncertainty-aware ICRL \citep{tarasov2025qlearning,liu2026sicql,berkes2026spice}, compositional or suboptimal-data ICRL \citep{xu2026icql,wang2026suboptimalicrl}, and model-aware ICRL \citep{son2025dicp,xu2026icprl}; retrieval-augmented ICRL \citep{schmied2024radt,sridhar2024regent}, history curation \citep{chen2025filtering}, and retrieval-based RL \citep{goyal2022retrievalrl,humphreys2022largescale}; large heterogeneous sequence-agent datasets \citep{nikulin2025xland,wang2025anymdp}, generalist sequence agents \citep{polubarov2025vintix,polubarov2026vintixii}, and mixture-of-experts routing \citep{wu2025t2mir}; and reward-feedback agents \citep{monea2024bandit,song2026reward}, safety-feedback agents \citep{chandak2020safepolicyimprovement,moeini2025safeicrl}, and barrier-style safe ICRL \citep{kwon2026qbarrier}.

The non-stationary lens changes how the field should be evaluated.
Many current ICRL evaluations test held-out task generalization or average improvement with longer prompts.
Those tests are useful, but they do not by themselves show whether an agent can recover after a shift, reject stale demonstrations, reuse old context when a mode recurs, or operate under a finite context budget.
For non-stationary ICRL, evaluation should report lifecycle curves, post-shift recovery, dynamic regret, stale-context sensitivity, retrieval utility, and adaptation under controlled changes in shift frequency and observability.
Recent work begins to address these issues directly through lifelong ICRL \citep{xu2024psbl}, continual ICRL \citep{wang2025cicrl}, classical non-stationary RL \citep{cheung2020nonstationary,fei2020dynamic,wei2021nonstationary}, near-optimal non-stationary control \citep{mao2021nearoptimal}, bandit tracking \citep{garivier2011ucb,russac2019weighted}, and dynamic-regret analyses \citep{chen2025dynamicregret}, but \textbf{the area still lacks a shared taxonomy and reporting standard.}

\begin{figure}[!htbp]
\centering
\includegraphics[width=0.85\linewidth]{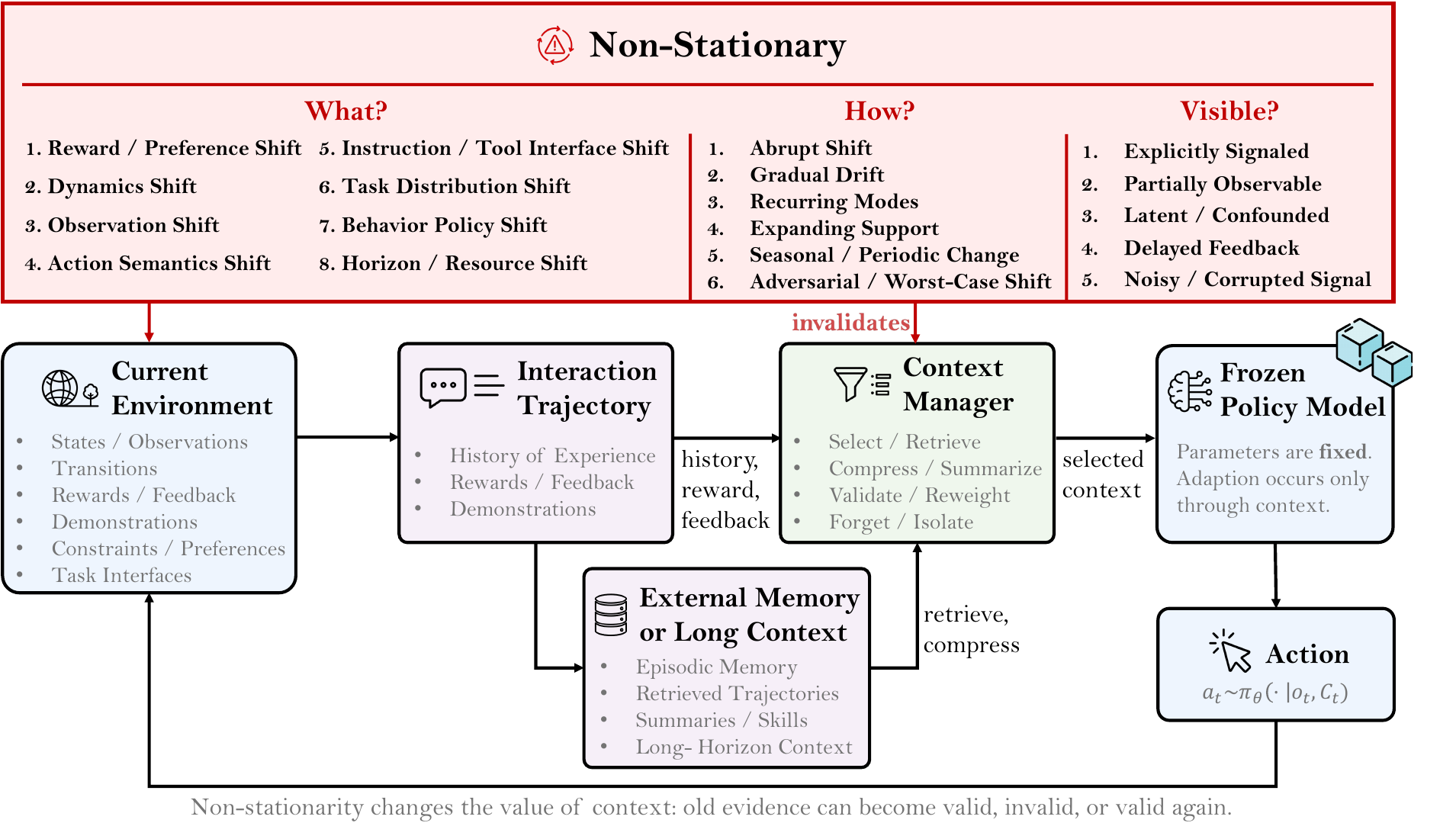}
\caption{Survey frame for non-stationary ICRL. A fixed-parameter policy model must infer and act from interaction trajectories while the decision process and the validity of accumulated evidence change.}
\label{fig:survey-frame}
\end{figure}

\begin{figure}[!htbp]
\centering
\includegraphics[width=0.8\linewidth]{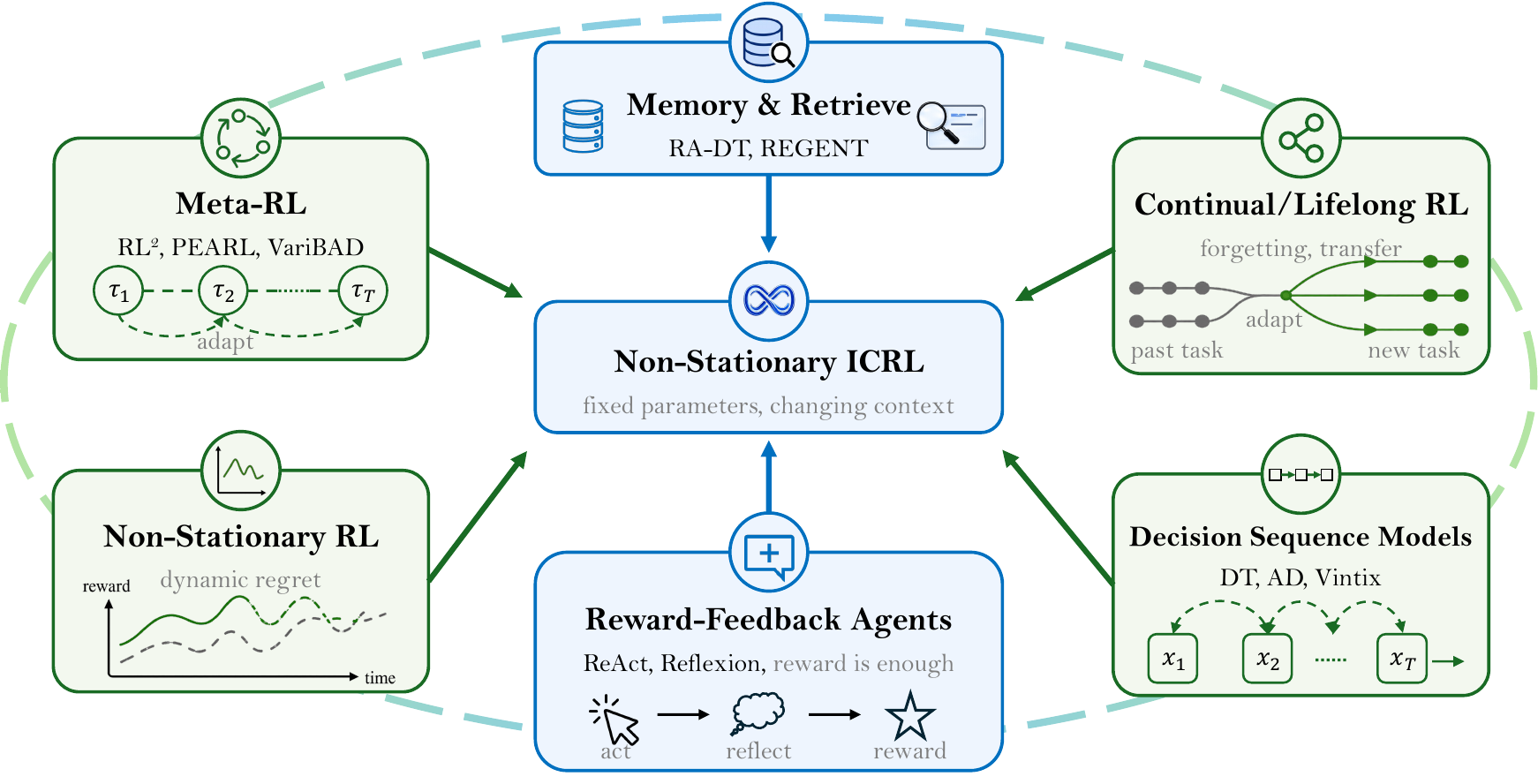}
\caption{Scope map for this survey, not a timeline. The paper studies the intersection where sequential decision agents adapt through context while the data-generating process changes; adjacent fields enter only when they clarify this mechanism.}
\label{fig:literature-domain-map}
\end{figure}

Figure~\ref{fig:survey-frame} gives the organizing picture.
The frozen decision model is only one part of the system.
The context pipeline decides which observations, demonstrations, rewards, retrieved memories, and summaries reach that model at each decision point.
Once the environment changes, this pipeline is both the adaptation mechanism and a common source of failure.
ICRL systems should therefore be evaluated by how their policy model, interaction history, memory, and feedback loop behave together under shifts.
Table~\ref{tab:survey-difference} makes this distinction explicit by contrasting the scope of this paper with a general ICRL survey.
Table~\ref{tab:related-surveys} then positions the paper against adjacent survey literatures, and Figure~\ref{fig:literature-domain-map} shows the resulting domain boundary.

\begin{table}[htbp]
\caption{Scope contrast between general ICRL surveys and this non-stationary ICRL survey.}
\label{tab:survey-difference}
\centering
\begingroup
\footnotesize
\setlength{\tabcolsep}{4pt}
\renewcommand{\arraystretch}{1.06}
\begin{tabularx}{\linewidth}{@{}
  >{\hsize=0.48\hsize\linewidth=\hsize\bfseries}Y
  >{\hsize=1.24\hsize\linewidth=\hsize}Y
  >{\hsize=1.26\hsize\linewidth=\hsize}Y
@{}}
\toprule
\textbf{Dimension} & \textbf{General ICRL Survey} & \textbf{This Survey} \\
\midrule
Core Focus & Context-driven improvement in pretrained decision models. & Validity of accumulated context after environmental change. \\
Axis & Pretraining objectives, architectures, context formats, and theory. & Sources, temporal patterns, and observability of non-stationarity. \\
Evaluation & Few-shot improvement, held-out tasks, context length, and architecture comparisons. & Post-shift recovery, dynamic regret, stale-context sensitivity, memory selection, and re-adaptation. \\
Failure Mode & Failure to use context or generalize to new tasks. & Over-trusting stale, contradictory, or distributionally invalid context. \\
Open Problems & Better training, scaling, architectures, and mechanistic theory. & Change-point-aware context policies, validity-aware retrieval, adaptive forgetting, and finite-context theory under drift. \\
\bottomrule
\end{tabularx}
\endgroup
\end{table}

\FloatBarrier
\begingroup
\footnotesize
\setlength{\LTpre}{0.4em}
\setlength{\LTpost}{0.4em}
\setlength{\tabcolsep}{4pt}
\renewcommand{\arraystretch}{1.15}
\begin{longtable}{@{}
  >{\strut\raggedright\arraybackslash\bfseries\color{nsblue}}p{0.18\linewidth}
  >{\strut\raggedright\arraybackslash}p{0.43\linewidth}
  >{\strut\raggedright\arraybackslash}p{0.31\linewidth}
@{}}
\caption{Related survey landscape.}
\label{tab:related-surveys}\\
\toprule
\textbf{\textcolor{black}{Literature}} & \textbf{Representative Work and Emphasis} & \textbf{Gap Addressed Here} \\
\midrule
\endfirsthead
\caption[]{Related survey landscape (continued).}\\
\toprule
\textbf{\textcolor{black}{Literature}} & \textbf{Representative Work and Emphasis} & \textbf{Gap Addressed Here} \\
\midrule
\endhead
\bottomrule
\endlastfoot
General ICRL & ICRL surveys organize pretraining, context construction, architectures, evaluation, and theory for ICRL broadly \citep{moeini2025surveyicrl}. & Re-centers the discussion on context validity after environmental change rather than ICRL as a whole. \\
Continual RL & Continual RL studies lifelong adaptation, plasticity, forgetting, and transfer \citep{khetarpal2022continual,abel2023definition,hadsell2020embracing,caccia2022taskagnostic}. & Imposes the stricter ICRL constraint: the deployed decision model adapts through context, not weight updates. \\
Non-Stationary RL & Non-stationary bandits, MDP algorithms, regret analyses, and NS-Gym benchmarks address changing rewards, dynamics, and latent regimes \citep{besbes2014nonstationarybandit,padakandla2020survey,lecarpentier2019nonstationary,cheung2020nonstationary,fei2020dynamic,feng2023nonstationary,wei2021nonstationary,keplinger2025nsgym}. & Adds learned context construction, retrieval, and stale evidence as first-class variables. \\
Change Detection And Bandits & Changepoint inference, sliding-window or discounted bandits, and restart methods discount old samples or react to abrupt shifts \citep{adams2007changepoint,garivier2011ucb,russac2019weighted,trovo2020sliding,auer2002nonstochastic,auer2019adaptively}. & Treats these tools as context-validity mechanisms inside fixed-parameter agents. \\
Meta-RL & Black-box, gradient-based, belief-based, and non-stationary meta-RL learn fast adaptation through hidden state, gradients, or latent task beliefs \citep{duan2016rl2,wang2016learning,finn2017maml,beck2023metarl,rakelly2019pearl,zintgraf2020varibad,humplik2019taskinference,mendonca2020metarlshift,bing2022nonstationarymetarl}. & Connects this lineage to modern sequence models and finite context-management failures. \\
Offline And Sequence RL & Offline RL, transformer-in-RL surveys, and trajectory models learn from fixed datasets or tokenize trajectories as supervised sequences \citep{levine2020offline,prudencio2022offlinesurvey,li2023surveytransformersrl,chen2021decision,janner2021trajectory,emmons2022rvs}. & Separates dataset-scale sequence modeling from true test-time context adaptation. \\
ICL Mechanisms & Work on formatting, function learning, learnability, and implicit optimization studies how transformers infer new functions from examples \citep{chan2022datadistribution,min2022rethinking,garg2022transformers,akyurek2023whatlearning,wies2023learnability,vonoswald2023gradient,kirsch2022gpicl,akyurek2024contextlanguage}. & Transfers the mechanism question to sequential decisions with rewards, stale evidence, and feedback loops. \\
RL Generalization & Zero-shot and procedural generalization work studies held-out environments, overfitting control, and distribution shift \citep{kirk2023zeroshot,cobbe2020procgen}. & Requires within-lifetime context invalidation, not only train-test diversity. \\
LLM Agents & Agent surveys and reflection or reward-feedback agents emphasize planning, memory, tool use, and reward-guided improvement \citep{wang2024llmagentsurvey,yao2023react,shinn2023reflexion,song2026reward}. & Includes only sequential feedback loops where actions, rewards, tool outcomes, or memories change future decisions; excludes general chatbot memory, one-shot prompting, and RAG without control feedback. \\
Retrieval And History Curation & Retrieval-augmented and history-filtered ICRL systems decide which evidence reaches the policy model \citep{schmied2024radt,sridhar2024regent,chen2025filtering}. & Asks whether retained evidence is valid under the current regime, not merely similar or high-return. \\
Generalist Sequence Agents & Cross-domain agents, large ICRL datasets, and generalist sequence models scale task breadth, action interfaces, and trajectory-token modeling \citep{lee2022multigame,reed2022gato,nikulin2025xland,wang2025anymdp,polubarov2025vintix,polubarov2026vintixii,wu2025t2mir}. & Treats scale as useful but insufficient without shift-aware evaluation and context-validity tests. \\
\end{longtable}
\endgroup

The survey builds its argument in four steps.
Figure~\ref{fig:literature-domain-map} is a domain-boundary view: it identifies which literatures overlap in the problem studied here.
The survey first casts non-stationary ICRL as a context-validity problem.
It then separates non-stationarity into three questions that are often conflated: what changes, how the change unfolds, and how directly the change can be observed by a frozen context-conditioned agent.
On that basis, it reviews methods by the mechanism that manages context--raw histories, curated histories, retrieval, compression, value-aware inference, model-based planning, reflection, and forgetting--rather than by publication chronology.
Finally, it argues for evaluation protocols that expose adaptation rather than memorization or domain-randomization-style generalization, and it highlights problems that stationary benchmarks tend to miss: action-semantics shift, context poisoning, memory isolation, uncertainty-calibrated retrieval, and theory for finite, stale, or conflicting context.
The same source/pattern/observability vocabulary is used throughout the survey so that method claims, benchmark claims, and open problems can be stated at the level of shift actually being tested.

The rest of the survey is organized as follows.
Section~\ref{sec:background} defines ICRL and its boundaries.
Section~\ref{sec:problem} formalizes non-stationary ICRL and introduces context validity.
Section~\ref{sec:taxonomy} develops the taxonomy.
Section~\ref{sec:context-validity} explains why non-stationarity changes the role of context.
Section~\ref{sec:methods} reviews method families through the lens of context management.
Section~\ref{sec:training} discusses training data and objectives.
Section~\ref{sec:evaluation} proposes evaluation protocols and metrics.
Section~\ref{sec:theory} summarizes theoretical links and gaps.
Section~\ref{sec:open-problems} closes with design principles and open problems.

\clearpage
\section{Foundations of In-Context Reinforcement Learning}
\label{sec:background}

\subsection{Fixed-Parameter Test-Time Adaptation}

We use an operational definition, defining the concept through observable, actionable, and testable criteria.
An agent performs ICRL when a fixed-parameter decision model uses the current episode or a broader interaction history to infer latent task structure and improve future action selection at test time.
For an observation $o_t$, action $a_t$, and context $\ctx_t$, the policy can be written as
\begin{equation}
  \policy(a_t \mid o_t, \ctx_t), \qquad \theta \ \text{fixed at test time}.
  \label{eq:icrl-policy}
\end{equation}
The context is usually a finite serialization of interaction evidence,
\begin{equation}
  \ctx_t
  =
  \operatorname{pack}_B
  \left(
    \tau_{1:k},
    d_{1:\ell},
    q_{1:m},
    z_t
  \right),
  \qquad
  \tau_i=(o_1^i,a_1^i,r_1^i,\ldots,o_{H_i}^i),
  \label{eq:context-serialization}
\end{equation}
where $\operatorname{pack}_B$ denotes a budget-limited formatting, retrieval, or compression operator; $\tau_{1:k}$ are trajectories, $d_{1:\ell}$ are demonstrations or descriptions, $q_{1:m}$ are retrieved memories or summaries, and $z_t$ denotes other task hints or feedback.
The context $\ctx_t$ may contain recent trajectories, older episodes, demonstrations, rewards, return prompts, task descriptions, retrieved memories, compressed summaries, value estimates, natural-language feedback, or other tokens that influence the model's action.
The defining constraint is not the token type but the adaptation channel: the agent adapts through $\ctx_t$, not through gradient updates to $\theta$.
Merely conditioning on a prompt is therefore not enough; the context must carry evidence from which the model can change later decisions in a task-relevant direction.
Equivalently, deployment-time adaptation is a forward-pass map from evidence to behavior,
\begin{equation}
  \ctx_t \mapsto \policy(\cdot \mid o_t,\ctx_t),
  \qquad
  \theta_{t+1}=\theta_t=\theta,
  \label{eq:fixed-parameter-adaptation}
\end{equation}
whereas online RL or gradient-based meta-learning would allow $\theta_{t+1}\neq\theta_t$ after observing new experience.

This definition covers several lines of work.
Algorithm distillation trains a model on histories generated by a learning algorithm, so that the trained model can imitate the algorithm's within-lifetime improvement through its context \citep{laskin2023algorithm}.
Decision-pretrained transformers and supervised-pretraining approaches show that trajectory models can learn useful in-context decision rules from offline data \citep{chen2021decision,janner2021trajectory,emmons2022rvs}, prompting and generalized decision-transformer objectives \citep{xu2022prompting,furuta2022gdt}, supervised ICRL theory and pretraining \citep{lee2023supervised,lin2023transformers}, agentic or variable-action formulations \citep{liu2023agentic,sinii2024variable}, and large multi-task decision datasets \citep{nikulin2025xland,polubarov2025vintix,polubarov2026vintixii}.
Long-context meta-RL agents connect the same idea to recurrent or transformer policies trained by reinforcement learning over multiple tasks or episodes, including transformer meta-RL \citep{melo2022transformersmetarl}, AMAGO-style agents \citep{grigsby2023amago,grigsby2024amago2}, and RELIC-style long-context systems \citep{elawady2024relic}.
Retrieval-augmented agents make the context partly external by selecting relevant experience from a memory database, including retrieval-augmented decision transformers \citep{schmied2024radt}, REGENT-style retrieval \citep{sridhar2024regent}, and earlier retrieval-based RL systems \citep{goyal2022retrievalrl,humphreys2022largescale}.
History curation and value-aware variants extend this design space by filtering learning histories \citep{chen2025filtering}, routing heterogeneous decision tokens \citep{wu2025t2mir}, or replacing pure imitation with value-sensitive objectives \citep{tarasov2025qlearning,liu2026sicql,berkes2026spice}.
Efficient long-sequence alternatives such as S4-style state-space models \citep{lu2023s4icrl,david2023decisions4}, recurrent-memory and S5-style models \citep{bulatov2022recurrentmemory,smith2023s5}, gated or xLSTM-style recurrent attention \citep{beck2024xlstm}, and Mamba-style sequence models \citep{ota2024decisionmamba,gu2023mamba} are also relevant because non-stationary ICRL often asks the agent to retain long histories while still reacting quickly to recent evidence.

ICRL is therefore a behavioral property of the deployment loop, not a single architecture.
A transformer with a trajectory prompt may fail to perform ICRL if the prompt is ignored.
A recurrent policy may perform ICRL if its hidden state or explicit context implements fast adaptation.
A large language model belongs in scope only when it is embedded in a sequential decision process with action, feedback, and reward-like signals, rather than merely recalling conversational state.
This includes LLM bandit agents \citep{monea2024bandit}, reward-feedback agents \citep{song2026reward}, ReAct-style agents that act in an environment \citep{yao2023react}, reflection or experience-library systems whose stored lessons change later trials \citep{shinn2023reflexion,zhao2024expel,wang2023voyager}, and tool-use or computer-control agents whose future actions depend on previous tool outcomes \citep{tan2025cradle}.
It excludes static prompting \citep{brown2020language,min2022rethinking}, one-shot retrieval-augmented question answering \citep{lewis2020rag}, generic chatbot memory \citep{zhong2023memorybank,wang2023longmem,packer2023memgpt}, and summaries that do not affect a repeated decision loop.
In latent-task notation, many ICRL systems can be read as amortizing a Bayes-adaptive control rule:
\begin{equation}
  b_t(\xi)
  =
  p(\xi \mid \ctx_t)
  \propto
  p(\xi)
  \prod_{(o,a,r,o') \in \ctx_t}
  p_\xi(o',r \mid o,a),
  \qquad
  \pi^\star(a \mid o,\ctx_t)
  =
  \int \pi_\xi^\star(a \mid o)\, b_t(\xi)\, d\xi,
  \label{eq:latent-task-belief}
\end{equation}
where $\xi$ is a hidden task, dynamics, reward, or goal parameter.
The learned policy need not explicitly represent $b_t$, but successful ICRL behavior requires some computation with the same effect: evidence in context changes the posterior over which decision rule is currently appropriate.
Recent theory strengthens this operational reading by showing that transformers can represent policy-improvement-like computations from in-context trajectories \citep{liang2026policyimprovement}, complementing chain-of-thought and temporal-difference analyses of ICRL \citep{xie2026cot,wang2024tdicrl}.
A fixed-parameter model can still implement an update-like decision rule inside the forward pass, so ICRL should not be dismissed as static behavior cloning.
The guarantees usually rely on coherent context drawn from a stable task family; non-stationary ICRL asks what happens when the same context window also contains obsolete, conflicting, or regime-mismatched evidence.
This behavioral definition is consistent with broader ICL work showing that in-context behavior can arise from distributional structure \citep{chan2022datadistribution}, meta-training and function learning \citep{garg2022transformers,akyurek2023whatlearning,wies2023learnability}, implicit linear-model fitting or gradient-descent-like computations \citep{vonoswald2023gradient,akyurek2024contextlanguage}, and demonstration formatting \citep{kirsch2022gpicl,min2022rethinking}.
For this survey, those results are useful background but not sufficient: reinforcement learning adds actions, rewards, exploration, credit assignment, and the possibility that old examples become actively wrong.

This behavioral definition also avoids a common ambiguity in the literature.
A method can be transformer-based, pretrained on trajectories, and evaluated on held-out tasks without demonstrating ICRL.
The decisive test is counterfactual: if the same observation is paired with different valid contexts, does the policy change in the direction implied by those contexts?
One compact diagnostic is a context-contrast score,
\begin{equation}
  \begin{aligned}
  \Delta_{\mathrm{ctx}}(o;\ctx,\ctx')
  &=
  D\!\left(
    \policy(\cdot \mid o,\ctx)
    \,\middle\|\,
    \policy(\cdot \mid o,\ctx')
  \right), \\
  J(\policy;\ctx) &> J(\policy;\ctx')
  \quad
  \text{when $\ctx$ contains better current evidence}.
  \end{aligned}
  \label{eq:context-contrast}
\end{equation}
where $D$ may be total variation, KL divergence, or another distance between action distributions.
Large distributional change alone is not enough; the changed action distribution must improve the downstream decision objective under the task implied by the context.
For non-stationary ICRL, the counterfactual is stronger.
If the prompt contains both old and new evidence, can the policy privilege the evidence that is valid under the current regime?
This is the reason the survey treats context construction, retrieval, and forgetting as first-class parts of the agent.

\subsection{Nearby Paradigms}

ICRL overlaps with several mature paradigms but is not identical to them.
Black-box meta-RL trains an agent over a distribution of tasks such that within-episode or within-lifetime computation implements a learning procedure \citep{duan2016rl2,wang2016learning}.
ICRL can be seen as a modern sequence-modeling version of this idea, with explicit context windows, retrieval, and trajectory tokens.

Classical RL and planning texts frame the stationary baseline through MDPs and value functions \citep{puterman1994markov,sutton2018reinforcement}, while POMDP work shows how hidden state can be converted into belief-state inference \citep{kaelbling1998pomdp}.
Contextual MDPs \citep{hallak2015contextualmdp} and hidden-parameter MDPs \citep{killian2017hipmdp,doshivelez2013hipmdp,perez2020generalizedhipmdp} are especially close to meta-RL because a latent or hidden parameter controls rewards and dynamics across related tasks.
Epistemic-POMDP \citep{ghosh2021epistemic} and task-inference views \citep{humplik2019taskinference,ren2020ocean} make the same point from a different angle: what looks like poor generalization can be implicit partial observability over an unobserved task, regime, or data-collection process.
For non-stationary ICRL this perspective is useful but incomplete.
The latent task belief is not only uncertain; it can expire, conflict with new evidence, or become useful again when a previous regime returns.
This is why recurrent memory baselines \citep{heess2015memory,ni2022recurrentpomdp}, trajectory-contrastive context encoders \citep{wang2021contrastivecontext}, and non-stationary meta-RL methods \citep{mendonca2020metarlshift,bing2022nonstationarymetarl} are important bridge literatures for the survey.

The older recurrent meta-RL line, including RL$^2$ \citep{duan2016rl2} and learning-to-reinforcement-learn systems \citep{wang2016learning}, already made the key move of representing fast adaptation in hidden state rather than in test-time weight updates.
Attentive and latent-belief meta-RL methods such as SNAIL \citep{mishra2018snail}, PEARL \citep{rakelly2019pearl}, and VariBAD \citep{zintgraf2020varibad} sharpened the same idea by treating recent experience as evidence for a latent task or Bayes-adaptive belief.
Gradient-based meta-RL, represented by MAML-style training \citep{finn2017maml} and later meta-RL variants \citep{beck2023metarl}, is adjacent but distinct because its adaptation channel is usually an explicit inner-loop parameter update rather than a fixed-parameter forward pass.
The operational difference can be summarized as
\begin{equation}
  \underbrace{\pi_{\theta'}(a_t\mid o_t),\quad
  \theta'=\theta-\alpha \nabla_\theta \mathcal{L}(\theta;\tau_{1:k})}_{\text{gradient-based adaptation}}
  \qquad\text{versus}\qquad
  \underbrace{\policy(a_t\mid o_t,\ctx_t),\quad
  \theta \ \text{unchanged}}_{\text{in-context adaptation}} .
  \label{eq:adaptation-channel-comparison}
\end{equation}
Decision Transformer \citep{chen2021decision}, Trajectory Transformer \citep{janner2021trajectory}, RvS \citep{emmons2022rvs}, and Multi-Game Decision Transformer \citep{lee2022multigame} then made trajectory sequence modeling a scalable substrate for offline decision models.
Offline RL also learns from previously collected data \citep{levine2020offline,prudencio2022offlinesurvey}, but standard offline RL usually produces a policy that acts without using new test-time examples as context; conservative Q-learning \citep{kumar2020cql} and implicit Q-learning \citep{kostrikov2022iql} are therefore background for value objectives, not ICRL by themselves.

Continual RL studies agents that persist across changing experiences and often update parameters, replay buffers, or other long-lived state \citep{khetarpal2022continual,abel2023definition}.
Non-stationary RL studies algorithms for changing MDPs, often with change-point tests \citep{adams2007changepoint,chen2019nonstationarycontextual}, sliding windows or discounting \citep{garivier2011ucb}, bandit and MDP variation-budget guarantees \citep{besbes2014nonstationarybandit,cheung2020nonstationary,fei2020dynamic}, dynamic or near-optimal regret bounds \citep{feng2023nonstationary,wei2021nonstationary,mao2021nearoptimal}, transformer-based regret analysis \citep{chen2025dynamicregret}, or restart and policy-search schedules \citep{auer2019adaptively,padakandla2020nonstationary,lecarpentier2019nonstationary}.
Survey work \citep{padakandla2020survey} and policy-search variants \citep{chandak2020optimizing,luo2024act,pettet2024policysearch} further broaden this algorithmic view.
ICRL imposes the additional constraint that the main policy parameters are fixed at test time.

The distinction among continual RL, lifelong RL, and ICRL is therefore operational rather than merely terminological.
Continual RL asks how an agent keeps learning across a stream of tasks or changing environments \citep{khetarpal2022continual,abel2023definition}; the usual toolbox may include online gradient updates in persistent agents \citep{khetarpal2022continual,abel2023definition}, replay \citep{lopezpaz2017gem,chaudhry2019agem}, regularization \citep{kirkpatrick2017overcoming,zenke2017synaptic,aljundi2018memoryaware}, architectural expansion \citep{rusu2016progressive}, or explicit task-free segmentation \citep{caccia2022taskagnostic}.
Lifelong RL is often used more broadly for persistent agents that accumulate and reuse experience over long horizons \citep{hadsell2020embracing,lesort2020continualrobotics,chandak2020changingaction}; it emphasizes retention \citep{kirkpatrick2017overcoming,zenke2017synaptic,aljundi2018memoryaware}, forward transfer \citep{wolczyk2022disentangling,powers2021cora}, and recovery or reuse when older skills or modes become relevant again \citep{steinparz2022reactive,schmied2024radt,sridhar2024regent}.
ICRL, in contrast, asks whether the deployed decision model can adapt with its parameters fixed, by changing only the context, hidden state, retrieval set, or summary supplied to the forward pass.
Non-stationary ICRL is the intersection of these ideas: it inherits continual and lifelong RL's changing-world concern, but it restricts the adaptation mechanism to context management.
For a fixed context budget $B$, the ICRL boundary can be written as the constrained deployment problem
\begin{equation}
  \max_{g}
  \ \mathbb{E}\!\left[
    \sum_{t=1}^{T} \gamma^{t-1} r_t
  \right]
  \quad
  \text{s.t.}\quad
  \ctx_t=g(h_{1:t-1}),\ |\ctx_t|\le B,\ \theta\ \text{fixed},
  \label{eq:context-budget-objective}
\end{equation}
which places the burden of adaptation on what is selected, compressed, retrieved, or forgotten before the policy's forward pass.
Recent direct work on lifelong ICRL \citep{xu2024psbl} and continual ICRL \citep{wang2025cicrl}, together with adjacent non-stationary in-context-learning theory \citep{qin2026beyondstationarity}, makes this intersection increasingly explicit.

\begin{figure}[t]
\centering
\resizebox{0.96\linewidth}{!}{\begin{tikzpicture}[
  x=1cm,
  y=1cm,
  axis/.style={
    line width=3.2pt,
    line cap=round
  },
  tick/.style={
    line width=0.55pt,
    line cap=round
  },
  phase/.style={
    rounded corners=2pt,
    inner xsep=4pt,
    inner ysep=2pt,
    align=center,
    font=\scriptsize\bfseries,
    text=white
  },
  stagebox/.style={
    draw=#1,
    rounded corners=4pt,
    line width=0.65pt,
    fill=#1!6,
    minimum width=2.90cm,
    minimum height=1.55cm,
    text width=2.68cm,
    align=center,
    font=\scriptsize,
    inner sep=4pt
  },
  stagecircle/.style={
    circle,
    draw=white,
    line width=1.35pt,
    fill=#1,
    minimum size=0.60cm,
    text=white,
    font=\bfseries\large,
    inner sep=0pt
  },
  connector/.style={
    line width=0.65pt,
    densely dotted
  },
  citefont/.style={
    font=\tiny,
    text=timelinegray
  }
]
\definecolor{timelineblue}{RGB}{45,110,190}
\definecolor{timelineteal}{RGB}{38,139,140}
\definecolor{timelinegreen}{RGB}{86,143,64}
\definecolor{timelinepurple}{RGB}{119,83,166}
\definecolor{timelinedeep}{RGB}{83,72,150}
\definecolor{timelinegray}{RGB}{92,96,104}
\definecolor{timelinepaper}{RGB}{247,249,252}

\fill[rounded corners=7pt, fill=timelinepaper] (-0.55,-0.58) rectangle (16.90,4.38);
\draw[rounded corners=7pt, draw=timelinegray!18, line width=0.35pt]
  (-0.55,-0.58) rectangle (16.90,4.38);

\draw[axis, draw=timelineblue, -{Latex[length=2.7mm,width=2.0mm]}]
  (-0.25,0) -- (3.05,0);
\draw[axis, draw=timelineteal, -{Latex[length=2.7mm,width=2.0mm]}]
  (3.05,0) -- (6.30,0);
\draw[axis, draw=timelinegreen, -{Latex[length=2.7mm,width=2.0mm]}]
  (6.30,0) -- (9.55,0);
\draw[axis, draw=timelinepurple, -{Latex[length=2.7mm,width=2.0mm]}]
  (9.55,0) -- (12.80,0);
\draw[axis, draw=timelinedeep, -{Latex[length=3.2mm,width=2.4mm]}]
  (12.80,0) -- (16.52,0);

\draw[tick, draw=timelineblue!72]   (1.38,-0.31) -- (1.38,0.31);
\draw[tick, draw=timelineteal!72]   (4.63,-0.31) -- (4.63,0.31);
\draw[tick, draw=timelinegreen!72]  (7.88,-0.31) -- (7.88,0.31);
\draw[tick, draw=timelinepurple!72] (11.13,-0.31) -- (11.13,0.31);
\draw[tick, draw=timelinedeep!72]   (14.38,-0.31) -- (14.38,0.31);

\draw[connector, draw=timelineblue!75]   (1.38,2.20) -- (1.38,0.43);
\draw[connector, draw=timelineteal!75]   (4.63,2.20) -- (4.63,0.43);
\draw[connector, draw=timelinegreen!75]  (7.88,2.20) -- (7.88,0.43);
\draw[connector, draw=timelinepurple!75] (11.13,2.20) -- (11.13,0.43);
\draw[connector, draw=timelinedeep!75]   (14.38,2.20) -- (14.38,0.43);

\node[stagecircle=timelineblue]   at (1.38,0) {1};
\node[stagecircle=timelineteal]   at (4.63,0) {2};
\node[stagecircle=timelinegreen]  at (7.88,0) {3};
\node[stagecircle=timelinepurple] at (11.13,0) {4};
\node[stagecircle=timelinedeep]   at (14.38,0) {5};

\node[phase, fill=timelineblue]   at (1.38,3.82) {Meta-RL};
\node[phase, fill=timelineteal]   at (4.63,3.82) {Sequence models};
\node[phase, fill=timelinegreen]  at (7.88,3.82) {Explicit ICRL};
\node[phase, fill=timelinepurple] at (11.13,3.82) {Scale + memory};
\node[phase, fill=timelinedeep]   at (14.38,3.82) {Non-stationarity};

\node[stagebox=timelineblue, anchor=north] at (1.38,3.56)
  {\textbf{2016--2020}\\Hidden-state adaptation\\RL$^2$, PEARL, VariBAD\\[-1pt]
  {\tiny\citep{duan2016rl2,rakelly2019pearl,zintgraf2020varibad}}};

\node[stagebox=timelineteal, anchor=north] at (4.63,3.56)
  {\textbf{2021--2022}\\Trajectories as tokens\\DT, TT, Multi-Game DT\\[-1pt]
  {\tiny\citep{chen2021decision,janner2021trajectory,lee2022multigame}}};

\node[stagebox=timelinegreen, anchor=north] at (7.88,3.56)
  {\textbf{2022--2023}\\Learning in context\\AD, DPT, theory\\[-1pt]
  {\tiny\citep{laskin2023algorithm,xu2022prompting,lee2023supervised}}};

\node[stagebox=timelinepurple, anchor=north] at (11.13,3.56)
  {\textbf{2024--2025}\\Long-horizon context\\AMAGO, RA-DT, XLand\\[-1pt]
  {\tiny\citep{grigsby2023amago,schmied2024radt,nikulin2025xland}}};

\node[stagebox=timelinedeep, anchor=north] at (14.38,3.56)
  {\textbf{2025--2026}\\Drift-aware ICRL\\CICRL, SPICE, regret\\[-1pt]
  {\tiny\citep{wang2025cicrl,berkes2026spice,chen2025dynamicregret}}};
\end{tikzpicture}}
\caption{Historical lineage of ICRL-related method families. The timeline orders recurrent meta-RL, sequence decision models, retrieval-augmented agents, and explicit non-stationary ICRL by development history.}
\label{fig:history-timeline}
\end{figure}
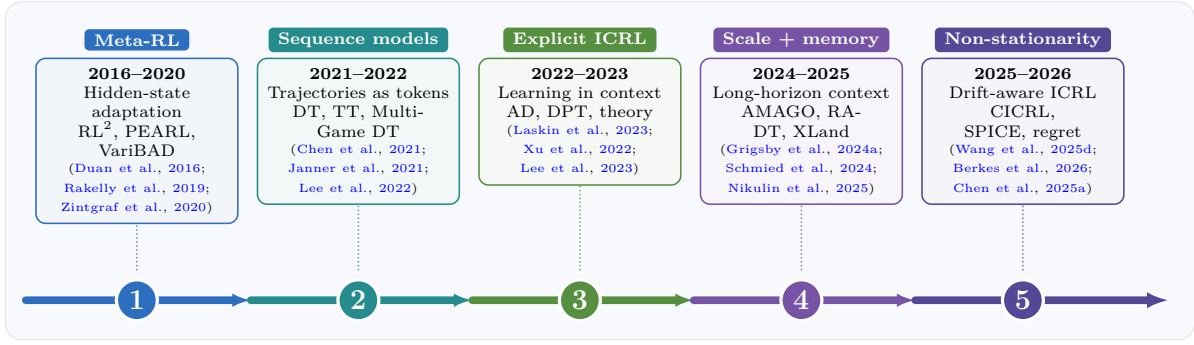

Figure~\ref{fig:history-timeline} gives a compact historical timeline for this lineage.
The important transition is from hidden-state adaptation in early meta-RL, to trajectory sequence modeling, to explicit ICRL, and finally to the current non-stationary turn where retrieval, value awareness, and lifecycle evaluation become central.
Table~\ref{tab:scope} uses this distinction to define the core, bridge, and excluded cases for the rest of the survey.

\begin{table}[H]
\caption{Scope boundary for this survey.}
\label{tab:scope}
\centering
\begingroup
\footnotesize
\setlength{\tabcolsep}{4pt}
\renewcommand{\arraystretch}{1.08}
\begin{tabularx}{\linewidth}{@{}
  >{\hsize=0.08\hsize\linewidth=\hsize\bfseries}Y
  >{\hsize=0.46\hsize\linewidth=\hsize}Y
  >{\hsize=0.46\hsize\linewidth=\hsize}Y
@{}}
\toprule
\textbf{Category} & \textbf{Included in This Survey} & \textbf{Boundary Condition} \\
\midrule
Core ICRL & Algorithm distillation, supervised ICRL, decision-pretrained transformers, long-context meta-RL, retrieval-augmented ICRL, value-aware ICRL. & Test-time behavior must improve or adapt from interaction context while model parameters fixed. \\
Bridge Cases & LLM reward-feedback agents, contextual-bandit ICRL, pure-exploration ICL, model-based in-context planning, long-memory agent systems. & Included only when the setting has sequential interaction and feedback relevant to future action \citep{monea2024bandit,krishnamurthy2024explore}; exploration and scaffolding variants are treated similarly \citep{dai2024icee,nie2025evolve}. \\
Excluded Cases & Pure supervised in-context learning, pure chatbot memory, standard offline RL, online fine-tuning, robotics-specific surveys, multi-agent adaptation. & Excluded because they lack the target adaptation channel or violate user-specified scope \citep{levine2020offline,prudencio2022offlinesurvey,luo2023modelbased}. \\
\bottomrule
\end{tabularx}
\endgroup
\end{table}

\subsection{Boundary Choices Under Non-Stationarity}

This survey draws the boundary around the adaptation channel rather than around the mere presence of non-stationarity: a method is central only when a fixed-parameter sequential decision agent must adapt to within-lifetime changes by selecting, compressing, retrieving, or forgetting context.
Boundary choices matter more under non-stationarity than under stationary task generalization.
If online parameter updates are allowed, adaptation can be attributed to continual learning rather than context use.
If the environment is not sequential, the problem becomes supervised in-context learning or contextual inference rather than reinforcement learning.
If the setting is purely offline and the agent receives no test-time context, then stale-context behavior cannot be studied.
This survey therefore uses a strict core definition and a conservative extended scope.
The aim is not to claim every context-conditioned model as ICRL, but to isolate the cases where changing context is the mechanism by which the agent adapts to changing decision processes.

The conservative boundary is also useful for negative results.
If a fixed model fails after a reward switch, the failure should not immediately be attributed to lack of model capacity.
It may be caused by a context policy that supplies the wrong evidence, a training distribution that never invalidates old context, or an evaluation protocol that rewards memorization.
Conversely, if a system succeeds only because it fine-tunes online, the result may be valuable for continual RL but does not answer the ICRL question.
Maintaining this boundary keeps the survey focused on the mechanisms that make adaptation through context possible or impossible while deployed policy parameters remain fixed.

\clearpage
\section{Problem Setting for Non-Stationary ICRL}
\label{sec:problem}

\subsection{From Stationary MDPs to Changing Decision Processes}

Let the agent interact with a sequence of decision processes
\begin{equation}
  \env_t = (\mathcal{S}_t, \mathcal{A}_t, P_t, R_t, \Omega_t, O_t, \gamma_t),
  \label{eq:changing-mdp}
\end{equation}
where $\mathcal{S}_t$ is the state space, $\mathcal{A}_t$ is the action space, $P_t$ is the transition kernel, $R_t$ is the reward function, $\Omega_t$ is the observation space, $O_t$ is the observation function, and $\gamma_t$ is the discount factor or horizon convention.
This notation extends the standard MDP formalism \citep{puterman1994markov} and allows partial observability \citep{kaelbling1998pomdp}, hidden contexts \citep{hallak2015contextualmdp}, or latent task parameters \citep{killian2017hipmdp,doshivelez2013hipmdp,ghosh2021epistemic} to enter through $O_t$ and through the unobserved variables that govern $P_t$ and $R_t$.
Non-stationarity means that at least one decision-relevant component changes over time or across phases of the same context-bearing lifetime, for example $P_t \neq P_{t+1}$, $R_t \neq R_{t+1}$, or a change in $O_t$, $\mathcal{A}_t$, goals, constraints, or action semantics.
Not every component must change.
The survey focuses on changes that alter which actions are good, which observations are informative, which memories are relevant, or which previous trajectories should be trusted.

At time $t$, the agent acts with a fixed policy class $\policy(a_t \mid o_t, \ctx_t)$.
The context $\ctx_t$ is produced by a context-management process:
\begin{equation}
  \ctx_t = g_\phi(h_{1:t-1}, m_{1:t-1}, b_t),
  \label{eq:context-manager}
\end{equation}
where $h_{1:t-1}$ denotes interaction history, $m_{1:t-1}$ denotes external memories or retrieved items, and $b_t$ denotes budget constraints such as context length, retrieval count, or latency.
In the strictest ICRL setting, both $\theta$ and $\phi$ are fixed after training.
Some systems use hand-designed retrieval or summarization rules \citep{schmied2024radt,sridhar2024regent}; others learn the context manager \citep{rakelly2019pearl,zintgraf2020varibad,chen2025filtering}.
The key distinction from continual RL remains that the adaptation channel at deployment is the constructed context, not online weight modification.
The policy must therefore infer task rules, reward meanings, transition structure, or action effects from history and express adaptation through future actions.

This formulation separates three objects that are often conflated.
The \emph{decision model} maps observations and context to actions.
The \emph{context manager} chooses what evidence is visible to the decision model.
The \emph{environment sequence} determines whether old evidence remains compatible with the present decision problem.
In stationary ICRL, the second object can often be treated as a formatting detail: include demonstrations, rewards, or previous episodes, and ask whether the model uses them.
In non-stationary ICRL, the context manager is a policy in its own right.
It must trade off recency, reliability, diversity, budget, and the possibility that a previously encountered regime has returned.
Classical non-stationary bandit and RL algorithms often encode this tradeoff through sliding windows and discount factors \citep{garivier2011ucb,russac2019weighted}, optimism bonuses \citep{besbes2014nonstationarybandit,cheung2020nonstationary,wei2021nonstationary}, restart schedules \citep{auer2019adaptively}, or explicit changepoint posteriors \citep{adams2007changepoint,chen2019nonstationarycontextual}.
The ICRL version differs because those operations may be implicit in the prompt, retrieval index, attention pattern, or memory state rather than in a hand-coded online estimator.

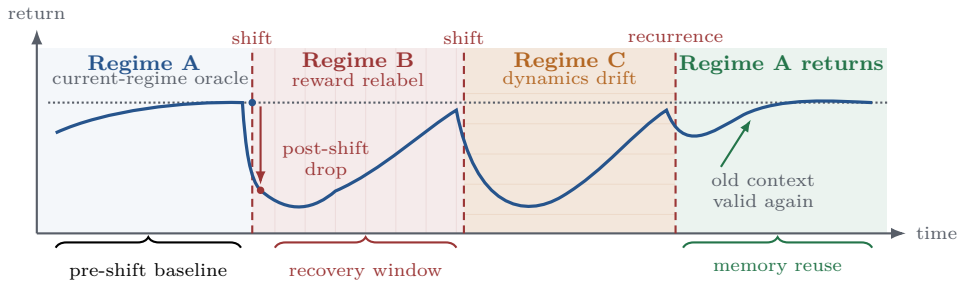
\begin{figure}[htbp]
\centering
\begin{tikzpicture}[
  x=1cm,
  y=1cm,
  every node/.style={font=\small},
  phase/.style={draw=none, rounded corners=2pt, minimum height=6mm, align=center, font=\footnotesize\bfseries},
  measure/.style={font=\scriptsize, align=center},
  axis/.style={-{Latex[length=2mm]}, thick, nsgray},
  shiftline/.style={densely dashed, thick, nsred},
  curve/.style={very thick, nsblue},
  oracle/.style={thick, nsgray, densely dotted},
  window/.style={decorate, decoration={brace, amplitude=4pt}, thick}
]
\fill[nslight] (0.4,0.45) rectangle (3.25,2.9);
\fill[nsred!10] (3.25,0.45) rectangle (6.05,2.9);
\fill[nsorange!16] (6.05,0.45) rectangle (8.85,2.9);
\fill[nsgreen!9] (8.85,0.45) rectangle (11.65,2.9);
\foreach \x in {3.55,3.95,...,5.95} {
  \draw[nsred!18, line width=0.35pt] (\x,0.45) -- (\x,2.9);
}
\foreach \y in {0.7,1.1,...,2.7} {
  \draw[nsorange!24, line width=0.35pt] (6.05,\y) -- (8.85,\y);
}

\node[phase, text=nsblue] at (1.82,2.68) {Regime A};
\node[phase, text=nsred] at (4.65,2.72) {Regime B};
\node[font=\scriptsize, text=nsred] at (4.65,2.49) {reward relabel};
\node[phase, text=nsorange] at (7.45,2.72) {Regime C};
\node[font=\scriptsize, text=nsorange] at (7.45,2.49) {dynamics drift};
\node[phase, text=nsgreen] at (10.25,2.68) {Regime A returns};

\draw[axis] (0.4,0.45) -- (11.9,0.45) node[right, font=\scriptsize] {time};
\draw[axis] (0.4,0.45) -- (0.4,3.15) node[above, font=\scriptsize, align=center] {return};
\draw[oracle] (0.55,2.18) -- (11.65,2.18);
\node[font=\scriptsize, nsgray, anchor=west] at (0.48,2.49) {current-regime oracle};

\draw[shiftline] (3.25,0.45) -- (3.25,2.9);
\draw[shiftline] (6.05,0.45) -- (6.05,2.9);
\draw[shiftline] (8.85,0.45) -- (8.85,2.9);
\foreach \x/\lab in {3.25/shift,6.05/shift,8.85/recurrence} {
  \node[font=\scriptsize, nsred, fill=white, inner sep=1pt] at (\x,3.04) {\lab};
}

\draw[curve]
  (0.65,1.78)
  .. controls (1.3,2.08) and (2.35,2.2) .. (3.12,2.18)
  .. controls (3.16,1.55) and (3.22,1.16) .. (3.36,1.02)
  .. controls (3.75,0.68) and (4.08,0.78) .. (4.35,1.01)
  .. controls (4.95,1.27) and (5.55,1.84) .. (5.95,2.08)
  .. controls (6.2,0.78) and (6.85,0.64) .. (7.35,0.94)
  .. controls (7.92,1.28) and (8.3,1.8) .. (8.73,2.08)
  .. controls (8.95,1.5) and (9.32,1.78) .. (9.75,2.02)
  .. controls (10.25,2.26) and (10.95,2.2) .. (11.45,2.18);

\fill[nsblue] (3.25,2.18) circle (1.4pt);
\fill[nsred] (3.36,1.02) circle (1.4pt);
\draw[-{Latex[length=1.8mm]}, nsred, thick] (3.36,2.13) -- (3.36,1.09);
\node[measure, nsred, anchor=west] at (3.52,1.42) {post-shift\\drop};

\draw[window, nsred] (3.55,0.26) -- (5.95,0.26)
  node[midway, below=2pt, measure, nsred] {recovery window};
\draw[window] (0.65,0.26) -- (3.1,0.26)
  node[midway, below=2pt, measure] {pre-shift baseline};
\draw[window, nsgreen] (8.95,0.26) -- (11.45,0.26)
  node[midway, below=2pt, measure, nsgreen] {memory reuse};

\node[measure, nsgray, anchor=west] at (9.2,1.0) {old context\\valid again};
\draw[-{Latex[length=1.7mm]}, thick, nsgreen] (9.45,1.33) -- (9.85,1.96);
\end{tikzpicture}
\caption{Lifecycle structure of a non-stationary ICRL episode. Average return hides where adaptation succeeds or fails.}
\label{fig:lifecycle}
\end{figure}

Figure~\ref{fig:lifecycle} frames the lifecycle perspective behind the adaptation metrics below, including dynamic regret, recovery time, context efficiency, and stale-context sensitivity.
A non-stationary benchmark should therefore reveal pre-shift competence, immediate post-shift degradation, recovery speed, and the reuse of context when a previous regime returns.

\subsection{Context Validity}

We define context validity as a decision-time property.
A context item $c_i \in \ctx_t$ is valid at time $t$ if conditioning on it improves, calibrates, or preserves decision quality under the current environment $\env_t$ compared with ignoring it.
It is invalid if it systematically worsens decisions, biases value estimates, hides a change point, or consumes scarce context budget without useful information.
One way to express this decision-relative notion is the marginal context value
\begin{equation}
  \nu_t(c_i)
  =
  \mathbb{E}\!\left[
    Q_t^{\policy}(s_t,a_t \mid \ctx_t)
    -
    Q_t^{\policy}(s_t,a_t \mid \ctx_t \setminus \{c_i\})
  \right],
  \label{eq:context-value}
\end{equation}
where the expectation is over the current state, action-selection randomness, and any retrieval randomness.
Positive $\nu_t(c_i)$ indicates helpful evidence, negative $\nu_t(c_i)$ indicates harmful or misleading evidence, and values near zero indicate context that may be syntactically relevant but decision-irrelevant.
This definition is intentionally performance-relative: old context is not invalid merely because it is old, and recent context is not valid merely because it is recent.
For recurring modes, an old trajectory may become valid again \citep{khetarpal2022continual,abel2023definition,xu2024psbl}.
For abrupt reward relabeling, even a high-return recent trajectory may be misleading \citep{besbes2014nonstationarybandit,mendonca2020metarlshift,song2026reward}.

This definition deliberately differs from syntactic relevance.
A trajectory can be syntactically similar to the current observation stream while being decision-invalid because the reward changed \citep{schmied2024radt,berkes2026spice}.
A demonstration can be high quality under its original regime while being invalid under a new action interface \citep{chandak2020changingaction,sinii2024variable}.
A recent item can be invalid because it was produced by an exploratory or corrupted behavior policy \citep{dorfman2020offlineexploration,fu2020d4rl,levine2020offline}.
The survey therefore uses validity as a causal and decision-theoretic notion: context is valid when it improves the action distribution for the present regime, not when it looks similar, recent, or successful in isolation.

The validity lens gives a compact statement of the non-stationary ICRL problem:
\begin{takeawayquote}
  A \textbf{non-stationary ICRL} agent must adapt through context while estimating which previously inferred rules, strategies, and memories still describe the current data-generating process.
\end{takeawayquote}
This is a stronger requirement than stationary few-shot generalization.
It requires the model or its context manager to solve a joint problem of inference, memory selection, and control.

\subsection{Objectives}

The objective can be expressed in several complementary ways.
The standard cumulative-reward objective is
\begin{equation}
  J(\policy, g) = \mathbb{E}\left[\sum_{t=1}^{T} \gamma_t r_t \right],
\end{equation}
where actions are produced by $\policy(a_t \mid o_t, g(h_{1:t-1},m_{1:t-1},b_t))$.
For non-stationary environments, average return hides important lifecycle behavior.
Two agents can have similar average return while one recovers immediately after shifts and the other fails catastrophically but performs well between shifts.
Ergodicity-focused work gives a second reason to distrust a single average-return number: when reward processes are non-ergodic, ensemble expectations can diverge from the time-average experience of an individual deployed agent \citep{baumann2026ergodicity,verbruggen2026nonergodic}.
In such settings, optimizing or reporting expected return can favor a policy that looks good across many hypothetical runs while producing poor long-run outcomes on the actual trajectory that carries the agent's context.
This concern is especially acute for ICRL because the context window is itself trajectory-dependent; early failures, stale memories, or unlucky regime exposure can change the evidence available for later decisions.
Dynamic regret is often more diagnostic:
\begin{equation}
  \mathrm{Reg}_T^{\mathrm{dyn}}
  =
  \sum_{t=1}^{T} V_t^{\pi_t^\star}(s_t)
  -
  \sum_{t=1}^{T} V_t^{\policy}(s_t),
  \label{eq:dynamic-regret}
\end{equation}
where $\pi_t^\star$ is an oracle policy for the current environment or current latent mode.
This objective follows the same motivation as adversarial and non-stationary bandit work \citep{auer2002nonstochastic,besbes2014nonstationarybandit,li2019online}, non-stationary MDP regret analyses \citep{cheung2020nonstationary,fei2020dynamic,wei2021nonstationary}, near-optimal or kernelized variants \citep{mao2021nearoptimal,domingues2021kernel}, and non-ergodic RL critiques of expectation-based objectives \citep{baumann2026ergodicity,verbruggen2026nonergodic}: a static comparator or ensemble average can make an agent look good simply because the environment changes or because failures are averaged away, whereas a dynamic comparator exposes the price paid for delayed detection, adaptation, and context repair along the realized lifetime.
The amount of environmental movement can be summarized by a variation budget, for example
\begin{equation}
  B_T
  =
  \sum_{t=2}^{T}
  \left(
  \|P_t-P_{t-1}\|_{\infty}
  +
  \|R_t-R_{t-1}\|_{\infty}
  \right),
  \label{eq:variation-budget}
\end{equation}
with analogous terms added when observation functions, action semantics, constraints, or task priors change.
This budget is not the only possible formalization, but it makes explicit that a method's adaptation claim depends on how often and how much the environment changes.
Post-shift recovery time, post-shift area under the learning curve, context efficiency, and stale-context sensitivity provide additional views that are more directly connected to ICRL.
Given a shift at $\tau$, one useful recovery statistic is
\begin{equation}
  T_{\mathrm{rec}}(\tau,\alpha)
  =
  \min\left\{k \geq 0:
  \frac{1}{w}\sum_{i=0}^{w-1} J_{\tau+k+i}^{\policy}
  \geq
  \alpha
  \frac{1}{w}\sum_{i=0}^{w-1} J_{\tau+k+i}^{\pi^\star}
  \right\},
  \label{eq:recovery-time}
\end{equation}
where $\alpha$ is a target fraction of oracle performance and $w$ is a smoothing window.
These objectives are summarized in Table~\ref{tab:objectives}, which shows why stationary ICRL metrics are insufficient once context validity changes over time.

\begin{table}[H]
\caption{Stationary ICRL and non-stationary ICRL emphasize different objectives.}
\label{tab:objectives}
\centering
\begingroup
\SurveyTableSetup
\begin{tabularx}{\linewidth}{@{}B{0.24\linewidth}
    @{}>{\raggedright\arraybackslash}p{0.3\linewidth}Y@{}}
\toprule
\textbf{Objective or Metric} & \textbf{Stationary ICRL Interpretation} & \textbf{Non-Stationary ICRL Interpretation} \\
\midrule
Final Return & Can the agent infer and exploit a fixed task from context \citep{lee2023supervised,lin2023transformers}? & Can the agent eventually recover after changes \citep{xu2024psbl,wang2025cicrl}? \\
Few-Shot Improvement & Does more task-relevant context improve performance \citep{xu2022prompting,raparthy2023generalization}? & Does the agent distinguish helpful new evidence from stale old evidence? \\
Context Length Scaling & Does a longer prompt help \citep{grigsby2023amago,grigsby2024amago2}? & Does a longer prompt help only when invalid context is filtered or downweighted \citep{chen2025filtering,schmied2024radt}? \\
Dynamic Regret & Usually secondary or absent. & Measures cost relative to a changing comparator policy \citep{besbes2014nonstationarybandit,feng2023nonstationary,chen2025dynamicregret}. \\
Recovery Time & Often not measured. & Central measure of adaptation after reward, dynamics, observation, or preference shifts \citep{adams2007changepoint,wang2025cicrl}. \\
Stale-Context Sensitivity & Rarely tested. & Measures whether outdated high-confidence context harms decisions \citep{schmied2024radt,berkes2026spice}. \\
\bottomrule
\end{tabularx}
\endgroup
\end{table}

\subsection{Non-Stationarity Is Not Domain Randomization}

Held-out task evaluation and domain randomization both test generalization across tasks or environment parameters, but they do not by themselves test non-stationary ICRL.
In non-stationary ICRL, the task may change during the agent's lifetime, so the agent must adjust its behavior from context rather than from a reset, retraining, or a clean task descriptor.
A model can therefore generalize well to a new stationary task while still failing when the task changes halfway through an episode; conversely, it may handle recurring change points in a narrow environment family without achieving broad held-out generalization.

This distinction is a useful diagnostic for benchmarks.
If changes occur only between episodes and the agent receives a clean task descriptor at the start of each episode, the benchmark mainly tests task-conditioned generalization.
If the task changes within a lifetime but the prompt or history is reset at the change point, it tests fast reconditioning rather than robustness to stale context \citep{wang2025cicrl,chen2025filtering}.
The central non-stationary ICRL regime arises when old and new regimes coexist in the same context window, forcing the agent to decide which evidence remains valid \citep{schmied2024radt,berkes2026spice}.
Papers reporting non-stationary ICRL results should therefore specify which regime they evaluate.

\clearpage
\section{A Taxonomy of Non-Stationarity for ICRL}
\label{sec:taxonomy}

Non-stationarity is too broad a label to support useful comparisons by itself.
The term can refer to qualitatively different stressors, including announced or inferable reward changes in non-stationary bandits and MDPs \citep{garivier2011ucb,besbes2014nonstationarybandit,fei2020dynamic,song2026reward}, gradual variation in transition dynamics \citep{lecarpentier2019nonstationary,chandak2020optimizing,feng2023nonstationary}, recurrent latent regimes in contextual or hidden-parameter MDPs \citep{hallak2015contextualmdp,killian2017hipmdp,doshivelez2013hipmdp}, and changes in the action interface or action-to-effect mapping \citep{chandak2020changingaction,sinii2024variable}.
These regimes do not exercise the same adaptation mechanism.
An append-all history may be adequate for gradual reward estimation and harmful after abrupt action remapping.
A retrieval-augmented model may be strong on recurring modes and still fail under subtle observation drift if its retrieval key is calibrated to the wrong surface features \citep{thakur2021beir,yoran2023robustretrieval}.
A value-aware model may react to reward shifts while remaining brittle when transition dynamics move outside its support \citep{fujimoto2019offpolicy,kidambi2020morel,yu2020mopo}.

We therefore describe a non-stationary ICRL setting along three axes: the component that changes, the temporal structure of the change, and the evidence by which the agent can observe it.
This is narrower than general RL generalization taxonomies, which mainly ask whether a trained policy transfers to held-out environments; here the issue is whether a fixed model can keep using, rejecting, or reactivating context during a lifetime \citep{kirk2023zeroshot}.
It is also more operational than POMDP and contextual-MDP formalisms \citep{kaelbling1998pomdp,hallak2015contextualmdp}, epistemic-POMDP views \citep{ghosh2021epistemic}, and hidden-parameter MDP formalisms \citep{killian2017hipmdp,doshivelez2013hipmdp}.
Those formalisms justify treating task identity or regime as latent.
They do not by themselves say how a finite prompt, memory index, or retrieval policy should be managed under drift.
They also do not reduce non-stationary ICRL to domain randomization, because the relevant change happens inside the interaction trajectory rather than only between sampled training or test instances.

\begin{figure}[htpb]
\centering
\includegraphics[width=0.8\linewidth]{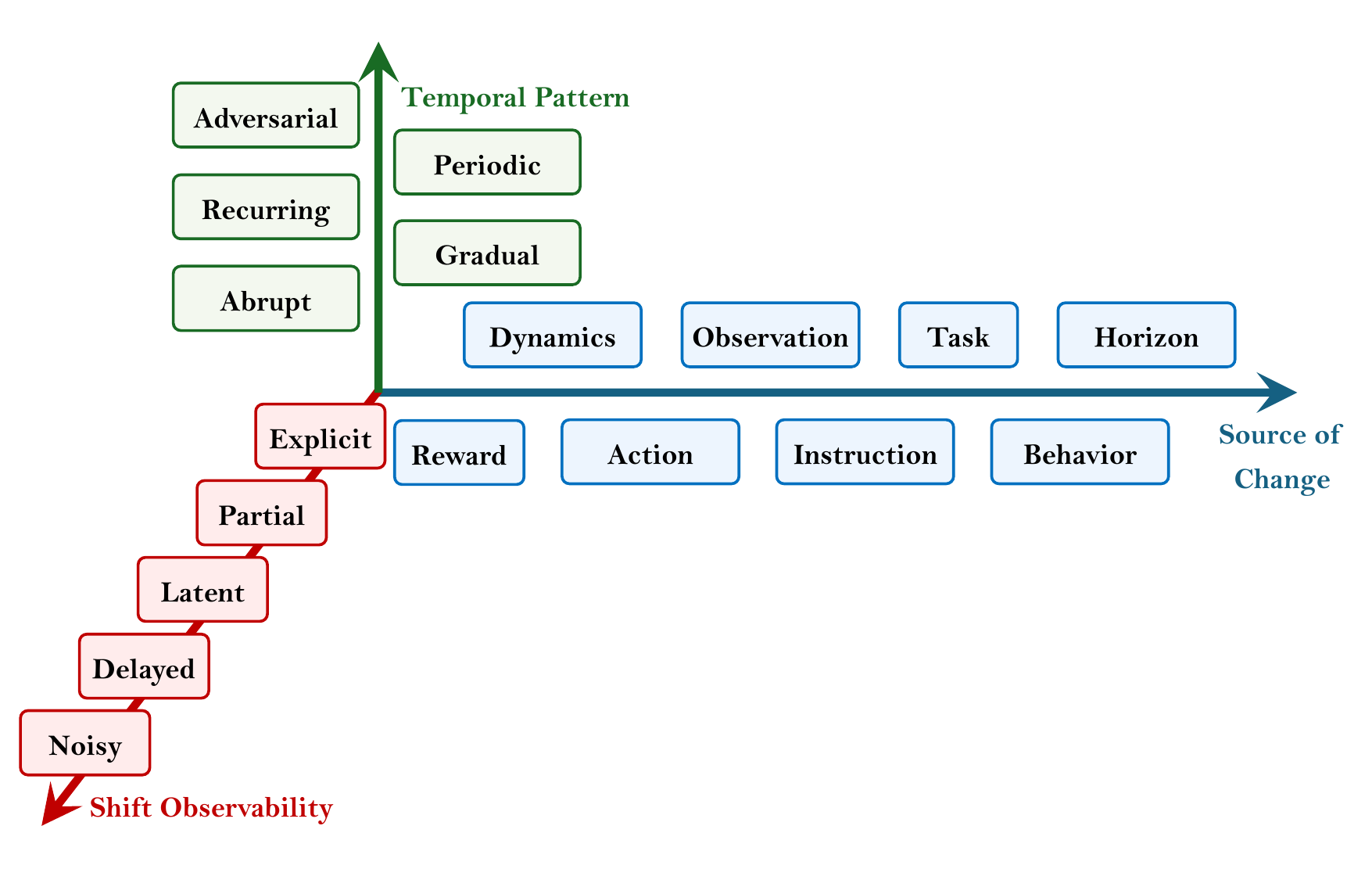}
\caption{Compact map of three independent axes for describing non-stationarity in ICRL. Short labels indicate representative categories; Tables~\ref{tab:what-changes}--\ref{tab:observability} give the full definitions.}
\label{fig:taxonomy-map}
\end{figure}

\begin{figure}[htpb]
\centering
\resizebox{\linewidth}{!}{\begingroup
\newcommand{\taxcite}[1]{{\tiny\citep{#1}}}

\definecolor{macbg}{RGB}{250,247,242}
\definecolor{macink}{RGB}{68,76,92}
\definecolor{macline}{RGB}{178,185,196}

\definecolor{macblue}{RGB}{118,176,218}
\definecolor{macgreen}{RGB}{122,199,176}
\definecolor{macpurple}{RGB}{198,184,228}      

\definecolor{macblueDark}{RGB}{82,139,186}
\definecolor{macgreenDark}{RGB}{82,157,137}
\definecolor{macpurpleDark}{RGB}{156,137,198}  

\begin{tikzpicture}[
  x=1cm,
  y=1cm,
  every node/.style={font=\scriptsize},
  root/.style={
    draw=macline!45,
    rounded corners=6pt,
    line width=0.65pt,
    align=center,
    fill=white,
    inner xsep=12pt,
    inner ysep=6pt,
    font=\small\bfseries,
    text=macink
  },
  axislabel/.style={
    rounded corners=4pt,
    minimum width=3.75cm,
    minimum height=0.58cm,
    align=center,
    font=\scriptsize\bfseries,
    text=white,
    inner sep=2pt
  },
  leaf/.style 2 args={
    draw=#1!58!macink,
    rounded corners=4pt,
    line width=0.48pt,
    fill=#1!13,
    text width=3.48cm,
    minimum height=1.00cm,
    align=left,
    inner xsep=5pt,
    inner ysep=3pt,
    font=\scriptsize,
    text=macink,
    path picture={
      \fill[#1!78]
        (path picture bounding box.north west)
        rectangle ([xshift=2.4pt]path picture bounding box.south west);
    }
  },
  stem/.style={
    line width=0.58pt,
    draw=macline!62,
    rounded corners=2pt
  },
  tick/.style={
    line width=0.58pt
  }
]

\fill[rounded corners=9pt, macbg] (-0.62,-6.95) rectangle (12.42,0.64);
\draw[rounded corners=9pt, draw=macline!28, line width=0.4pt]
  (-0.62,-6.95) rectangle (12.42,0.64);

\node[root] (root) at (5.9,0.14) {Non-stationary ICRL};

\coordinate (sourceTop) at (1.55,-1.02);
\coordinate (patternTop) at (5.90,-1.02);
\coordinate (visibleTop) at (10.25,-1.02);

\draw[stem] (root.south) -- +(0,-0.28) -| (sourceTop);
\draw[stem] (root.south) -- +(0,-0.28) -| (patternTop);
\draw[stem] (root.south) -- +(0,-0.28) -| (visibleTop);

\node[axislabel, fill=macblueDark] (source) at (1.55,-1.02) {What changes?};
\node[axislabel, fill=macgreenDark] (pattern) at (5.90,-1.02) {How does it change?};
\node[axislabel, fill=macpurpleDark] (visible) at (10.25,-1.02) {How visible is it?};

\node[leaf={macblue}{}, anchor=north] (s1) at (1.55,-1.52)
  {\textbf{Reward / Preference}\\reward feedback; safe ICRL\\[-1pt]\taxcite{song2026reward}};
\node[leaf={macblue}{}, anchor=north] (s2) at (1.55,-2.95)
  {\textbf{Dynamics / Observation}\\DICP; non-stationary RL\\[-1pt]\taxcite{lecarpentier2019nonstationary,son2025dicp}};
\node[leaf={macblue}{}, anchor=north] (s3) at (1.55,-4.38)
  {\textbf{Action Interface}\\changing action sets\\[-1pt]\taxcite{chandak2020changingaction}};
\node[leaf={macblue}{}, anchor=north] (s4) at (1.55,-5.81)
  {\textbf{Task / Data Source}\\PSBL; AnyMDP; XLand\\[-1pt]\taxcite{xu2024psbl,nikulin2025xland}};

\node[leaf={macgreen}{}, anchor=north] (p1) at (5.90,-1.52)
  {\textbf{Abrupt}\\change-point recovery\\[-1pt]\taxcite{adams2007changepoint,wang2025cicrl}};
\node[leaf={macgreen}{}, anchor=north] (p2) at (5.90,-2.95)
  {\textbf{Gradual}\\variation budget\\[-1pt]\taxcite{besbes2014nonstationarybandit}};
\node[leaf={macgreen}{}, anchor=north] (p3) at (5.90,-4.38)
  {\textbf{Recurring}\\retrieval and reuse\\[-1pt]\taxcite{schmied2024radt}};
\node[leaf={macgreen}{}, anchor=north] (p4) at (5.90,-5.81)
  {\textbf{Adversarial}\\stale-memory stress tests\\[-1pt]\taxcite{chen2025dynamicregret}};

\node[leaf={macpurple}{}, anchor=north] (v1) at (10.25,-1.52)
  {\textbf{Explicit}\\task or goal token\\[-1pt]\taxcite{xu2022prompting,lee2023supervised}};
\node[leaf={macpurple}{}, anchor=north] (v2) at (10.25,-2.95)
  {\textbf{Partial}\\reward or transition evidence\\[-1pt]\taxcite{kaelbling1998pomdp}};
\node[leaf={macpurple}{}, anchor=north] (v3) at (10.25,-4.38)
  {\textbf{Latent}\\ambiguous modes\\[-1pt]\taxcite{hallak2015contextualmdp,killian2017hipmdp}};
\node[leaf={macpurple}{}, anchor=north] (v4) at (10.25,-5.81)
  {\textbf{Delayed / Noisy}\\long-horizon evidence\\[-1pt]\taxcite{grigsby2024amago2,elawady2024relic}};

\draw[tick, draw=macblueDark!55] (source.south) -- ++(0,-0.16);
\draw[tick, draw=macgreenDark!55] (pattern.south) -- ++(0,-0.16);
\draw[tick, draw=macpurpleDark!45] (visible.south) -- ++(0,-0.16);

\end{tikzpicture}
\endgroup}
\caption{Tree-structured taxonomy for non-stationary ICRL. The leaves are representative problem families rather than mutually exclusive classes.}
\label{fig:taxonomy-tree}
\end{figure}
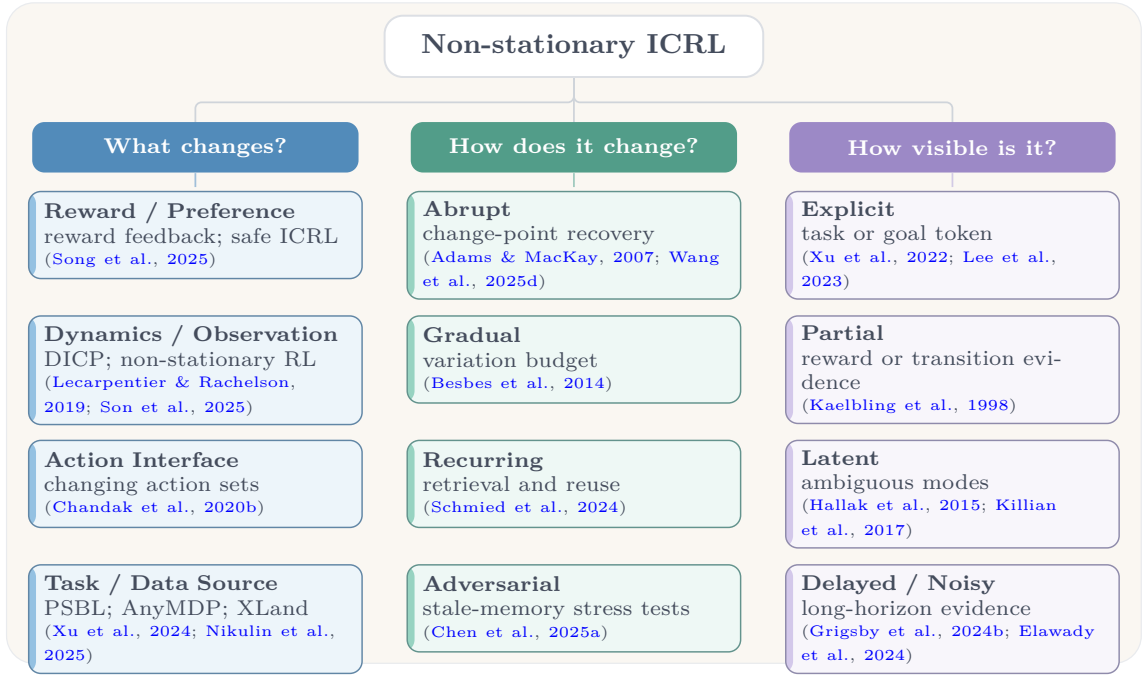

Figure~\ref{fig:taxonomy-map} shows a compact map of the axes, with representative labels for each dimension, and Figure~\ref{fig:taxonomy-tree} gives a tree view for placing papers and benchmarks. The full category definitions appear in Tables~\ref{tab:what-changes}--\ref{tab:observability}. A concrete non-stationary setting can be described along all three independent dimensions simultaneously. For example, a setting labeled as dynamics/recurring/latent denotes a task in which the transition function changes across recurring regimes, while the current regime is not directly observed and must be inferred from the interaction context. The main practical point is that these axes should be reported independently. Explicit reward switches and latent recurring dynamics are both non-stationary, but they are not interchangeable evidence for the same adaptation claim.

\subsection{Changing Components}

The first axis characterizes where change enters the decision process.
Table~\ref{tab:what-changes} summarizes the main cases.
The rows are not mutually exclusive, since deployed systems often face multiple forms of change at the same time.
The central question is which part of the accumulated context has become unreliable.

\begingroup
\SurveyLongTableSetup
\begin{longtable}{@{}
  B{0.14\linewidth}
  L{0.18\linewidth}
  L{0.22\linewidth}
  L{0.38\linewidth}@{}}
\caption{Taxonomy axis 1: sources of non-stationarity and their context consequences.}
\label{tab:what-changes}\\
\toprule
\textbf{Shift Source} & \textbf{Changing Object} & \textbf{Context Failure Mode} & \textbf{Typical Evaluation Probe} \\
\midrule
\endfirsthead
\caption[]{Taxonomy axis 1: sources of non-stationarity and their context consequences (continued).}\\
\toprule
\textbf{Shift Source} & \textbf{Changing Object} & \textbf{Context Failure Mode} & \textbf{Typical Evaluation Probe} \\
\midrule
\endhead
\bottomrule
\endlastfoot
Reward or Preference & $R_t$; goal; utility; constraint weights; user preference. & Previously good trajectories become wrong demonstrations; reward labels may need reinterpretation. & Reward relabeling and goal switches in non-stationary bandits or MDPs \citep{garivier2011ucb,besbes2014nonstationarybandit,fei2020dynamic}; preference reversal or reward-feedback changes \citep{song2026reward}; safety-cost shifts \citep{chandak2020safepolicyimprovement,moeini2025safeicrl}. \\
\addlinespace[2pt]
Dynamics & $P_t$; latent physics; transition stochasticity; exogenous process. & Old action--outcome evidence misleads planning, credit assignment, or value estimation. & Abrupt and gradual transition changes with fixed observations \citep{padakandla2020nonstationary,lecarpentier2019nonstationary,chandak2020optimizing}; regret and policy-search variants \citep{cheung2020nonstationary,feng2023nonstationary,luo2024act,pettet2024policysearch}; latent and model-based dynamics inference \citep{hallak2015contextualmdp,killian2017hipmdp,doshivelez2013hipmdp,son2025dicp,xu2026icprl}. \\
\addlinespace[2pt]
Observation & $O_t$; sensor mapping; feature availability; observability level. & Context describes a mode that is no longer identifiable through the same observed features. & Feature corruption, sensor remapping, and missing observations in POMDP-style settings \citep{kaelbling1998pomdp,ghosh2021epistemic}; recurrent memory probes \citep{heess2015memory,ni2022recurrentpomdp}; long-context generalization tests \citep{grigsby2024amago2,raparthy2023generalization}. \\
\addlinespace[2pt]
Action Semantics & $\mathcal{A}_t$; available actions; action-to-effect map. & Old action tokens no longer denote the same intervention. & Action permutation, action masking, and continuous-action rescaling \citep{chandak2020changingaction,sinii2024variable}. \\
\addlinespace[2pt]
Instruction or Tool Interface & System instructions; user constraints; tool schemas; website state; feedback format. & Old instructions, reflections, or tool-use memories no longer describe valid actions. & Hidden instruction changes, preference drift, tool-schema remapping, website distribution shift, and feedback reliability changes in interactive language environments \citep{yao2023react,wang2022scienceworld,yao2022webshop,shridhar2020alfworld,tan2025cradle}. \\
\addlinespace[2pt]
Task Distribution & Latent task sequence; support; mixture weights. & Pretraining or retrieval priors overweight outdated task modes. & Changing task families in meta-RL \citep{alshedivat2017continuous,mendonca2020metarlshift}; out-of-support lifelong ICRL and generated-MDP sequences \citep{xu2024psbl,wang2025anymdp,nikulin2025xland}. \\
\addlinespace[2pt]
Behavior Policy & Demonstrator; data collector; exploration or logging policy. & Retrieved or demonstrated trajectories reflect policy mismatch rather than current task evidence. & Mixed-quality demonstrations and weak-data histories in offline RL \citep{levine2020offline,kumar2020cql,kostrikov2022iql}; value-aware ICRL \citep{tarasov2025qlearning}; Bayesian context use \citep{berkes2026spice}. \\
\addlinespace[2pt]
Horizon or Resource & Episode length; discount; budget; latency; context length. & A context-use rule that works with long horizons fails when adaptation opportunities are scarce. & Short-horizon shift and delayed-feedback stress tests \citep{grigsby2023amago,grigsby2024amago2,elawady2024relic}; context-budget and retrieval ablations \citep{schmied2024radt}. \\
\end{longtable}
\endgroup

These sources of change stress different failure modes of context use.
Reward and preference shifts are the cleanest validity test: a trajectory that was optimal before a reward switch can become actively misleading afterward, especially for models trained to imitate high-return histories.
Dynamics and observation shifts test whether the agent infers latent environment structure from context rather than replaying familiar action sequences or overfitting to superficial tokens.
Action-semantics shifts are particularly sharp for sequence models, because the same action token or embedding may denote a different intervention after the interface changes.
Task-distribution, behavior-policy, and resource shifts instead alter the credibility of accumulated evidence: the agent must decide whether demonstrations, retrieved trajectories, and old priors still describe the current decision problem.

\subsection{Temporal Patterns of Change}

The second axis is temporal structure.
An abrupt change asks for change-point detection, gradual drift for tracking, recurring modes for memory indexing, and adversarial or strategically timed changes for robustness.
Table~\ref{tab:temporal-patterns} links these patterns to the context-management implication and to metrics that reveal the corresponding failure.

\begin{table}[htbp]
\caption{Taxonomy axis 2: temporal patterns of non-stationarity.}
\label{tab:temporal-patterns}
\centering
\begingroup
\SurveyTableSetup
\begin{tabularx}{\linewidth}{@{}B{0.18\linewidth}L{0.3\linewidth}Y@{}}
\toprule
\textbf{Temporal Pattern} & \textbf{Context-Management Demand} & \textbf{Representative Evaluation Signal} \\
\midrule
Abrupt Shift & Detect the change quickly; suppress pre-shift evidence before it dominates post-shift decisions. & Detection delay and changepoint quality \citep{adams2007changepoint,chen2019nonstationarycontextual}; restart or adaptive-window behavior \citep{auer2019adaptively}; post-shift AUC in continual ICRL \citep{wang2025cicrl}. \\
\addlinespace[2pt]
Gradual Drift & Track a moving target; smooth noisy evidence without locking onto stale estimates. & Bandit and MDP dynamic regret \citep{besbes2014nonstationarybandit,cheung2020nonstationary,fei2020dynamic}; refined tracking bounds \citep{feng2023nonstationary,wei2021nonstationary,mao2021nearoptimal}; kernelized and sliding-window variants \citep{domingues2021kernel,garivier2011ucb,russac2019weighted,trovo2020sliding}. \\
\addlinespace[2pt]
Recurring Modes & Preserve mode-specific memory; retrieve old context when a regime returns without causing interference. & Re-adaptation latency and old-mode reuse \citep{steinparz2022reactive}; retrieval-augmented ICRL quality \citep{schmied2024radt,sridhar2024regent}; retrieval-based RL memory quality \citep{goyal2022retrievalrl,humphreys2022largescale}. \\
\addlinespace[2pt]
Expanding Support & Recognize tasks, observations, or actions that were absent from early context. & Out-of-support return and calibration under novelty \citep{xu2024psbl,wang2025anymdp}. \\
\addlinespace[2pt]
Seasonal or Periodic Change & Learn when old context becomes useful again rather than merely old. & Performance by cycle phase; retrieval precision across cycles \citep{khetarpal2022continual,abel2023definition,wolczyk2022disentangling}. \\
\addlinespace[2pt]
Adversarial Or Worst-Case Shift & Avoid being exploited by stale, strategically timed, or poisoned context. & Worst-case regret, robust return, and stale-context stress tests \citep{moeini2025safeicrl,chen2025dynamicregret}. \\
\bottomrule
\end{tabularx}
\endgroup
\end{table}

These temporal patterns imply different forms of memory.
Abrupt non-stationarity favors short-term evidence and explicit forgetting.
Gradual drift favors smoothing and recency-weighted estimation.
Recurring modes favor long-term memory with mode-specific retrieval.
Expanding support favors exploration and uncertainty calibration.
A single context-length ablation cannot distinguish these cases.

\subsection{Shift Observability}

The third axis asks how directly the agent can observe that a shift has occurred.
At one extreme, the change is announced by a task descriptor, instruction, goal token, or other explicit signal.
At the other, it must be inferred from prediction errors, transition mismatches, or delayed outcomes that may appear only after consequential actions have already been taken.
This distinction is central for ICRL because the same context must often serve two roles at once: diagnosing the current regime and choosing actions under that diagnosis.
Table~\ref{tab:observability} separates the resulting observability regimes and the abilities they stress.

\begin{table}[htbp]
\caption{Taxonomy axis 3: observability regimes.}
\label{tab:observability}
\centering
\begingroup
\SurveyTableSetup
\begin{tabularx}{\linewidth}{@{}
  B{0.2\linewidth}
  L{0.3\linewidth}Y@{}}
\toprule
\textbf{Regime} & \textbf{Shift Evidence} & \textbf{ICRL-Specific Failure Mode} \\
\midrule
Explicitly Signaled & Task ID; instruction; goal token; system message. & The model may ignore the signal as irrelevant metadata, or overfit to the signal without using interaction evidence \citep{xu2022prompting,yao2023react}. \\
\addlinespace[2pt]
Partially Observable & Rewards, transitions, or observations reveal the change only after several interactions. & Adaptation speed depends on belief-state memory \citep{kaelbling1998pomdp,heess2015memory,ni2022recurrentpomdp}, algorithm-distillation or supervised ICRL context use \citep{laskin2023algorithm,lee2023supervised}, and exploration scaffolds \citep{dai2024icee,krishnamurthy2024explore}. \\
\addlinespace[2pt]
Latent And Confounded & The same observations remain compatible with multiple shift explanations. & Context must preserve competing hypotheses in contextual and hidden-parameter MDPs \citep{hallak2015contextualmdp,killian2017hipmdp,doshivelez2013hipmdp}, epistemic or task-inference POMDPs \citep{ghosh2021epistemic,humplik2019taskinference,ren2020ocean}, latent-belief meta-RL \citep{rakelly2019pearl,zintgraf2020varibad}, and value-aware uncertainty models \citep{berkes2026spice}. \\
\addlinespace[2pt]
Delayed Feedback & Shift evidence arrives only after a long horizon or after irreversible early choices. & Early post-shift actions can be dominated by stale pre-shift context in long-context meta-RL \citep{grigsby2023amago,grigsby2024amago2}, S4-style ICRL \citep{lu2023s4icrl}, and long-term memory settings \citep{pasukonis2022longterm}. \\
\addlinespace[2pt]
Noisy Or Corrupted Signal & Change indicators are unreliable, missing, or adversarially perturbed. & The context manager must estimate trustworthiness, not simply recency \citep{zisman2024noise,chen2025filtering}. \\
\bottomrule
\end{tabularx}
\endgroup
\end{table}

Observability also changes what a benchmark can legitimately claim to measure.
When every shift is announced by a clean descriptor, success mostly demonstrates task conditioning, instruction use, or prompt parsing.
When shifts are hidden behind sparse delayed feedback, the same label instead covers exploration, belief updating, and long-horizon credit assignment.
Both regimes are valuable, but they should not be treated as interchangeable evidence for a single adaptation capability.

\subsection{A Compact Naming Scheme}

For reporting experiments and organizing the literature, it is useful to name a non-stationary setting with a compact three-part notation:
\[
  \begingroup
  \setlength{\fboxsep}{2.2pt}
  \text{\fcolorbox{nsblue}{nslight}{\textcolor{nsblue}{\normalfont\scshape source}}}
  \;\textcolor{nsgray}{\boldsymbol{/}}\;
  \text{\fcolorbox{nsgreen}{nslight}{\textcolor{nsgreen}{\normalfont\scshape temporal pattern}}}
  \;\textcolor{nsgray}{\boldsymbol{/}}\;
  \text{\fcolorbox{nsorange}{nslight}{\textcolor{nsorange}{\normalfont\scshape observability}}}
  \endgroup .
\]
For example, an unannounced reward switch is `reward/abrupt/partial', a recurring hidden dynamics mode is `dynamics/recurring/latent', and an action remapping announced by an interface token is `action/abrupt/explicit'.
The notation is not a substitute for a full benchmark description.
Its purpose is more modest: it makes the adaptation claim legible before the details of the environment, dataset, or model architecture are introduced.

\subsection{Using the Taxonomy Analytically}

The taxonomy is intended to discipline comparisons, not merely to label benchmarks.
Success on `reward/abrupt/explicit' tasks may reflect effective use of a goal token, whereas success on `dynamics/gradual/partial' tasks requires tracking transition evidence over time.
Performance on `task/recurring/latent' tasks may come from memory indexing and mode reuse; performance on `behavior-policy/abrupt/noisy' tasks instead depends on judging whether demonstrations and retrieved trajectories are trustworthy.
These are distinct capabilities, so they should support distinct claims.

The same taxonomy suggests a minimal stress-test rule.
For each claimed source of non-stationarity, evaluation should include at least one temporal pattern where stale context is harmful and at least one observability setting where trivial task-ID conditioning is unavailable.
For each claimed memory mechanism, evaluation should include one case where old context should be discarded and one where old context should be recovered.
This two-sided requirement prevents degenerate solutions from looking general: a recent-only policy can excel on one-way drift, and an append-all policy can excel on stationary or recurring tasks, but neither by itself solves context validity.

\clearpage
\section{The Role of Context Under Non-Stationarity}
\label{sec:context-validity}

\subsection{Context Is Evidence, Memory, and Intervention}

In stationary ICRL, context is usually treated as evidence about a fixed latent task.
Relevant examples improve inference; irrelevant ones are expected to be ignored.
Non-stationarity breaks that tidy picture.
The context is now a time-stamped record of decisions, rewards, and observations that may have been generated under different regimes.
It is also an intervention: the retrieved or summarized context changes what the frozen model attends to, and therefore changes the policy it implements.
This is why external-memory work in language agents is relevant as a systems analogy even when it is not itself reinforcement learning: MemoryBank \citep{zhong2023memorybank}, LongMem \citep{wang2023longmem}, and MemGPT \citep{packer2023memgpt} make persistent memory explicit.
Retrieval-augmented language models \citep{lewis2020rag,guu2020realm,borgeaud2022retro}, nearest-neighbor language models \citep{khandelwal2019knnlm}, and in-context retrieval \citep{ram2023incontextretrieval} make retrieval and evidence selection explicit.
The same evidence-management problem appears in older sequential decision theory under different names: POMDP belief states \citep{kaelbling1998pomdp}, contextual MDP latent contexts \citep{hallak2015contextualmdp}, hidden-parameter MDP embeddings \citep{killian2017hipmdp,doshivelez2013hipmdp}, task-inference variables \citep{humplik2019taskinference,ren2020ocean}, and Bayesian changepoint posteriors \citep{adams2007changepoint} all encode the idea that recent observations should update beliefs about the current regime.
Non-stationary ICRL differs in its engineering constraint: the belief update may be only an implicit computation inside a fixed sequence model, retrieval rule, or prompt summary.

Context therefore plays three roles at once.
It is evidence about rewards, dynamics, goals, preferences, and uncertainty.
It is memory, preserving information that may become useful again if a mode recurs.
It is also a control input to a frozen policy, because changing the context can change behavior without changing weights.
Non-stationary ICRL is hard because these roles do not always agree.
The best evidence for the current mode may require discarding long-term memory; the best long-term memory may exceed the context budget; the most salient trajectory may be precisely the one that should no longer be trusted.
Thus the core question is not whether the agent remembers history, but whether it can infer when history has stopped supporting the current policy.
In a finite-budget system, this tradeoff can be written as a context-selection problem:
\begin{equation}
  \ctx_t^\star
  =
  \arg\max_{\ctx \subseteq \mathcal{M}_{t-1},\ |\ctx|\leq b_t}
  \mathbb{E}\!\left[V_t^{\policy}(s_t \mid \ctx)\right]
  -
  \lambda\,\mathrm{cost}(\ctx),
  \label{eq:context-selection}
\end{equation}
where $\mathcal{M}_{t-1}$ is the memory available before acting, $b_t$ is the context budget, and the cost term can represent tokens, retrieval calls, latency, or risk of stale evidence.
Most existing systems approximate this objective implicitly; non-stationary evaluation should make the approximation visible.

\subsection{Failure Modes}

Table~\ref{tab:failure-modes} summarizes common failure modes that are easy to miss in stationary evaluation.

\begin{table}[!htbp]
\caption{Failure modes induced by stale or invalid context.}
\label{tab:failure-modes}
\centering
\begingroup
\SurveyTableSetup
\begin{tabularx}{\linewidth}{@{}
  B{0.15\linewidth}
  L{0.22\linewidth}
  L{0.2\linewidth}
  Y@{}}
\toprule
\textbf{Failure Mode} & \textbf{Mechanism} & \textbf{Symptom} & \textbf{Diagnostic Ablation} \\
\midrule
Stale Imitation & The model imitates high-return pre-shift trajectories after the reward or dynamics changed. & Strong performance before shift, slow or no recovery afterward. & Insert old optimal demonstrations into a post-shift prompt for decision-transformer \citep{chen2021decision}, algorithm-distillation \citep{laskin2023algorithm}, or suboptimal-ICRL \citep{wang2026suboptimalicrl} settings. \\
Context Inertia & Long context dominates new evidence. & New rewards are observed but behavior changes late. & Vary pre-shift context length while holding post-shift evidence fixed in AMAGO-style agents \citep{grigsby2023amago,grigsby2024amago2}, S4-style ICRL \citep{lu2023s4icrl}, or RELIC-style long-context agents \citep{elawady2024relic}. \\
Mode Collapse & The agent maps several latent modes to one familiar behavior. & Recurring or mixed modes are handled as if stationary. & Evaluate mode-conditioned returns and confusion matrix for latent-belief meta-RL methods such as PEARL \citep{rakelly2019pearl} and VariBAD \citep{zintgraf2020varibad}. \\
Retrieval Contamination & External memory retrieves superficially similar but invalid trajectories. & Retrieval helps in stationary tasks but hurts after interface or reward shift. & Compare similarity-only retrieval with validity-filtered retrieval in RA-DT \citep{schmied2024radt} and REGENT-style systems \citep{sridhar2024regent}. \\
Compression Loss & Summaries discard the change evidence needed for adaptation. & Compressed context performs well on average but misses rare shifts. & Preserve raw post-shift events while compressing old context in memory systems \citep{zhong2023memorybank,wang2023longmem,packer2023memgpt}, communication-based compression \citep{martinezlopez2025coral}, or filtered-history training \citep{chen2025filtering}. \\
Action-Token Mismatch & The model assumes old action semantics still hold. & Correct high-level plan but wrong low-level actions. & Permute, mask, or rescale actions at known change points \citep{chandak2020changingaction,sinii2024variable}. \\
Over-Forgetting & The model discards old context that becomes useful again. & Slow recovery when a previous mode returns. & Use recurring-mode benchmarks and measure re-adaptation latency \citep{khetarpal2022continual,steinparz2022reactive,schmied2024radt}. \\
\bottomrule
\end{tabularx}
\endgroup
\end{table}

Several of these failures come from context management rather than from the policy class itself.
A sufficiently expressive model can still fail when its prompt is dominated by invalid evidence, when retrieval keys optimize surface similarity, or when compression removes a rare but decisive reward event.
Conversely, a modest model can look robust if it receives clean, recent, mode-specific evidence.
The policy and the context pipeline therefore have to be evaluated together.

The failure modes also explain why longer context is an ambiguous intervention.
Increasing the context window can improve adaptation by making more evidence available, but it can also increase the fraction of invalid evidence, make attention more diffuse, and amplify high-salience pre-shift trajectories.
The right comparison is therefore not ``short context versus long context'' in isolation.
It is short context, full history, recent-only history, oracle-truncated history, learned retrieval, and adversarially contaminated retrieval under the same shift schedule.
Only this comparison reveals whether scale is solving adaptation or merely hiding the cost of invalid context.
This point is consistent with the broader ICL literature: training distributions \citep{chan2022datadistribution}, transformer function learning \citep{garg2022transformers}, gradient-free or probabilistic ICL mechanisms \citep{kirsch2022gpicl,wies2023learnability}, demonstration ordering \citep{min2022rethinking}, and contextual learning dynamics \citep{akyurek2024contextlanguage} can encourage in-context inference, memorization, or failure depending on how examples are clustered, repeated, ordered, or distributed across tasks.
For ICRL, the analogous distributional feature is not only task diversity but whether context windows contain the right conflicts: pre-shift versus post-shift evidence, high-return stale trajectories versus low-return current evidence, and recurring modes separated by misleading intervening data.

Retrieval and long context make the problem sharper rather than eliminating it.
Retrieval-augmented language systems \citep{izacard2022atlas}, retrieval-based RL systems \citep{goyal2022retrievalrl,humphreys2022largescale}, and retrieval-augmented ICRL systems \citep{schmied2024radt,sridhar2024regent} show that external memories can improve sample efficiency and reuse, but they also introduce a new decision variable: which memory should be treated as evidence for the current regime.
Long-context studies similarly warn that simply appending more examples can dilute or distort in-context learning when the relevant evidence is far from the query or mixed with distractors \citep{li2024longcontext,bertsch2024longcontext}.
For non-stationary ICRL, this means retrieval precision and attention allocation should be reported as adaptation mechanisms, not only as engineering choices.

\subsection{The Context Lifecycle}

A useful system-level view is to give context a lifecycle:
\begin{enumerate}
  \item \textbf{Write:} which interactions, demonstrations, and feedback are stored?
  \item \textbf{Index:} how are stored items keyed for future retrieval?
  \item \textbf{Retrieve:} which items are selected for the current decision?
  \item \textbf{Compress:} how are long histories summarized within a finite context budget?
  \item \textbf{Trust or weight:} after evidence is available to the policy, how much influence should each item have on action selection?
  \item \textbf{Forget or isolate:} when should context be removed, downweighted, or stored in a mode-specific partition?
\end{enumerate}
These entries should not be read as a mandatory serial architecture.
Write, index, retrieve, compress, and forget or isolate are often discrete context-manager operations: a system can log an item, create a key, retrieve a subset, summarize a history, or move an item out of the active memory.
Trust is different.
It denotes the \emph{influence assignment} given to available evidence once it reaches the decision computation.
In an explicit system, trust may be implemented by a reliability score, uncertainty gate, value-fusion weight, safety filter, or regime posterior.
In a plain transformer or recurrent policy, it may be an emergent property of attention, hidden-state dynamics, value prediction, and decoding rather than a separately exposed module.
The lifecycle is therefore an analytic decomposition of responsibilities, not a claim that every ICRL agent contains six separate processing blocks.

\begin{figure}[htbp]
\centering
\includegraphics[width=0.666\linewidth]{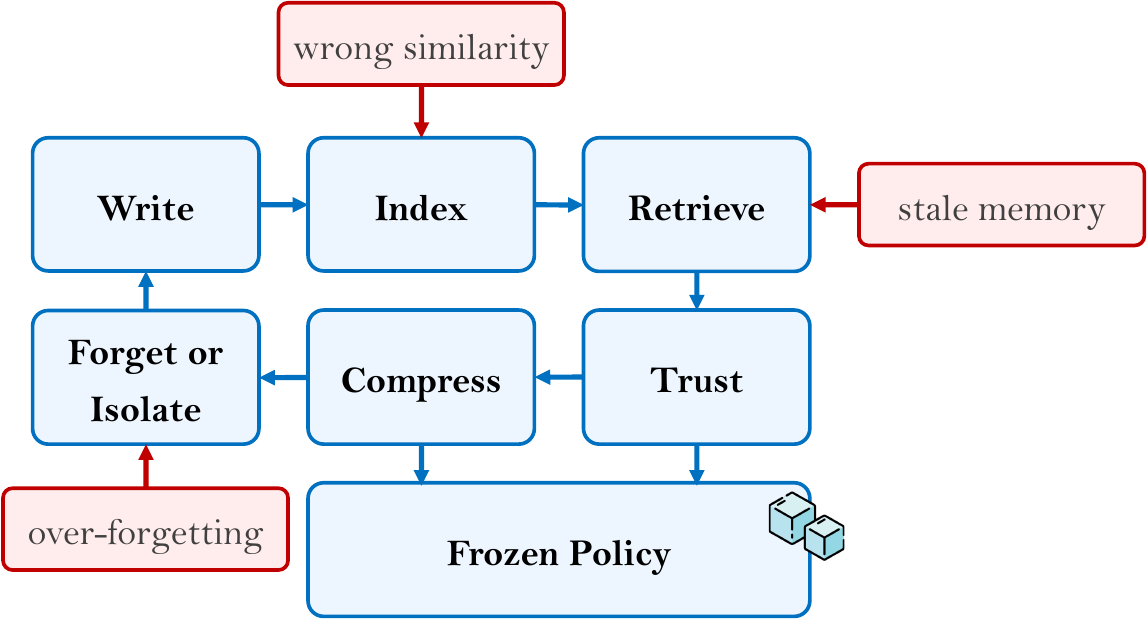}
\caption{Context lifecycle. Under non-stationarity, every operation can improve adaptation or inject invalid evidence.}
\label{fig:context-lifecycle}
\end{figure}

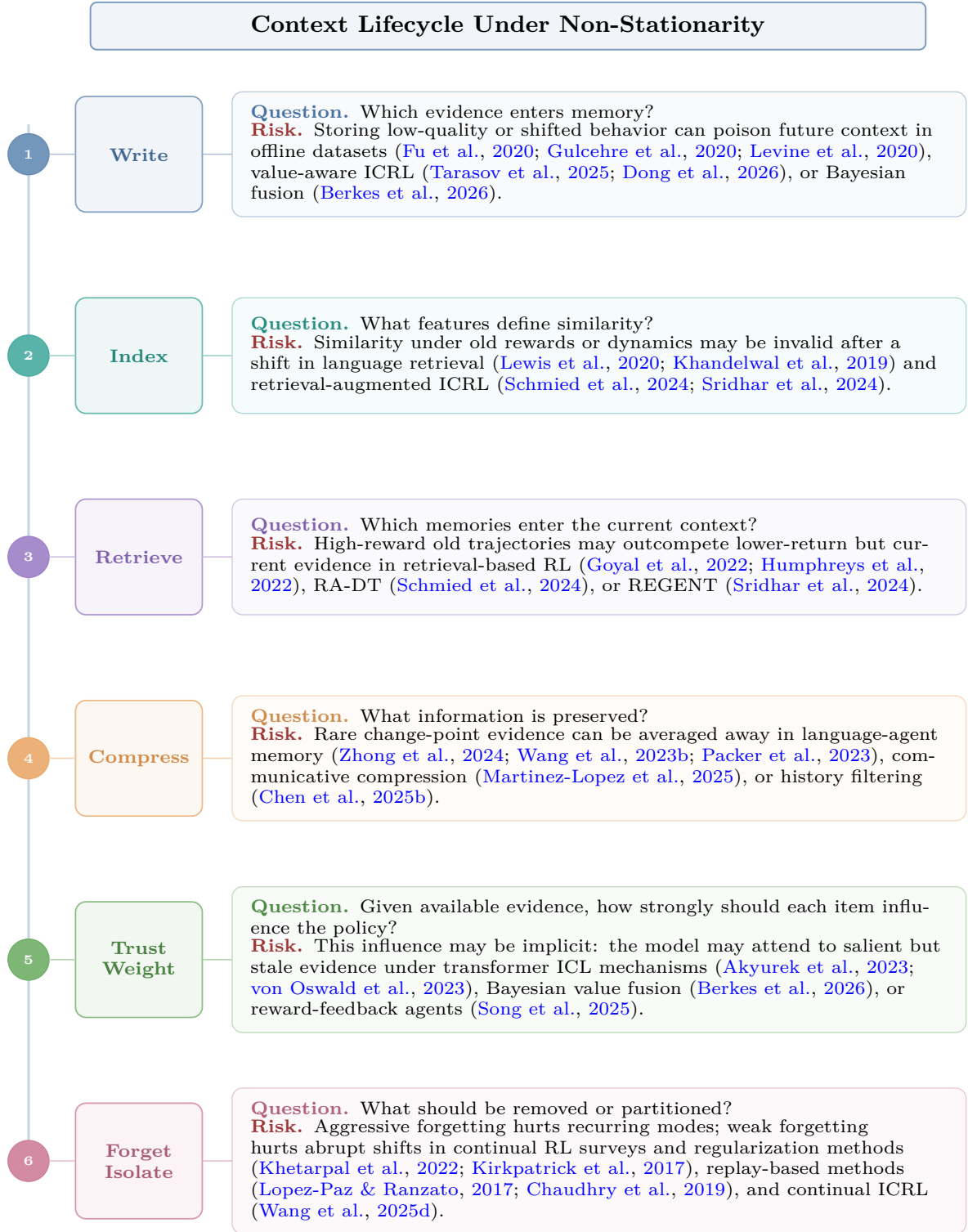
\begin{figure}[htpb]
\centering
\begingroup
\resizebox{\linewidth}{!}{\begingroup
\definecolor{lifeBlue}{HTML}{4C78A8}
\definecolor{lifeTeal}{HTML}{2A9D8F}
\definecolor{lifePurple}{HTML}{8E6BBE}
\definecolor{lifeOrange}{HTML}{E59B52}
\definecolor{lifeGreen}{HTML}{59A14F}
\definecolor{lifeRose}{HTML}{C76A8A}

\begin{tikzpicture}[
  x=1cm,
  y=1cm,
  every node/.style={outer sep=0pt},
  titlebox/.style={
    draw=lifeBlue!75,
    fill=lifeBlue!8,
    rounded corners=2.5pt,
    line width=0.6pt,
    align=center,
    font=\bfseries\small,
    inner xsep=8pt,
    inner ysep=5pt
  },
  dot/.style={
    circle,
    text=white,
    line width=0.45pt,
    minimum size=5.5mm,
    font=\bfseries\tiny,
    align=center
  },
  opbox/.style={
    rounded corners=4pt,
    line width=0.55pt,
    minimum width=1.70cm,
    minimum height=1.55cm,
    align=center,
    font=\bfseries\scriptsize,
    inner sep=3pt
  },
  contentbox/.style={
    rounded corners=4pt,
    line width=0.4pt,
    text width=9.35cm, 
    minimum height=1.55cm,
    align=left,
    font=\scriptsize,
    inner xsep=7pt,
    inner ysep=5pt
  },
  spine/.style={
    draw=lifeBlue!30,
    line width=1.0pt
  }
]

\def\xdot{0.82}
\def\xop{2.30}      
\def\xcontent{3.55} 

\node[titlebox, anchor=west, minimum width=11.20cm] at (1.65,1.62)
  {Context Lifecycle Under Non-Stationarity};

\draw[spine] (\xdot,0.30) -- (\xdot,-13.60);

\def\yone{-0.10}
\def\ytwo{-2.80}
\def\ythree{-5.50}
\def\yfour{-8.20}
\def\yfive{-10.90}
\def\ysix{-13.60}

\node[dot, draw=lifeBlue!85, fill=lifeBlue!78] (d1) at (\xdot,\yone) {1};
\node[opbox, draw=lifeBlue!70, fill=lifeBlue!8, text=lifeBlue!85!black] (op1) at (\xop,\yone) {Write};
\node[contentbox, draw=lifeBlue!40, fill=lifeBlue!4, anchor=west] (c1) at (\xcontent,\yone) {
  \textcolor{lifeBlue!95!black}{\textbf{Question.}} Which evidence enters memory?\\[-1pt]
  \textcolor{nsred}{\textbf{Risk.}} Storing low-quality or shifted behavior can poison future context in offline datasets
  \citep{fu2020d4rl,gulcehre2020rlunplugged,levine2020offline}, value-aware ICRL
  \citep{tarasov2025qlearning,wang2026suboptimalicrl}, or Bayesian fusion \citep{berkes2026spice}.
};
\draw[draw=lifeBlue!45, line width=0.5pt] (d1.east) -- (op1.west);
\draw[draw=lifeBlue!45, line width=0.5pt] (op1.east) -- (c1.west);

\node[dot, draw=lifeTeal!85, fill=lifeTeal!78] (d2) at (\xdot,\ytwo) {2};
\node[opbox, draw=lifeTeal!70, fill=lifeTeal!8, text=lifeTeal!85!black] (op2) at (\xop,\ytwo) {Index};
\node[contentbox, draw=lifeTeal!40, fill=lifeTeal!4, anchor=west] (c2) at (\xcontent,\ytwo) {
  \textcolor{lifeTeal!90!black}{\textbf{Question.}} What features define similarity?\\[-1pt]
  \textcolor{nsred}{\textbf{Risk.}} Similarity under old rewards or dynamics may be invalid after a shift in language retrieval
  \citep{lewis2020rag,khandelwal2019knnlm} and retrieval-augmented ICRL
  \citep{schmied2024radt,sridhar2024regent}.
};
\draw[draw=lifeTeal!45, line width=0.5pt] (d2.east) -- (op2.west);
\draw[draw=lifeTeal!45, line width=0.5pt] (op2.east) -- (c2.west);

\node[dot, draw=lifePurple!85, fill=lifePurple!78] (d3) at (\xdot,\ythree) {3};
\node[opbox, draw=lifePurple!70, fill=lifePurple!8, text=lifePurple!90!black] (op3) at (\xop,\ythree) {Retrieve};
\node[contentbox, draw=lifePurple!40, fill=lifePurple!4, anchor=west] (c3) at (\xcontent,\ythree) {
  \textcolor{lifePurple!90!black}{\textbf{Question.}} Which memories enter the current context?\\[-1pt]
  \textcolor{nsred}{\textbf{Risk.}} High-reward old trajectories may outcompete lower-return but current evidence in retrieval-based RL
  \citep{goyal2022retrievalrl,humphreys2022largescale}, RA-DT \citep{schmied2024radt}, or REGENT
  \citep{sridhar2024regent}.
};
\draw[draw=lifePurple!45, line width=0.5pt] (d3.east) -- (op3.west);
\draw[draw=lifePurple!45, line width=0.5pt] (op3.east) -- (c3.west);

\node[dot, draw=lifeOrange!85, fill=lifeOrange!78] (d4) at (\xdot,\yfour) {4};
\node[opbox, draw=lifeOrange!70, fill=lifeOrange!8, text=lifeOrange!90!black] (op4) at (\xop,\yfour) {Compress};
\node[contentbox, draw=lifeOrange!40, fill=lifeOrange!4, anchor=west] (c4) at (\xcontent,\yfour) {
  \textcolor{lifeOrange!90!black}{\textbf{Question.}} What information is preserved?\\[-1pt]
  \textcolor{nsred}{\textbf{Risk.}} Rare change-point evidence can be averaged away in language-agent memory
  \citep{zhong2023memorybank,wang2023longmem,packer2023memgpt}, communicative compression
  \citep{martinezlopez2025coral}, or history filtering \citep{chen2025filtering}.
};
\draw[draw=lifeOrange!45, line width=0.5pt] (d4.east) -- (op4.west);
\draw[draw=lifeOrange!45, line width=0.5pt] (op4.east) -- (c4.west);

\node[dot, draw=lifeGreen!85, fill=lifeGreen!78] (d5) at (\xdot,\yfive) {5};
\node[opbox, draw=lifeGreen!70, fill=lifeGreen!8, text=lifeGreen!80!black] (op5) at (\xop,\yfive) {Trust\\Weight};
\node[contentbox, draw=lifeGreen!40, fill=lifeGreen!4, anchor=west] (c5) at (\xcontent,\yfive) {
  \textcolor{lifeGreen!85!black}{\textbf{Question.}} Given available evidence, how strongly should each item influence the policy?\\[-1pt]
  \textcolor{nsred}{\textbf{Risk.}} This influence may be implicit: the model may attend to salient but stale evidence under transformer ICL mechanisms
  \citep{akyurek2023whatlearning,vonoswald2023gradient}, Bayesian value fusion \citep{berkes2026spice}, or
  reward-feedback agents \citep{song2026reward}.
};
\draw[draw=lifeGreen!45, line width=0.5pt] (d5.east) -- (op5.west);
\draw[draw=lifeGreen!45, line width=0.5pt] (op5.east) -- (c5.west);

\node[dot, draw=lifeRose!85, fill=lifeRose!78] (d6) at (\xdot,\ysix) {6};
\node[opbox, draw=lifeRose!70, fill=lifeRose!8, text=lifeRose!85!black] (op6) at (\xop,\ysix) {Forget\\Isolate};
\node[contentbox, draw=lifeRose!40, fill=lifeRose!4, anchor=west] (c6) at (\xcontent,\ysix) {
  \textcolor{lifeRose!90!black}{\textbf{Question.}} What should be removed or partitioned?\\[-1pt]
  \textcolor{nsred}{\textbf{Risk.}} Aggressive forgetting hurts recurring modes; weak forgetting hurts abrupt shifts in continual RL surveys
  and regularization methods \citep{khetarpal2022continual,kirkpatrick2017overcoming}, replay-based methods
  \citep{lopezpaz2017gem,chaudhry2019agem}, and continual ICRL \citep{wang2025cicrl}.
};
\draw[draw=lifeRose!45, line width=0.5pt] (d6.east) -- (op6.west);
\draw[draw=lifeRose!45, line width=0.5pt] (op6.east) -- (c6.west);

\end{tikzpicture}
\endgroup}
\endgroup
\caption{Context lifecycle operations and their non-stationary design questions. The layout follows a taxonomy-style view: each lifecycle operation is treated as a stage whose design decision determines how stale or shifted evidence can affect the policy.}
\label{fig:context-lifecycle-taxonomy}
\end{figure}

This distinction clarifies the relation between retrieve and trust.
Retrieval answers a candidate-selection question: which memories or summaries enter the context window at all?
Trust answers an influence question: conditional on evidence being present, how strongly does it shape the action distribution?
A retrieval-augmented agent can retrieve the right item and still fail by over-weighting a stale high-return trajectory; conversely, a long-context agent with no explicit retriever can still succeed if attention or value inference suppresses invalid evidence.
For non-stationary ICRL, both questions should be reported when possible: selection quality through retrieval or context-ablation metrics, and influence quality through attention, gating, counterfactual prompting, value sensitivity, or stale-evidence stress tests.
Under non-stationarity, every operation or influence facet in this lifecycle can become either an adaptation mechanism or a failure point.
Figure~\ref{fig:context-lifecycle} visualizes this lifecycle, and Figure~\ref{fig:context-lifecycle-taxonomy} states the design question and non-stationary risk at each operation or facet.

\subsection{Implication for Survey Organization}

The rest of the survey therefore organizes methods by how they manage context.
Algorithm distillation learns from histories that contain improvement \citep{laskin2023algorithm}.
Supervised ICRL learns mappings from contextual examples to decisions \citep{lee2023supervised,lin2023transformers,huang2024icdt}.
Retrieval-augmented methods select external memories \citep{schmied2024radt,sridhar2024regent,goyal2022retrievalrl}.
Value-aware and Bayesian methods combine context with priors \citep{tarasov2025qlearning,liu2026sicql,berkes2026spice}.
Model-based methods infer dynamics or planning computations in context \citep{son2025dicp,xu2026icprl}.
World-model and communication-based methods compress experience \citep{zhong2023memorybank,wang2023longmem,packer2023memgpt,martinezlopez2025coral}.
Reflection-style agents use textual or symbolic summaries of feedback \citep{shinn2023reflexion,zhao2024expel,wang2023voyager,song2026reward}.
For non-stationarity, the key question for each family is the same: what makes the context valid, and what happens when it is not?

This lifecycle view suggests a decomposition of responsibility.
The write operation controls what evidence can ever be used.
The index and retrieve operations control which evidence is considered similar or relevant enough to expose.
The compress operation controls which rare events survive a finite budget.
The trust facet controls how strongly available evidence affects the frozen policy, even when that weighting is implicit in the model rather than implemented as a separate module.
The forget or isolate operation controls whether old evidence disappears, is downweighted, or remains available for future recurrence.
A non-stationary ICRL paper need not solve all parts of the lifecycle, but it should state which responsibility its contribution improves, which parts are held fixed, and whether trust is explicitly modeled or only inferred from behavior.

\clearpage
\section{Method Families Through the Lens of Context Management}
\label{sec:methods}

Methods for ICRL differ less by their names than by where they place the burden of adaptation.
Some rely on the sequence model to infer the right update from raw history.
Others move part of the problem into retrieval, summarization, memory partitioning, value estimation, or model inference.
For non-stationary settings, the useful organizing question is simple: when the world changes, what mechanism decides which context still matters?

\begin{figure}[htpb]
\centering
\resizebox{\linewidth}{!}{

\begingroup
\newcommand{\taxcite}[1]{{\scriptsize\citep{#1}}}

\definecolor{nsgray}{RGB}{90,95,100}
\definecolor{nsred}{RGB}{210,80,80}
\definecolor{nslight}{RGB}{247,249,251}

\definecolor{methodblue}{RGB}{48,112,180}
\definecolor{methodgreen}{RGB}{45,145,96}
\definecolor{methodpurple}{RGB}{125,96,180}
\definecolor{methodorange}{RGB}{196,123,54}
\definecolor{methodteal}{RGB}{38,145,150}

\begin{tikzpicture}[
  x=1cm,
  y=1cm,
  every node/.style={font=\small},
  axis/.style={
    -{Latex[length=3mm,width=2.2mm]},
    line width=0.9pt,
    draw=nsgray!85
  },
  quadrant/.style 2 args={
    rounded corners=8pt,
    draw=#1!34,
    fill=#1!4,
    line width=0.6pt,
    inner sep=0pt
  },
  qtitle/.style 2 args={
    rounded corners=3pt,
    fill=#1!13,
    draw=#1!42,
    line width=0.4pt,
    text=#1!85!black,
    font=\scriptsize\bfseries,
    align=center,
    inner xsep=6pt,
    inner ysep=3pt
  },
  item/.style 2 args={
    draw=#1!70,
    fill=white,
    rounded corners=4pt,
    line width=0.6pt,
    text width=#2,
    minimum height=0.95cm,
    align=left,
    inner xsep=10pt,
    inner ysep=6pt,
    font=\small,
    text=nsgray,
    path picture={
      \fill[#1!86]
        (path picture bounding box.north west)
        rectangle ([xshift=3.8pt]path picture bounding box.south west);
    }
  },
  gapitem/.style={
    draw=nsred!72,
    fill=nsred!6,
    rounded corners=4pt,
    line width=0.7pt,
    text width=6.2cm,
    minimum height=0.95cm,
    align=left,
    inner xsep=10pt,
    inner ysep=6pt,
    font=\small,
    text=nsgray,
    path picture={
      \fill[nsred!72]
        (path picture bounding box.north west)
        rectangle ([xshift=3.8pt]path picture bounding box.south west);
    }
  },
  axislab/.style={
    font=\large\bfseries,
    text=nsgray,
    align=center
  },
  ticklab/.style={
    font=\large\bfseries,
    text=nsgray!90,
    align=center
  }
]

\node[quadrant={methodblue}{}, minimum width=7.25cm, minimum height=4.2cm, anchor=south west]
  (qbl) at (0.8,0.8) {};
\node[quadrant={methodgreen}{}, minimum width=7.25cm, minimum height=4.2cm, anchor=south west]
  (qbr) at (8.35,0.8) {};
\node[quadrant={methodpurple}{}, minimum width=7.25cm, minimum height=4.2cm, anchor=south west]
  (qtl) at (0.8,5.4) {};
\node[quadrant={methodorange}{}, minimum width=7.25cm, minimum height=4.2cm, anchor=south west]
  (qtr) at (8.35,5.4) {};

\node[qtitle={methodpurple}{}, anchor=west] at (1.2,9.6)
  {implicit history + value/model evidence};

\node[qtitle={methodorange}{}, anchor=west] at (8.75,9.6)
  {managed memory + value/model evidence};

\node[qtitle={methodblue}{}, anchor=west] at (1.2,5.0)
  {implicit history + behavioral evidence};

\node[qtitle={methodgreen}{}, anchor=west] at (8.75,5.0)
  {managed memory + behavioral evidence};

\draw[axis] (0.2,0.3) -- (15.9,0.3);

\node[axislab, anchor=west] at (16.1,0.3)
  {explicit context\\selection};

\draw[axis] (0.2,0.3) -- (0.2,10.2);

\node[axislab, anchor=south] at (0.2,10.4)
  {value/model\\awareness};

\draw[draw=nsgray!40, densely dotted, line width=0.6pt]
  (8.2,0.6) -- (8.2,9.8);

\draw[draw=nsgray!40, densely dotted, line width=0.6pt]
  (0.6,5.2) -- (15.8,5.2);

\node[ticklab, anchor=north] at (4.4,0.0) {implicit history};
\node[ticklab, anchor=north] at (12.0,0.0) {managed memory};

\node[ticklab, rotate=90] at (-0.4,2.9) {behavioral evidence};
\node[ticklab, rotate=90] at (-0.4,7.5) {value/model evidence};

\node[item={methodblue}{6.2cm}, anchor=west] at (1.2,4.0)
  {\textbf{Decision Transformers}: trajectory tokens
  \taxcite{chen2021decision,janner2021trajectory}};

\node[item={methodblue}{6.2cm}, anchor=west] at (1.2,2.7)
  {\textbf{Algorithm Distillation}: learning histories
  \taxcite{laskin2023algorithm}};

\node[item={methodteal}{6.2cm}, anchor=west] at (1.2,1.4)
  {\textbf{Long-context meta-RL}: extended histories
  \taxcite{grigsby2024amago2,elawady2024relic}};

\node[item={methodgreen}{6.2cm}, anchor=west] at (8.75,3.6)
  {\textbf{Retrieval-augmented ICRL}: external memory
  \taxcite{schmied2024radt,sridhar2024regent}};

\node[item={methodgreen}{6.2cm}, anchor=west] at (8.75,2.0)
  {\textbf{Filtering / compression}: budgeted histories
  \taxcite{chen2025filtering,martinezlopez2025coral}};

\node[item={methodpurple}{6.2cm}, anchor=west] at (1.2,8.6)
  {\textbf{Model-based ICRL}: dynamics inference
  \taxcite{son2025dicp,xu2026icprl}};

\node[item={methodpurple}{6.2cm}, anchor=west] at (1.2,7.3)
  {\textbf{Value-aware ICRL}: value-sensitive context
  \taxcite{tarasov2025qlearning,berkes2026spice}};

\node[item={methodpurple}{6.2cm}, anchor=west] at (1.2,6.0)
  {\textbf{Reward-feedback agents}: preference evidence
  \taxcite{song2026reward}};

\node[gapitem, anchor=west] at (8.75,8.1)
  {\textbf{Open gap}: validity-aware retrieval and forgetting
  \taxcite{schmied2024radt,wang2025cicrl}};

\node[item={methodorange}{6.2cm}, anchor=west] at (8.75,6.5)
  {\textbf{World-model memory}: selected model evidence
  \taxcite{son2025dicp,xu2026icprl}};

\end{tikzpicture}
\endgroup}
\caption{Method landscape organized by context-selection explicitness and value/model awareness.}
\label{fig:method-landscape}
\end{figure}
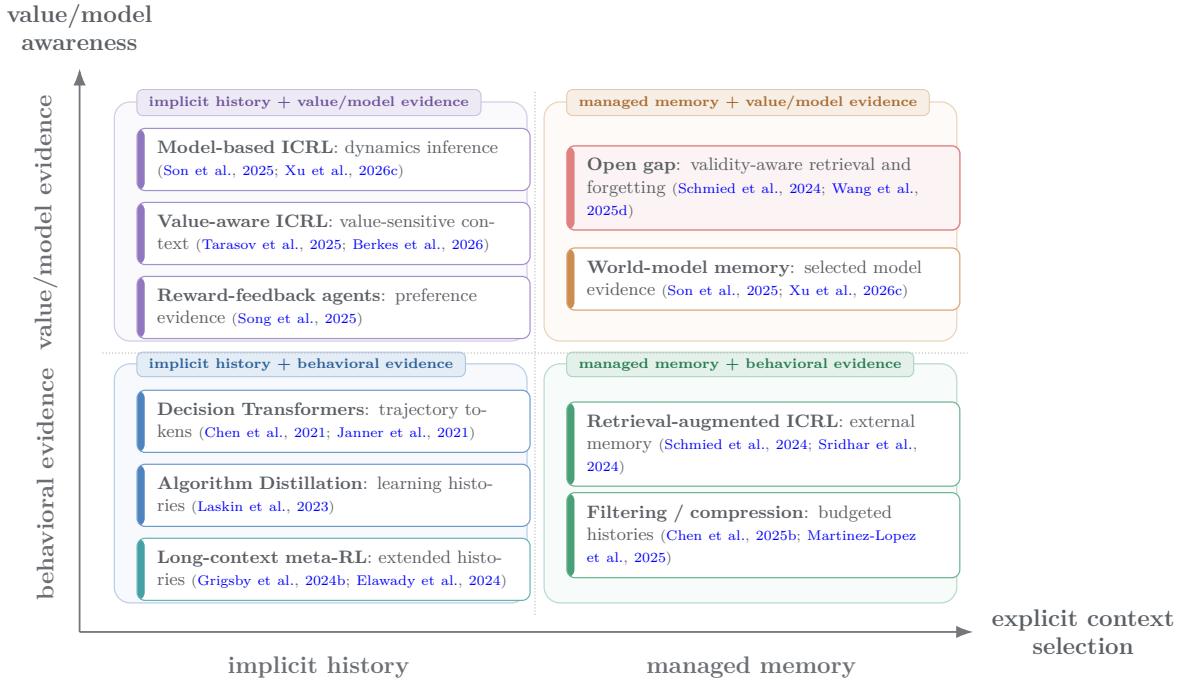

Figure~\ref{fig:method-landscape} summarizes the resulting landscape.
The horizontal axis separates implicit history use from explicit context selection through retrieval, summaries, and memory policies.
The vertical axis separates raw behavioral evidence from value- or model-aware evidence.
The empty region is as informative as the occupied one: many systems scale context length or memory access without estimating validity, while value- and model-aware systems still depend on the evidence selected for them.

\subsection{Algorithm Distillation and Learning Histories}

Algorithm distillation trains a sequence model on histories generated by a learning algorithm, with the aim that the trained model internalizes the algorithm's within-lifetime improvement \citep{laskin2023algorithm}.
For non-stationary ICRL, this family is important because learning histories can include exploration, failure, reward feedback, and policy improvement.
If the pretraining histories include changes in task or reward, the model may learn a pattern of adaptation through context.
If they are stationary, the model may instead learn that older high-return behavior should remain trusted.
This distinction mirrors broader ICL evidence that models can shift between memorization, system identification, and genuine in-context learning depending on distributional burstiness \citep{chan2022datadistribution}, probabilistic or gradient-free ICL structure \citep{kirsch2022gpicl,wies2023learnability}, example ordering \citep{min2022rethinking}, and contextual learning dynamics \citep{akyurek2024contextlanguage}.

The critical design variable is the distribution of histories.
Histories generated by a strong learner on stationary tasks provide clean examples of improvement, but they may not teach change detection.
Histories generated under weak data or noisy policies may better reflect deployment, but they can also create ambiguous supervision.
Cross-episodic curricula \citep{shi2023cec}, noise distillation \citep{zisman2024noise}, contrastive context encoders \citep{wang2021contrastivecontext}, and filtered learning histories \citep{chen2025filtering} all modify the history distribution to make within-context improvement easier to learn.
From the non-stationary perspective, filtering should not merely remove low-return episodes; it should preserve transitions that reveal when old context became invalid.
ICPRL adds a physically grounded variant of this idea by training a vision-language policy on multi-episode trial-and-error histories, so that later actions can condition on previous physical failures and successes without test-time weight updates \citep{xu2026icprl}.
The adjacent meta-exploration literature is useful here because it explicitly treats early actions as information-gathering, not only as return maximization, through exploration objectives \citep{stadie2018exploration,dorfman2020offlineexploration,kamienny2020adaptiveexploration}, decoupled exploration \citep{liu2021decoupling}, hyperstate inference \citep{zintgraf2021hyperstate}, or first-explore strategies \citep{norman2023firstexplore}.

\subsection{Supervised ICRL and Decision-Pretrained Transformers}

Supervised pretraining can induce ICRL when the model is trained on datasets where context examples identify a latent decision problem \citep{lee2023supervised,lin2023transformers}.
Decision Transformer \citep{chen2021decision}, Trajectory Transformer \citep{janner2021trajectory}, RvS \citep{emmons2022rvs}, prompting variants \citep{xu2022prompting}, generalized decision transformers \citep{furuta2022gdt}, chain-of-hindsight or agentic variants \citep{liu2023agentic}, and ICDT-style models \citep{huang2024icdt} provide the broader sequence-modeling substrate.
These methods are attractive because they can use offline data and scale with standard supervised objectives.
Recent scaling and routing variants, including multi-game decision transformers \citep{lee2022multigame}, online decision transformers \citep{zheng2022online}, hierarchical prompt transformers \citep{wang2024hierarchicalprompt}, Vintix-style action-model formulations \citep{polubarov2025vintix}, and mixture-of-experts formulations \citep{wu2025t2mir}, further test whether the same substrate can support broader task mixtures.
Theoretical results on in-context policy improvement clarify why this family is more than sequence imitation: under structured conditions, a transformer can use trajectory context to implement a policy-improvement procedure rather than simply replaying the behavior policy \citep{liang2026policyimprovement}.
This gives supervised and decision-pretrained ICRL a principled bridge to classical policy iteration, but it also exposes the non-stationary missing assumption.
Policy improvement is well grounded when the trajectories describe the same decision process; after reward, dynamics, or action-semantics shifts, the model must first infer which trajectories are still valid before any improvement step is meaningful.
Their weakness under non-stationarity is objective mismatch.
Next-action prediction can reward copying behavior from context even when that behavior is no longer optimal.
Return conditioning can encode a desired outcome, but it does not by itself solve which past examples are relevant after a shift.

The non-stationary version of supervised ICRL should therefore include contrastive context designs.
For example, prompts can contain pre-shift and post-shift trajectories with identical observations but different rewards.
Training can require the model to attend to the most recent valid evidence, infer a latent mode, or predict change-point indicators.
Without such designs, strong held-out-task performance may reflect static generalization rather than adaptation to changing environments.
Demonstration-conditioned sequential decision studies show that task diversity, dataset size, and trajectory structure matter for in-context generalization, but those results still need explicit stale-context and shift tests before they establish non-stationary adaptation \citep{raparthy2023generalization}.
The weak-label and suboptimal-data variants are particularly important for this survey.
Offline ICRL rarely has perfect learning histories or optimal labels at scale; offline RL coverage issues \citep{levine2020offline}, conservative and implicit value learning \citep{kumar2020cql,kostrikov2022iql}, return-conditioned value learning \citep{brandfonbrener2022returnconditioned}, Q-learning-style ICRL \citep{tarasov2025qlearning}, suboptimal-data ICRL \citep{wang2026suboptimalicrl}, random-policy state-action distillation \citep{chen2024randompolicyicrl}, and TD-style ICRL \citep{wang2024tdicrl} can turn imperfect historical trajectories into useful adaptation signals, but only if the context window exposes which behavior was suboptimal under the current regime.

\subsection{Efficient Sequence Backbones}

The default ICRL backbone is a causal transformer, but long non-stationary lifetimes make the quadratic cost of attention a practical bottleneck.
Structured state-space models, recurrent-memory transformers, gated recurrent attention, and Mamba-style sequence models therefore deserve explicit treatment.
S4-style ICRL \citep{lu2023s4icrl} and Decision S4 \citep{david2023decisions4} show that state-space sequence models can support long-horizon memory and sequence-based RL, while Decision Mamba \citep{ota2024decisionmamba,huang2024decisionmamba} and Mamba-based scaling of algorithm distillation \citep{beaussant2025mambaad} test selective state-space decision models.
Related selective-state, recurrent-memory, and xLSTM-style models \citep{gu2023mamba,smith2023s5,bulatov2022recurrentmemory} and transformer-RL memory studies \citep{beck2024xlstm,ni2024transformersrl} test whether recurrent state can replace or complement transformer attention in decision modeling.
For non-stationarity, the architectural question is not only speed.
A compressed recurrent state can make old information cheap to carry, but it can also make invalid information hard to remove.
Thus efficient backbones should be evaluated with the same stale-context, oracle-reset, and recurring-mode tests as transformer systems.
Recent non-stationary ICL theory \citep{qin2026beyondstationarity} makes this architectural point explicit: adaptive filtering \citep{sayed2011adaptive}, gating \citep{yang2023gla,katsch2023gateloop}, retention \citep{sun2023retnet}, and recurrent attention \citep{peng2024eagle} can be interpreted as learnable forgetting factors when the target function changes over time.

\subsection{Long-Context Meta-RL Agents}

Long-context agents such as AdA-style systems \citep{bauer2023humantimescale}, AMAGO-style systems \citep{grigsby2023amago,grigsby2024amago2}, RELIC-style systems \citep{elawady2024relic}, transformer meta-RL systems \citep{melo2022transformersmetarl}, and hierarchical meta-RL agents \citep{shala2024hierarchicalmetarl} use recurrent, transformer, or external-memory mechanisms to solve adaptive RL problems across long horizons and open-ended task spaces.
They are natural candidates for non-stationary ICRL because they can integrate long histories and delayed feedback.
However, long context is not automatically better.
When the environment changes, a longer history increases both evidence and contamination.
The relevant question becomes whether the model has learned attention patterns that separate current-mode evidence from stale-mode evidence.
Large context, open-ended learning \citep{stooke2021openended}, dataset efforts such as XLand and XLand-100B \citep{nikulin2025xland}, broader generalist-agent sequence models \citep{reed2022gato,polubarov2025vintix,polubarov2026vintixii}, and episodic or long-term recall systems \citep{ritter2018episodicrecall,pasukonis2022longterm} make this question sharper because they scale learning histories far beyond toy settings while still requiring tests of whether scaled histories teach invalidation rather than only broad task recognition.

Long-context meta-RL also raises a reporting issue.
If a model is trained online with RL across non-stationary episodes, then test-time adaptation may be partly due to learned recurrent dynamics rather than explicit context retrieval or prompting.
This is still within our ICRL boundary when parameters are fixed at deployment, but papers should report the lifetime state, context window, reset rules, and whether memory persists across episodes.

\subsection{Retrieval-Augmented ICRL}

Retrieval-augmented methods externalize part of the context problem.
Instead of fitting the entire history into the model's input, they retrieve trajectories, transitions, or summaries from memory in RA-DT \citep{schmied2024radt}, REGENT \citep{sridhar2024regent}, in-context Q-learning with retrieval \citep{xu2026icql}, and earlier retrieval-based RL systems \citep{goyal2022retrievalrl,humphreys2022largescale}.
This is one of the most promising families for non-stationarity because retrieval can, in principle, select mode-relevant memories and ignore stale ones.
It is also one of the most fragile.
If retrieval is based on surface similarity, it may select trajectories from the wrong reward regime.
If retrieval is based on return, it may prefer high-return old behavior that is now invalid.
If retrieval is based on recentness, it may fail when an old mode recurs.

For non-stationary ICRL, retrieval should be evaluated with validity metrics, not only downstream return.
Precision of retrieved context, counterfactual performance without retrieval, and adversarial stale-memory tests are needed.
The best retrieval key may combine observations, actions, rewards, inferred latent modes, uncertainty, and temporal metadata.
This creates a bridge between ICRL and memory systems: retrieval is not just a scaling trick, but a learned or designed adaptation policy.
The language-model retrieval literature is a useful warning because nearest-neighbor language models \citep{khandelwal2019knnlm}, retrieval-augmented generation and pretraining \citep{lewis2020rag,guu2020realm,borgeaud2022retro}, in-context retrieval \citep{ram2023incontextretrieval}, and Atlas-style retrieval \citep{izacard2022atlas} can improve prediction while leaving relevance and trust as separate design problems.

\subsection{Value-Aware and Bayesian Context Use}

Value-aware methods attempt to make ICRL less dependent on imitation of raw trajectories.
Q-learning objectives can improve offline ICRL \citep{tarasov2025qlearning} by teaching the model value-sensitive behavior rather than pure behavior cloning, building on the broader offline RL lesson that conservative and implicit value regularization matter under support mismatch \citep{kumar2020cql,kostrikov2022iql}.
Scalable in-context Q-learning \citep{liu2026sicql} and compositional Q-learning \citep{xu2026icql} extend this idea by conditioning value estimates on compact context or retrieved transitions.
Bayesian-fusion approaches combine context with value priors, aiming to decide how much to trust contextual evidence under uncertainty \citep{berkes2026spice}.

This family is especially relevant under reward and dynamics shifts.
If the model estimates uncertainty over values, it can downweight context that conflicts with the current reward signal.
If it only predicts next actions, it may reproduce old behavior even after feedback changes.
The limitation is that value-aware objectives still need the right evidence.
A Q-function conditioned on invalid retrieved context can still produce invalid actions.
Thus value awareness and context validity are complementary: values help interpret context, while context management determines which evidence values are conditioned on.

\subsection{Action-Space and Interface Adaptation}

Action-space variation deserves separate treatment because it is both a scaling issue and a non-stationarity issue.
Variable-action ICRL shows that models can be designed to generalize across discrete action spaces with different size, order, and semantic content \citep{sinii2024variable}.
For this survey, the non-stationary version is stricter: the action interface may change during a lifetime while old context still contains action tokens from the previous interface.
This makes action embeddings, action descriptions, and interface metadata part of context validity rather than a mere implementation detail.
Benchmarks that only switch rewards or tasks miss this failure mode.

\subsection{Model-Based In-Context Planning}

Model-based ICRL distills dynamics inference \citep{son2025dicp} or planning computation \citep{xu2026icprl} into context-conditioned computation.
Under dynamics shifts, this is a direct fit: the agent must infer how actions change the world from recent evidence and plan accordingly.
Model-based context also gives a clearer diagnostic than action prediction.
If a model predicts transitions correctly after a shift but still acts poorly, the failure may lie in planning.
If it predicts old transitions after observing new ones, the failure is context adaptation.
ICPRL is useful here because it separates an adaptive policy from a world model used for lookahead search in visual physics tasks \citep{xu2026icprl}.
For a non-stationary benchmark, the same architecture would need to show that the world-model-guided search reacts to changed physical parameters or action outcomes, rather than merely exploiting a static simulator prior.
World-model work \citep{hafner2019planet}, simulator-learning work \citep{yang2023simulators}, Genie-style interactive generation \citep{bruce2024genie}, and instructable simulators \citep{raad2024instructable} provide adjacent mechanisms for this direction, but the survey treats them as relevant only when the learned model is used in a repeated decision loop with context-conditioned adaptation.

The challenge is compounding error.
A model-based in-context planner may infer the wrong latent dynamics from stale history and then plan confidently under that wrong model.
Non-stationary evaluation should therefore report both predictive accuracy and control performance across pre-shift, immediately post-shift, and recovered phases.

\subsection{Context Compression and Communicative World Models}

Finite context makes compression unavoidable.
Communication- or world-model-based approaches can summarize interaction histories into compact language-agent memory \citep{zhong2023memorybank,wang2023longmem,packer2023memgpt}, predictive video representations \citep{wu2024ivideogpt}, or communicative world-model representations \citep{martinezlopez2025coral}.
Compression is useful under long horizons and recurring tasks, but it can erase the rare evidence that marks a shift.
The key question is not compression ratio alone, but preservation of decision-relevant change evidence.
For example, a summary that preserves average reward may miss a sudden preference reversal; a summary that preserves the most recent transition may lose an older mode that will recur.

\subsection{LLM and Foundation Agents as Extended ICRL Cases}

LLM-based and foundation-agent systems are in scope only when the model participates in a sequential decision loop where context, actions, feedback, and later behavior are coupled.
Recent work frames LLMs as in-context reinforcement learners under bandit feedback \citep{monea2024bandit}, reward feedback \citep{song2026reward}, or online regret criteria \citep{park2024llmregret}.
In these settings, the frozen model weights are not the adaptation channel.
The adaptation channel is the prompt, transcript, retrieved memory, reflection, executable skill, or summary that changes future actions after previous outcomes.
ReAct-style agents \citep{yao2023react} connect reasoning traces to environment actions, while Reflexion \citep{shinn2023reflexion}, ExpeL \citep{zhao2024expel}, Voyager \citep{wang2023voyager}, and general computer control agents \citep{tan2025cradle} store verbal reinforcement, experience summaries, executable skills, or interaction memories for later trials.

Non-stationarity makes this bridge case more than a systems detail.
User preferences can drift, hidden instructions can change, reward feedback can be reinterpreted, tool schemas and website layouts can move, and old reflections can become stale policy hints.
A memory may be semantically similar to the current situation but invalid because it was produced under a different reward function, tool interface, or user constraint.
Natural-language summaries are especially risky because they compress many observations into declarative advice; after a shift, the advice can remain fluent and authoritative while becoming wrong.
This is the LLM-agent analogue of stale trajectories in decision-transformer systems and stale retrieved transitions in retrieval-augmented ICRL.

The survey still treats these systems as extended cases rather than the core of ICRL.
Many LLM-agent papers do not report scalar rewards, controlled action spaces, lifecycle curves, no-memory baselines, or context ablations.
Success in an interactive language environment may reflect instruction following, prompt parsing, static task recognition, or retrieval of a near duplicate rather than reinforcement learning through context.
Bandit studies provide an important caution: frozen LLMs often need explicit summaries \citep{krishnamurthy2024explore,dai2024icee} or algorithmic scaffolding \citep{nie2025evolve} to explore reliably in context.
Multi-turn agent training methods such as PAPRIKA \citep{tajwar2025curious}, ICAL \citep{sarch2024ical}, ArCHer \citep{zhou2024archer}, VAGEN \citep{wang2025vagen}, SCoRe-style self-correction \citep{kumar2025score}, and cultural-accumulation systems \citep{cook2024generational} are therefore adjacent unless the experiment shows that repeated feedback changes future decisions through the context channel.

For a large-model result to support a non-stationary ICRL claim, the evidence should show post-shift behavioral change without parameter updates, compare memory policies, and expose stale-context failure.
Useful baselines include no memory, last-$k$ transcript, full transcript, summary memory, retrieval memory, oracle-current-rule context, and stale-memory adversaries.
Small-scale reinforcement-tuned transformer studies that report adaptation to changing gridworld conditions are useful proof-of-concept evidence \citep{rentschler2025rltransformer}, but a convincing large-model benchmark must also test whether verbal memories, tool histories, and retrieved examples remain valid under preference, instruction, reward, or interface shifts.
Without environment interaction, scalar or preference feedback, and repeated decisions, LLM memory is better viewed as supervised or conversational in-context learning.

\begin{table}[htbp]
\caption{Large-model bridge cases for non-stationary ICRL.}
\label{tab:foundation-agent-bridge}
\centering
\begingroup
\SurveyTableSetup
\begin{tabularx}{\linewidth}{@{}
  B{0.2\linewidth}
  L{0.4\linewidth}
  Y@{}}
\toprule
\textbf{Category} & \textbf{ICRL Entry Condition} & \textbf{Non-Stationary Risk} \\
\midrule
Decision Foundation Models / Generalist Sequence Agents & Trajectory context changes future actions with fixed parameters in broad decision-model settings \citep{reed2022gato,polubarov2025vintix,polubarov2026vintixii}. & Scale can mask stale-context failure when performance is reported only as broad task success. \\
LLM Reward-Feedback Agents & Feedback, reward, regret, or reflection changes later decisions in bandit or interactive settings \citep{monea2024bandit,song2026reward,park2024llmregret}. & Old verbal feedback or reflection can remain over-trusted after preference or reward shifts. \\
Tool-Use / Computer-Control Agents & Tool outcomes and action histories shape later tool calls or environment actions \citep{yao2023react,tan2025cradle}. & Tool APIs, website layouts, action schemas, or parser conventions can drift. \\
VLM / World-Model Agents & Visual observations and transition evidence update planning through context \citep{xu2026icprl,martinezlopez2025coral}. & Old physical, visual, or causal evidence can mislead after dynamics or observation shifts. \\
Long-Memory Agents & Retrieved memories, summaries, or executable skills influence later decisions \citep{zhong2023memorybank,wang2023longmem,packer2023memgpt,wang2023voyager}. & Similar memories may be invalid under the current regime, especially when summaries hide when evidence was collected. \\
\bottomrule
\end{tabularx}
\endgroup
\end{table}

\subsection{Method Matrix}

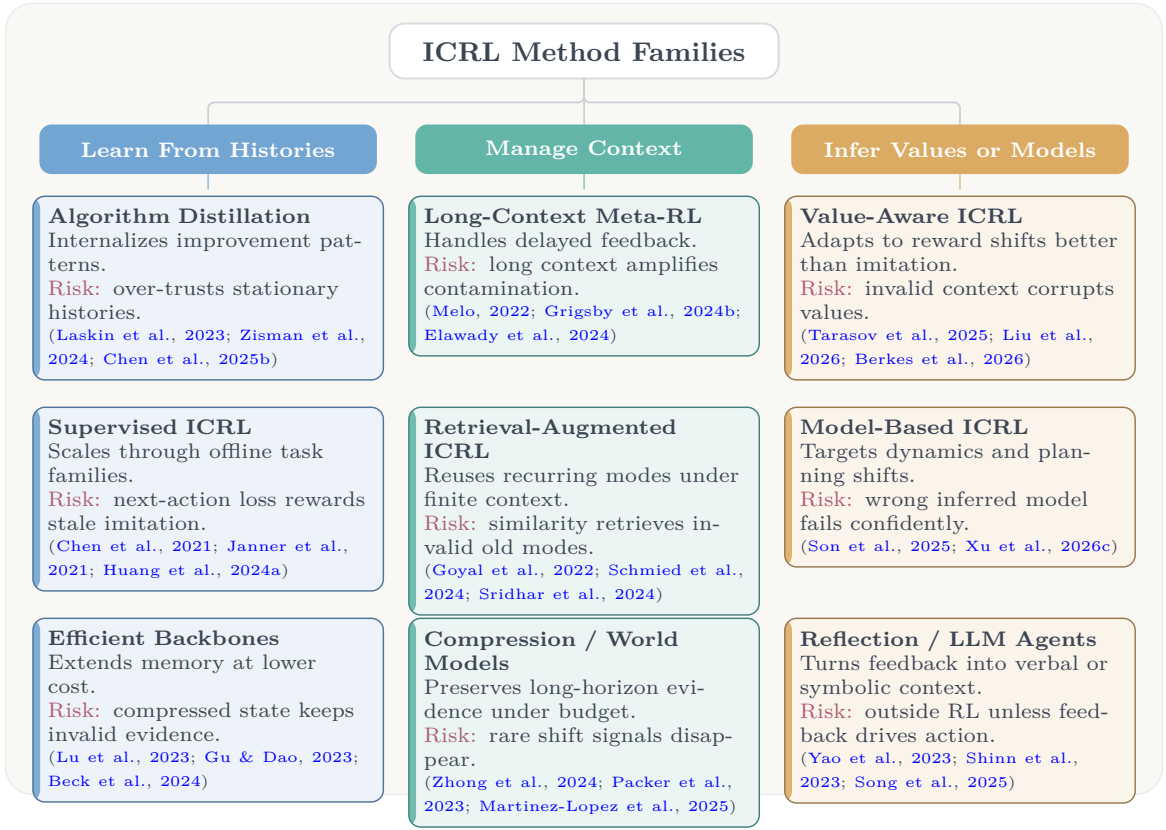
\begin{figure}[htbp]
\centering
\begingroup
\resizebox{\linewidth}{!}{\begingroup
\newcommand{\methodcite}[1]{{\tiny\citep{#1}}}

\definecolor{methodBg}{RGB}{250,248,244}
\definecolor{methodInk}{RGB}{60,69,84}
\definecolor{methodLine}{RGB}{174,181,192}

\definecolor{methodBlue}{RGB}{92,150,203}
\definecolor{methodTeal}{RGB}{72,169,153}
\definecolor{methodGold}{RGB}{214,157,74}
\definecolor{methodRose}{RGB}{194,105,128}

\begin{tikzpicture}[
  x=1cm,
  y=1cm,
  every node/.style={font=\scriptsize, outer sep=0pt},
  root/.style={
    draw=methodLine!50,
    rounded corners=5pt,
    line width=0.65pt,
    align=center,
    fill=white,
    inner xsep=11pt,
    inner ysep=6pt,
    font=\small\bfseries,
    text=methodInk
  },
  branch/.style 2 args={
    rounded corners=4pt,
    minimum width=3.95cm,
    minimum height=0.58cm,
    align=center,
    font=\scriptsize\bfseries,
    fill=#1!86,
    text=white,
    inner sep=2pt
  },
  leaf/.style 2 args={
    draw=#1!58!methodInk,
    rounded corners=4pt,
    line width=0.48pt,
    fill=#1!11,
    text width=3.75cm,
    minimum height=1.72cm,
    align=left,
    inner xsep=5pt,
    inner ysep=4pt,
    font=\scriptsize,
    text=methodInk,
    path picture={
      \fill[#1!78]
        (path picture bounding box.north west)
        rectangle ([xshift=2.4pt]path picture bounding box.south west);
    }
  },
  stem/.style={
    line width=0.58pt,
    draw=methodLine!64,
    rounded corners=2pt
  },
  tick/.style={
    line width=0.58pt
  }
]

\fill[rounded corners=9pt, methodBg] (-0.82,-8.58) rectangle (12.72,0.68);
\draw[rounded corners=9pt, draw=methodLine!28, line width=0.4pt]
  (-0.82,-8.58) rectangle (12.72,0.68);

\node[root] (root) at (5.95,0.12) {ICRL Method Families};

\coordinate (learnTop) at (1.55,-1.03);
\coordinate (manageTop) at (5.95,-1.03);
\coordinate (modelTop) at (10.35,-1.03);

\draw[stem] (root.south) -- +(0,-0.28) -| (learnTop);
\draw[stem] (root.south) -- +(0,-0.28) -| (manageTop);
\draw[stem] (root.south) -- +(0,-0.28) -| (modelTop);

\node[branch={methodBlue}{}] (learn) at (1.55,-1.03) {Learn From Histories};
\node[branch={methodTeal}{}] (manage) at (5.95,-1.03) {Manage Context};
\node[branch={methodGold}{}] (model) at (10.35,-1.03) {Infer Values or Models};

\node[leaf={methodBlue}{}, anchor=north] (l1) at (1.55,-1.58)
  {\textbf{Algorithm Distillation}\\Internalizes improvement patterns.\\\textcolor{methodRose!90!black}{Risk:} over-trusts stationary histories.\\[-1pt]\methodcite{laskin2023algorithm,zisman2024noise,chen2025filtering}};
\node[leaf={methodBlue}{}, anchor=north] (l2) at (1.55,-4.05)
  {\textbf{Supervised ICRL}\\Scales through offline task families.\\\textcolor{methodRose!90!black}{Risk:} next-action loss rewards stale imitation.\\[-1pt]\methodcite{chen2021decision,janner2021trajectory,huang2024icdt}};
\node[leaf={methodBlue}{}, anchor=north] (l3) at (1.55,-6.52)
  {\textbf{Efficient Backbones}\\Extends memory at lower cost.\\\textcolor{methodRose!90!black}{Risk:} compressed state keeps invalid evidence.\\[-1pt]\methodcite{lu2023s4icrl,gu2023mamba,beck2024xlstm}};

\node[leaf={methodTeal}{}, anchor=north] (m1) at (5.95,-1.58)
  {\textbf{Long-Context Meta-RL}\\Handles delayed feedback.\\\textcolor{methodRose!90!black}{Risk:} long context amplifies contamination.\\[-1pt]\methodcite{melo2022transformersmetarl,grigsby2024amago2,elawady2024relic}};
\node[leaf={methodTeal}{}, anchor=north] (m2) at (5.95,-4.05)
  {\textbf{Retrieval-Augmented ICRL}\\Reuses recurring modes under finite context.\\\textcolor{methodRose!90!black}{Risk:} similarity retrieves invalid old modes.\\[-1pt]\methodcite{goyal2022retrievalrl,schmied2024radt,sridhar2024regent}};
\node[leaf={methodTeal}{}, anchor=north] (m3) at (5.95,-6.52)
  {\textbf{Compression / World Models}\\Preserves long-horizon evidence under budget.\\\textcolor{methodRose!90!black}{Risk:} rare shift signals disappear.\\[-1pt]\methodcite{zhong2023memorybank,packer2023memgpt,martinezlopez2025coral}};

\node[leaf={methodGold}{}, anchor=north] (v1) at (10.35,-1.58)
  {\textbf{Value-Aware ICRL}\\Adapts to reward shifts better than imitation.\\\textcolor{methodRose!90!black}{Risk:} invalid context corrupts values.\\[-1pt]\methodcite{tarasov2025qlearning,liu2026sicql,berkes2026spice}};
\node[leaf={methodGold}{}, anchor=north] (v2) at (10.35,-4.05)
  {\textbf{Model-Based ICRL}\\Targets dynamics and planning shifts.\\\textcolor{methodRose!90!black}{Risk:} wrong inferred model fails confidently.\\[-1pt]\methodcite{son2025dicp,xu2026icprl}};
\node[leaf={methodGold}{}, anchor=north] (v3) at (10.35,-6.52)
  {\textbf{Reflection / LLM Agents}\\Turns feedback into verbal or symbolic context.\\\textcolor{methodRose!90!black}{Risk:} outside RL unless feedback drives action.\\[-1pt]\methodcite{yao2023react,shinn2023reflexion,song2026reward}};

\draw[tick, draw=methodBlue!58] (learn.south) -- ++(0,-0.18);
\draw[tick, draw=methodTeal!58] (manage.south) -- ++(0,-0.18);
\draw[tick, draw=methodGold!58] (model.south) -- ++(0,-0.18);

\end{tikzpicture}
\endgroup}
\endgroup
\caption{Method families organized by context-management mechanism. Each leaf states the adaptation advantage and the main non-stationary failure mode.}
\label{fig:method-matrix-taxonomy}
\end{figure}

Figure~\ref{fig:method-matrix-taxonomy} condenses the preceding method discussion by mapping each family to its context mechanism, non-stationary strength, and main risk.

\subsection{Cross-Family Lessons}

Across these families, architecture is not the cleanest dividing line.
Algorithm distillation and supervised ICRL ask the pretrained model to infer the meaning of context inside its forward pass.
Retrieval and compression make the context pipeline more explicit: the model can adapt only to evidence that was selected and preserved.
Value-aware and model-based methods change what the evidence means, treating context as information about values, dynamics, or uncertainty rather than as behavior to imitate.
These choices are complementary under non-stationarity, but each needs different evidence.
Table~\ref{tab:method-requirements} turns this point into a standard for evaluating claims about each method family.

The common pattern is that useful systems need three things at once.
Their training histories must contain invalidation events, otherwise old evidence is always presented as trustworthy.
Their context managers need validity cues--time, shift likelihood, reward compatibility, uncertainty, or latent-mode metadata--instead of relying only on similarity.
Their evaluations must test both forgetting and reuse.
Aggressive forgetting can look good after an abrupt shift and fail when a mode recurs; preserving everything can look good on recurrence and fail after an irreversible preference change.
The target is not more memory in the abstract, but controlled influence of memory.

\begin{table}[htbp]
\caption{Required evidence for method families under non-stationarity.}
\label{tab:method-requirements}
\centering
\begingroup
\SurveyTableSetup
\begin{tabularx}{\linewidth}{@{}
    B{0.18\linewidth}
    L{0.57\linewidth}
    Y@{}}
\toprule
\textbf{Family} & \textbf{Necessary Demonstration} & \textbf{Weak Evidence That Is Often Insufficient} \\
\midrule
Algorithm Distillation & Learns to change behavior after context invalidation in AD \citep{laskin2023algorithm}, contrastive context encoders \citep{wang2021contrastivecontext}, or continual ICRL \citep{wang2025cicrl}, not only after ordinary trial-and-error. & Improvement on stationary held-out tasks. \\
Supervised ICRL & Uses conflicting prompt evidence according to current rewards or dynamics in supervised ICRL \citep{lee2023supervised,lin2023transformers}, TD-style ICRL \citep{wang2024tdicrl}, or suboptimal-data ICRL \citep{wang2026suboptimalicrl}. & Higher return with more demonstrations. \\
Efficient Sequence Backbones & Show that retained hidden state or compressed sequence state can be reset, gated, or corrected after shifts in S4-style models \citep{lu2023s4icrl,david2023decisions4}, Mamba-style models \citep{ota2024decisionmamba,gu2023mamba}, or gated attention \citep{yang2023gla}. & Faster inference or longer context on stationary tasks. \\
Long-Context Meta-RL & Separates useful delayed evidence from stale early evidence in AMAGO-style systems \citep{grigsby2023amago,grigsby2024amago2} and long-term memory settings \citep{pasukonis2022longterm}. & Better performance with longer memory. \\
Retrieval-Augmented ICRL & Retrieves valid memories when similar invalid memories exist in retrieval-based RL \citep{goyal2022retrievalrl,humphreys2022largescale}, RA-DT \citep{schmied2024radt}, or REGENT \citep{sridhar2024regent}. & Retrieval beats no-retrieval on stationary tasks. \\
Value-Aware ICRL & Updates value estimates when reward semantics shift in Q-learning ICRL \citep{tarasov2025qlearning}, scalable in-context Q-learning \citep{liu2026sicql}, or Bayesian value fusion \citep{berkes2026spice}. & Better average offline return. \\
Model-Based ICRL & Revises inferred dynamics after transition changes in DICP \citep{son2025dicp} or ICPRL \citep{xu2026icprl}. & Accurate prediction before the shift. \\
Compression & Preserves rare change evidence under budget in communicative compression \citep{martinezlopez2025coral} or filtered histories \citep{chen2025filtering}. & High compression ratio with small average loss. \\
\bottomrule
\end{tabularx}
\endgroup
\end{table}

\clearpage
\section{Training Data and Objectives}
\label{sec:training}

\subsection{Training Data as an Implicit Non-Stationarity Assumption}

Every ICRL training pipeline encodes an assumption about how context should be interpreted.
Stationary trajectories teach that old evidence remains relevant.
Learning histories teach that behavior can improve across time.
Mixed-task datasets teach that context identifies a latent task.
Retrieval databases teach that similar past experience can help the present.
Non-stationary training data must teach something more specific: the validity of context can change.
This point is easy to miss if ICRL is treated only as offline RL at scale: offline RL datasets \citep{fu2020d4rl,gulcehre2020rlunplugged,agarwal2020optimistic}, offline RL surveys \citep{levine2020offline,prudencio2022offlinesurvey}, and offline sequence-modeling substrates \citep{chen2021decision,janner2021trajectory,emmons2022rvs} provide the data substrate, but they do not automatically create within-context invalidation events.

This suggests a simple audit question for training data:
\begin{takeawayquote}
  Does the dataset contain examples in which previously useful context becomes invalid, and does the supervision reward the model for changing its behavior accordingly?
\end{takeawayquote}
If the answer is no, then robust non-stationary adaptation is unlikely to emerge reliably from scale alone.
The model may still generalize, but the benchmark should not interpret this as evidence of context-validity reasoning.

This audit question should be applied at the level of individual context windows, not only at the dataset level.
A corpus may contain many tasks and still present each window as internally coherent.
Such a corpus teaches task inference but not conflict resolution.
To train non-stationary ICRL, windows should sometimes contain pre-shift and post-shift evidence together, with supervision or rewards that make the distinction consequential.
The model must experience cases where the correct action is not the action most often seen, not the action seen in the highest-return old trajectory, and not necessarily the action associated with the nearest retrieved neighbor.
General ICL studies make the same lesson visible in a simpler setting: example arrangement \citep{min2022rethinking}, task diversity and burstiness \citep{chan2022datadistribution}, transformer function learning \citep{garg2022transformers,akyurek2023whatlearning}, gradient-free or probabilistic ICL mechanisms \citep{kirsch2022gpicl,wies2023learnability}, and contextual learning dynamics \citep{akyurek2024contextlanguage} can determine whether a transformer learns from context or memorizes a distributional shortcut.
For ICRL, dataset construction must additionally expose rewards, interventions, and control consequences.

\subsection{Stationary Pretraining, Non-Stationary Testing}

A common pattern is to pretrain on stationary tasks and test on related but changing tasks.
This creates a useful stress test, but it should be interpreted carefully.
Failure may indicate that the model never learned change detection.
Success may indicate that the non-stationary benchmark is solvable through recency heuristics or static generalization.
To separate these cases, evaluations should include context-order permutations, stale-context injections, repeated-mode tests, and ablations that remove post-shift reward evidence.

\subsection{Non-Stationary Pretraining}

Direct non-stationary pretraining can vary reward functions, dynamics, observations, action mappings, and task distributions across the lifetime.
Work on lifelong ICRL \citep{xu2024psbl} and continual ICRL \citep{wang2025cicrl} moves in this direction.
The main design choice is whether changes are random, curriculum-based, or adversarial.
Randomized worlds \citep{wang2025anymdp}, contextual MDP families \citep{hallak2015contextualmdp}, hidden-parameter MDPs \citep{killian2017hipmdp,doshivelez2013hipmdp}, open-ended or human-timescale task generators \citep{stooke2021openended,bauer2023humantimescale}, unsupervised environment design \citep{dennis2020ued,jiang2021plr,jiang2021replayguided}, PLR-style or evolving generators \citep{parkerholder2022evolving}, and non-stationary MDP benchmark suites \citep{keplinger2025nsgym} can provide scale, but the generator must expose the shift types the survey cares about.
Otherwise, a large randomized dataset can still underrepresent action-semantics shifts, hidden preference changes, or recurring modes.
Recent curriculum-design work makes this point concrete: generators can prioritize levels near the competence frontier \citep{rutherford2024noregrets}, improve regret approximations for task discovery \citep{garcin2024dred}, regularize environment design \citep{frans2023powderworld}, or evolve the mechanics of the game itself \citep{earle2024autoverse}.

\subsection{Objective Alignment}

The training objective should align with the desired adaptation behavior.
Behavior cloning rewards reproducing the data policy.
Return-conditioned sequence modeling rewards action sequences associated with high returns in Decision Transformer \citep{chen2021decision}, Trajectory Transformer \citep{janner2021trajectory}, and RvS \citep{emmons2022rvs}.
Q-learning-style objectives reward value-sensitive decisions in ICRL \citep{tarasov2025qlearning,liu2026sicql,wang2026suboptimalicrl} and TD-style variants \citep{wang2024tdicrl}.
They can inherit useful lessons from conservative \citep{kumar2020cql}, optimistic \citep{agarwal2020optimistic}, implicit \citep{kostrikov2022iql}, or return-conditioned offline value learning \citep{brandfonbrener2022returnconditioned}.
Model-based objectives reward predictive structure in DICP \citep{son2025dicp} and ICPRL \citep{xu2026icprl}.
Contrastive or auxiliary objectives can reward change-point identification, latent-mode inference, or retrieval validity.
Table~\ref{tab:training} summarizes how common training choices shape expected behavior under non-stationarity.

\begin{table}[H]
\caption{Training choices and their expected non-stationary behavior.}
\label{tab:training}
\centering
\begingroup
\SurveyTableSetup
\begin{tabularx}{\linewidth}{@{}
  B{0.16\linewidth}
  L{0.45\linewidth}
  L{0.15\linewidth}Y@{}}
\toprule
\textbf{Training Choice} & \textbf{What It Encourages} & \textbf{Non-Stationary Benefit} & \textbf{Non-Stationary Risk} \\
\midrule
Stationary Expert Histories & Imitation of good behavior in trajectory models \citep{chen2021decision,janner2021trajectory,emmons2022rvs} and prompting variants \citep{xu2022prompting}. & Strong in familiar stable tasks. & Over-trusts old demonstrations after shifts. \\
Learning Histories & Improvement from feedback over time in algorithm distillation \citep{laskin2023algorithm}, cross-episodic curricula \citep{shi2023cec}, noise distillation \citep{zisman2024noise}, RELIC-style agents \citep{elawady2024relic}, and ICPRL \citep{xu2026icprl}. & Teaches adaptation trajectories. & May not teach invalidation unless histories shift. \\
Weak Or Mixed-Quality Data & Robustness to imperfect behavior policies in offline datasets \citep{fu2020d4rl,gulcehre2020rlunplugged,levine2020offline}, offline value learning \citep{kumar2020cql,kostrikov2022iql}, random-policy distillation \citep{chen2024randompolicyicrl}, and value-aware ICRL \citep{tarasov2025qlearning,wang2026suboptimalicrl,berkes2026spice}. & Forces value or feedback use. & Can confuse low quality with non-stationarity. \\
Procedural Task Generation & Broad task support and scalable diversity from open-ended environments \citep{stooke2021openended,bauer2023humantimescale}, unsupervised environment design \citep{dennis2020ued,jiang2021plr,jiang2021replayguided}, evolving or replay-guided generators \citep{parkerholder2022evolving}, randomized MDPs \citep{wang2025anymdp}, XLand-MiniGrid \citep{nikulin2024xlandminigrid}, and Kinetix \citep{matthews2024kinetix}. & Can cover many latent modes. & Coverage depends entirely on generator design. \\
Retrieval Database Training & External memory use in RA-DT \citep{schmied2024radt} and REGENT \citep{sridhar2024regent}. & Enables recurring-mode reuse. & Retrieval shortcuts can mask lack of adaptation. \\
Value-Aware Objectives & Action selection through inferred values in Q-learning ICRL \citep{tarasov2025qlearning}, scalable in-context Q-learning \citep{liu2026sicql}, and compositional in-context Q-learning \citep{xu2026icql}. & Helps reward and preference shifts. & Still brittle if context selection is invalid. \\
Change-Point Auxiliaries & Explicit shift detection through Bayesian changepoints \citep{adams2007changepoint}, adaptive restarts \citep{auer2019adaptively}, non-stationary contextual bandits \citep{chen2019nonstationarycontextual}, continual ICRL \citep{wang2025cicrl}, or dynamic-regret objectives \citep{chen2025dynamicregret}. & Improves recovery and interpretability. & Labels may be unavailable or oversimplified. \\
State-Space Sequence Training & Long-horizon history modeling with S4-style ICRL \citep{lu2023s4icrl,david2023decisions4}, Mamba-style models \citep{ota2024decisionmamba,beaussant2025mambaad,gu2023mamba}, or gated recurrent attention \citep{yang2023gla}. & Makes long lifetimes cheaper to model. & Hidden compression can make invalid evidence harder to audit. \\
\bottomrule
\end{tabularx}
\endgroup
\end{table}

\subsection{Reporting Requirements}

Papers should report the stationarity assumptions of both training and testing.
At minimum, a non-stationary ICRL paper should specify which components can change, whether changes appear in training, whether changes are signaled, whether context persists across changes, and whether the agent has a finite context or retrieval budget.
Without these details, it is difficult to know whether a method has learned non-stationary adaptation or has simply exploited a benchmark convention.

\subsection{Data Construction Patterns}

Several practical data-construction patterns follow from the survey's taxonomy.
Contrastive windows pair two contexts that share observations but differ in rewards, dynamics, or action semantics.
Shift-labeled windows include explicit change-point metadata, useful for supervised auxiliary losses and for measuring whether a model can use clean signals.
Shift-unlabeled windows remove metadata and require inference from feedback.
Recurring-mode windows force the model to store old evidence without letting it dominate every later decision.
Contaminated windows insert invalid demonstrations or retrieved trajectories, making robustness to stale evidence part of the training objective.

These patterns can be mixed with existing ICRL training objectives.
For behavior cloning, the key is to avoid rewarding stale imitation after a shift.
For return conditioning, the key is to ensure that high desired returns cannot be achieved by repeating pre-shift actions.
For Q-learning objectives, the key is to provide enough post-shift evidence for the value function to disambiguate regimes.
For model-based objectives, the key is to evaluate predictions separately before and after changes.
The training objective need not explicitly name ``context validity'', but it must create a learning signal where validity affects action quality.

\clearpage
\section{Evaluation of Adaptation Beyond Memorization}
\label{sec:evaluation}

\subsection{Limits of Average Return}

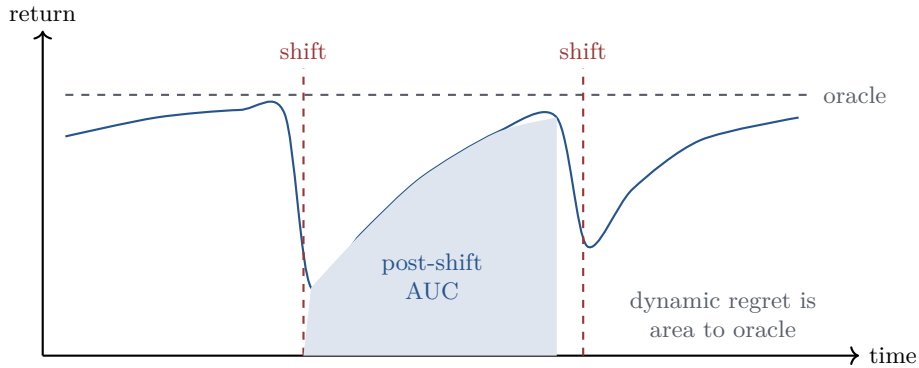
\begin{figure}[htpb]
\centering
\begin{tikzpicture}[
  every node/.style={font=\small},
  arr/.style={-{Latex[length=2mm]}, thick}
]
\draw[->, thick] (0,0) -- (10.8,0) node[right] {time};
\draw[->, thick] (0,0) -- (0,4.3) node[above] {return};
\draw[thick, nsgray, dashed] (0.3,3.45) -- (10.2,3.45) node[right, font=\small] {oracle};
\draw[thick, nsblue] plot[smooth] coordinates {
  (0.3,2.9) (1.5,3.15) (2.6,3.25) (3.2,3.2)
  (3.55,0.9) (4.2,1.6) (5.0,2.35) (6.0,2.95) (6.8,3.15)
  (7.2,1.45) (7.8,2.2) (8.7,2.85) (10.0,3.15)
};
\draw[thick, nsred, dashed] (3.45,0) -- (3.45,3.8);
\draw[thick, nsred, dashed] (7.15,0) -- (7.15,3.8);
\node[nsred, above] at (3.45,3.8) {shift};
\node[nsred, above] at (7.15,3.8) {shift};
\draw[<->, thick, nsorange] (3.55,0.75) -- (5.85,0.75);
\node[nsorange, below] at (4.7,0.75) {recovery time};
\fill[nsblue!15] (3.45,0) -- plot[smooth] coordinates {(3.55,0.9) (4.2,1.6) (5.0,2.35) (6.0,2.95) (6.8,3.15)} -- (6.8,0) -- cycle;
\node[align=center, font=\small, nsblue] at (5.15,1.05) {post-shift\\AUC};
\node[align=center, font=\small, nsgray] at (9.0,0.55) {dynamic regret is\\area to oracle};
\end{tikzpicture}
\caption{Lifecycle evaluation quantities for non-stationary ICRL. The shape of the curve is more informative than its average.}
\label{fig:evaluation-curve}
\end{figure}

Average return is necessary, but it is a poor summary of adaptation.
It collapses the lifetime into one number and can hide the exact interval where non-stationarity matters.
One agent can score well by performing strongly before a shift and failing afterward.
Another can recover after each shift but have lower steady-state return.
For non-stationary ICRL, the curve itself is the evidence: behavior before the shift, damage immediately afterward, recovery, and performance when an old mode returns.
This aligns with non-stationary bandits' move from static to dynamic comparators \citep{garivier2011ucb,besbes2014nonstationarybandit,li2019online} and with analogous non-stationary RL analyses \citep{lecarpentier2019nonstationary,cheung2020nonstationary,fei2020dynamic}.
Related MDP regret work \citep{wei2021nonstationary} and deep RL generalization studies \citep{packer2018generalization,kirk2023zeroshot} make the same warning concrete: held-out performance alone can hide shortcut learning.
Figure~\ref{fig:evaluation-curve} illustrates the lifecycle quantities that should be read from this curve.

\subsection{Core Metrics}

Several metrics have to be read together because each exposes a different failure mode.
For a shift at $\tau$, post-shift area under the curve can be reported as
\begin{equation}
  \mathrm{AUC}_{\tau:K}
  =
  \sum_{k=0}^{K-1} J_{\tau+k}^{\policy},
  \label{eq:post-shift-auc}
\end{equation}
while stale-context sensitivity can be measured by the drop between a valid context and a deliberately contaminated one:
\begin{equation}
  S_{\mathrm{stale}}
  =
  J\!\left(\policy,g_{\mathrm{valid}}\right)
  -
  J\!\left(\policy,g_{\mathrm{valid}\cup\mathrm{stale}}\right).
  \label{eq:stale-sensitivity}
\end{equation}
To separate memory scale from adaptation quality, context efficiency can be normalized as
\begin{equation}
  \eta_{\mathrm{ctx}}
  =
  \frac{J(\policy,g)-J(\policy,g_{\emptyset})}
       {\mathrm{tokens}(\ctx)+\rho\,\mathrm{retrievals}(\ctx)},
  \label{eq:context-efficiency}
\end{equation}
where $\rho$ converts retrieval calls into a token-equivalent cost.
Table~\ref{tab:metrics} lists these metrics and explains what each one measures.

\begin{table}[htpb]
\caption{Recommended metrics for non-stationary ICRL.}
\label{tab:metrics}
\centering
\begingroup
\SurveyTableSetup
\begin{tabularx}{\linewidth}{@{}
  B{0.18\linewidth}
  L{0.2\linewidth}Y@{}}
\toprule
\textbf{Metric} & \textbf{Measure} & \textbf{Use} \\
\midrule
Post-Shift AUC & Return area after a shift. & Summarizes damage and recovery \citep{wang2025cicrl}. \\
Recovery Time & Steps to near-oracle return. & Reports adaptation latency \citep{adams2007changepoint,wang2025cicrl}. \\
Dynamic Regret & Gap to a changing comparator. & Aligns ICRL with non-stationary regret theory \citep{besbes2014nonstationarybandit,cheung2020nonstationary,chen2025dynamicregret}. \\
Stale-Context Sensitivity & Drop under invalid old context. & Detects over-reliance on stale evidence \citep{schmied2024radt,berkes2026spice}. \\
Context Efficiency & Return per context or memory cost. & Separates adaptation from long-context scale \citep{lu2023s4icrl,grigsby2024amago2,sridhar2024regent}. \\
Retrieval Utility & Gain from valid versus bad retrieval. & Audits memory selection \citep{goyal2022retrievalrl,schmied2024radt,sridhar2024regent}. \\
Re-Adaptation Latency & Recovery when an old mode returns. & Tests recurrent-mode retention \citep{khetarpal2022continual,abel2023definition}. \\
Change-Detection Quality & Precision, recall, delay, calibration. & Evaluates inferred-shift reliability \citep{wang2025cicrl,chen2025dynamicregret}. \\
\bottomrule
\end{tabularx}
\endgroup
\end{table}

\subsection{Benchmark Dimensions}

A useful benchmark suite has to vary four dimensions at once: source of change, temporal pattern, observability, and context budget.
The first includes reward, dynamics, observation, action semantics, task distribution, and behavior policy.
The second includes abrupt, gradual, recurring, expanding, and adversarial changes.
The third distinguishes explicit, partial, latent, delayed, and noisy evidence.
The fourth covers short context, long context, external retrieval, compressed memory, and no-memory baselines.
Without this structure, a benchmark can easily support claims that are technically correct but much narrower than the label ``non-stationary ICRL'' suggests.
Table~\ref{tab:benchmark-blueprint} translates the factorial design into concrete diagnostic benchmark blocks.

Procedural environment families are especially useful for this purpose.
XLand-MiniGrid \citep{nikulin2024xlandminigrid} provides scalable JAX meta-RL environments with many task instances, and XLand-100B \citep{nikulin2025xland} turns this family into a large-scale ICRL learning-history dataset.
Kinetix \citep{matthews2024kinetix} provides an open-ended 2D physics-based control space with procedurally generated tasks.
NS-Gym \citep{keplinger2025nsgym} is a more direct non-stationary MDP substrate, while PHYRE \citep{bakhtin2019phyre}, I-PHYRE \citep{li2024iphyre}, and DeepPHY \citep{xu2025deepphy} provide interactive physical reasoning tasks where repeated attempts and visual dynamics can be repackaged into context-adaptation protocols \citep{xu2026icprl}.
These environments are not automatically non-stationary merely because they are procedural; they become non-stationary ICRL benchmarks when the evaluator constructs within-lifetime schedules that switch, drift, recur, or corrupt rewards, dynamics, observations, action interfaces, or task distributions.
Offline datasets such as D4RL \citep{fu2020d4rl}, RL Unplugged \citep{gulcehre2020rlunplugged}, and Atari replay \citep{agarwal2020optimistic} can also be useful when they are repackaged into contextual windows with controlled policy quality, support mismatch, and post-shift evidence, but they should not be confused with non-stationary benchmarks unless the test protocol actually changes the task within the context lifetime.

\subsection{Candidate Environment Families}

Several existing environment families are useful substrates for non-stationary ICRL, even if they were not originally introduced under that label.
Their value is not that they automatically solve the benchmark-design problem, but that they are programmable, procedurally diverse, memory-demanding, or already tied to continual-learning metrics.
Table~\ref{tab:candidate-environments} groups these candidates by the kind of non-stationary ICRL evidence they can support.

The critical distinction is between an \emph{environment substrate} and a \emph{non-stationary ICRL protocol}.
Procgen \citep{cobbe2020procgen}, MiniGrid \citep{chevalierboisvert2023minigrid}, Craftax \citep{matthews2024craftax}, MiniHack \citep{samvelyan2021minihack}, and Jumanji \citep{bonnet2024jumanji} are not evidence of non-stationary ICRL merely because they generate many tasks.
They become relevant when the evaluator specifies a within-lifetime shift process, keeps old context visible after the shift, prevents online parameter updates, and measures recovery through context.
Continual RL benchmarks such as CRLMaze \citep{lomonaco2019crlmaze} and CORA \citep{powers2021cora} provide useful shift schedules and metrics, but they need the same care because many continual RL baselines assume online learning.
An ICRL protocol should freeze the main policy parameters and compare raw-history, recent-only, retrieval, compression, oracle-change-point, and stale-context adversary variants.

\begingroup
\SurveyLongTableSetup
\begin{longtable}{@{}B{0.14\linewidth}L{0.36\linewidth}L{0.28\linewidth}L{0.18\linewidth}@{}}
\caption{Candidate environment families for non-stationary ICRL evaluation.}
\label{tab:candidate-environments}\\
\toprule
\textbf{Environment Family} & \textbf{Why It Is Useful} & \textbf{Non-Stationary Protocol to Add} & \textbf{Best Role in This Survey} \\
\midrule
\endfirsthead
\caption[]{Candidate environment families for non-stationary ICRL evaluation (continued).}\\
\toprule
\textbf{Environment Family} & \textbf{Why It Is Useful} & \textbf{Non-Stationary Protocol to Add} & \textbf{Best Role in This Survey} \\
\midrule
\endhead
\midrule
\multicolumn{4}{r}{Continued on next page}\\
\endfoot
\bottomrule
\endlastfoot
AnyMDP / OmniRL & Randomized MDPs are generated by varying transitions and rewards, directly matching the formal ICRL setting of adapting to unseen decision processes \citep{wang2025anymdp}. & Switch or recur latent MDPs within a lifetime; vary reward and transition budgets; keep pre-shift context in the window; compare oracle restart with learned context filtering. & Controlled mechanism layer for reward, transition, and latent-mode shift. \\
Contextual / Hidden-Parameter MDP Generators & Contextual MDPs \citep{hallak2015contextualmdp} and hidden-parameter MDPs \citep{killian2017hipmdp,doshivelez2013hipmdp,perez2020generalizedhipmdp} use latent variables to define rewards and dynamics while preserving shared structure. & Let the latent parameter drift, switch, or recur within a lifetime; hide the parameter from the agent and require inference from context. & Formal mechanism layer for latent-regime inference. \\
MiniGrid / BabyAI & MiniGrid \citep{chevalierboisvert2023minigrid} and BabyAI \citep{chevalierboisvert2019babyai} are lightweight grid and instruction environments that are modular and easy to customize. & Change goals, object rules, door-key dependencies, instructions, observation masks, or action permutations after context has accumulated. & Minimal diagnostic layer for action semantics, instruction shift, and stale demonstrations. \\
Alchemy & Latent causal rules are procedurally resampled and require online inference, hypothesis testing, and action sequencing \citep{wang2021alchemy}. & Change the hidden chemistry during a lifetime, make old chemistry recur, or insert examples from incompatible chemistries into retrieval memory. & Latent-rule inference layer for context ambiguity and recurring modes. \\
XLand-MiniGrid / XLand-100B & XLand-MiniGrid \citep{nikulin2024xlandminigrid} and XLand-100B \citep{nikulin2025xland} make broad ICRL pretraining and evaluation possible through scalable JAX meta-RL environments and large learning-history data. & Schedule task, layout, rule, reward, or action-interface changes across a single lifetime; hold context budget fixed across append-all and retrieval methods. & Large-scale meta-RL and learning-history layer. \\
Kinetix & Open-ended 2D physics-based control tasks provide procedural diversity with controllable physical structure \citep{matthews2024kinetix}. & Drift gravity, mass, joints, morphology, object layout, goals, or reward weights; test whether old physics evidence is downweighted. & Open-ended physics layer for dynamics and morphology-like shift without making robotics central. \\
PHYRE / I-PHYRE / DeepPHY & PHYRE \citep{bakhtin2019phyre}, I-PHYRE \citep{li2024iphyre}, and DeepPHY \citep{xu2025deepphy} expose action-outcome evidence, visual dynamics, repeated attempts, and VLM-compatible observations relevant to world-model-guided ICRL \citep{xu2026icprl}. & Change friction, gravity, object properties, action discretization, visual annotations, or reward rules across attempts; keep failed pre-shift attempts in context and measure whether the agent updates physical intuition. & Physically grounded dynamics and observation layer for world-model-guided ICRL. \\
Procgen & Procedurally generated games test visual and level generalization across many instances \citep{cobbe2020procgen}. & Change level generators, visual themes, object layouts, difficulty, or reward rules during a lifetime; include recurring generator families. & Visual procedural layer for observation and task-distribution shift. \\
Craftax & Fast JAX open-ended crafting tasks require exploration, long-term planning, memory, and adaptation to newly discovered situations \citep{matthews2024craftax}. & Drift resource distributions, recipes, tool effects, biome rules, survival objectives, or inventory constraints; test old-recipe contamination. & Open-ended memory and planning layer, complementary to Kinetix. \\
MiniHack / NetHack Learning Environment & MiniHack \citep{samvelyan2021minihack} is a sandbox for designing rich NetHack-based tasks, while NLE \citep{kuttler2020nethack} provides a stochastic, procedurally generated roguelike environment. & Change object semantics, dungeon families, monster/item distributions, reward shaping, or action effects; test whether retrieved dungeon memories remain valid. & Rich single-agent procedural layer for long-horizon memory and sparse rewards. \\
Jumanji & JAX environments for combinatorial and general decision problems are scalable and customizable in initial-state distribution and problem complexity \citep{bonnet2024jumanji}. & Drift problem size, constraints, cost functions, instance distributions, or resource budgets; measure context efficiency under fixed retrieval cost. & Non-visual combinatorial layer for budget and constraint shift. \\
POPGym / POPGym Arcade & POPGym \citep{morad2023popgym} explicitly stresses memory in POMDPs; POPGym Arcade \citep{wang2025popgymarcade} provides pixel-based fully and partially observable variants and studies memory contamination by old observations. & Switch observability regimes, delay change evidence, inject stale observations, or alternate fully observable and partially observable modes. & Memory-validity layer for partial observability and context poisoning. \\
NS-Gym & NS-Gym \citep{keplinger2025nsgym} is a benchmark suite explicitly designed for non-stationary MDPs; it connects naturally to non-stationary reward and dynamics algorithms \citep{lecarpentier2019nonstationary,chandak2020optimizing}, latent-process algorithms \citep{luo2024act}, and policy-search settings \citep{pettet2024policysearch}. & Freeze the agent after pretraining; expose regime changes only through interaction history; compare append-all, recent-only, oracle-change, retrieval, and learned context filters. & Direct bridge from non-stationary MDP evaluation to frozen-parameter ICRL. \\
CRLMaze & A 3D continual RL benchmark explicitly built around non-stationary environmental changes \citep{lomonaco2019crlmaze}. & Freeze the policy after pretraining and expose changes only through interaction history; compare ICRL context policies against online continual RL baselines. & Bridge case for explicit non-stationary RL environments. \\
CORA & A continual RL platform with benchmarks, baselines, and metrics for forgetting, plasticity, generalization, and forward transfer \citep{powers2021cora}. & Recast Atari, Procgen, or NetHack task streams as frozen-parameter context-adaptation tests; report CORA-style forgetting and transfer with ICRL context ablations. & Protocol and metric bridge from continual RL to ICRL. \\
ScienceWorld / WebShop / ALFWorld & ScienceWorld \citep{wang2022scienceworld}, WebShop \citep{yao2022webshop}, and ALFWorld \citep{shridhar2020alfworld} expose feedback, tool/action histories, and task instructions in interactive language environments. & Drift user preferences, reward functions, hidden instructions, website/product distributions, or feedback reliability across trials. & Extended LLM-agent layer, included only when feedback drives repeated sequential decisions. \\
\end{longtable}
\endgroup

This table also clarifies what should be excluded from the core claim.
Robotics-specific suites and controllers such as Meta-World \citep{yu2020metaworld}, Continual World \citep{wolczyk2022disentangling}, LIBERO \citep{liu2023libero}, RoboMimic \citep{mandlekar2021robomimic}, Open X-Embodiment \citep{openxembodiment2024}, RoboMME \citep{dai2026robomme}, and humanoid whole-body-control systems \citep{ding2025jaeger} are informative for continual learning and generalist action modeling, but they are outside the main scope of this survey.
They can be mentioned as adjacent benchmarks when discussing evaluation traditions, but the paper should not rely on them for its central non-stationary ICRL argument.
Similarly, multi-agent suites should be avoided because they introduce strategic non-stationarity, for example when concurrently updating agents make the environment non-stationary from each individual agent's perspective \citep{luo2023modelbased}, which is explicitly outside scope here.

\begin{table}[htbp]
\caption{A benchmark blueprint for separating adaptation mechanisms.}
\label{tab:benchmark-blueprint}
\centering
\begingroup
\SurveyTableSetup
\begin{tabularx}{\linewidth}{@{}
  B{0.12\linewidth}
    L{0.5\linewidth}
      L{0.16\linewidth}
  Y@{}}
\toprule
\textbf{Test Block} & \textbf{Controlled Change} & \textbf{Intended Capability} & \textbf{Shortcut It Rules Out} \\
\midrule
Reward Switch & Same observations and dynamics, new reward or preference in bandit settings \citep{garivier2011ucb,besbes2014nonstationarybandit}, safe-policy settings \citep{chandak2020safepolicyimprovement,moeini2025safeicrl}, or reward-feedback agents \citep{song2026reward}. & Reinterpret feedback and stop stale imitation. & Static task recognition from observations. \\
Dynamics Drift & Same reward, changed transition kernel or latent physical parameters in non-stationary RL \citep{lecarpentier2019nonstationary,chandak2020optimizing,feng2023nonstationary}, DICP \citep{son2025dicp}, or ICPRL \citep{xu2026icprl}. & Infer action-outcome evidence in context. & Copying old high-return action sequences. \\
Action Remapping & Same goal, permuted or rescaled action effects in changing-action-set settings \citep{chandak2020changingaction} or variable-action ICRL \citep{sinii2024variable}. & Detect interface change and relearn action semantics. & Treating action tokens as globally stable. \\
Latent Recurrence & Earlier mode returns after intervening modes in Alchemy-style tasks \citep{wang2021alchemy}, reactive or recurring-mode settings \citep{steinparz2022reactive}, retrieval-based RL \citep{goyal2022retrievalrl}, RA-DT \citep{schmied2024radt}, or REGENT \citep{sridhar2024regent}. & Retrieve old valid memory without over-forgetting. & Recency-only adaptation. \\
Stale-Memory Attack & Insert high-confidence but invalid observations in POPGym-style memory tasks \citep{morad2023popgym,wang2025popgymarcade}, invalid demonstrations through history filtering or suboptimal-data tests \citep{chen2025filtering,wang2026suboptimalicrl}, or conflicting Bayesian evidence \citep{berkes2026spice}. & Downweight misleading context. & Long-context or retrieval scale alone. \\
Delayed Feedback & Shift is only visible after long-horizon outcomes in S4-style ICRL \citep{lu2023s4icrl}, AMAGO-style agents \citep{grigsby2023amago,grigsby2024amago2}, RELIC-style agents \citep{elawady2024relic}, or POPGym \citep{morad2023popgym}. & Maintain uncertainty and explore diagnostically. & Myopic reward matching. \\
Procedural Task Stream & Generate scheduled task, layout, physics, rule, or goal changes in XLand-MiniGrid/XLand-100B \citep{nikulin2024xlandminigrid,nikulin2025xland}, Kinetix \citep{matthews2024kinetix}, Procgen \citep{cobbe2020procgen}, Craftax \citep{matthews2024craftax}, MiniHack \citep{samvelyan2021minihack}, or Jumanji \citep{bonnet2024jumanji}. & Test adaptation under controlled diversity and repeated shift seeds. & Overfitting to a small hand-written set of shifts. \\
\bottomrule
\end{tabularx}
\endgroup
\end{table}

\subsection{Constructing Stale-Context Stress Tests}

A non-stationary ICRL benchmark should be specified as a protocol over lifetimes, not only as a set of environment IDs.
The core intervention is simple: create a period in which old context is genuinely useful, change the decision process, and then keep some of the old evidence visible so the agent must decide whether to trust it.
This makes the failure modes in Table~\ref{tab:failure-modes} operational.
For example, to test stale imitation, the evaluator can collect high-return trajectories under regime $m_0$, switch to a regime $m_1$ with the same observations but different rewards or dynamics, and inject the $m_0$ trajectories into the post-shift context while also providing a small amount of valid $m_1$ evidence.
If performance drops relative to the same post-shift context without the injected trajectories, the agent is not merely slow to learn; it is being actively pulled by stale evidence.
This protocol connects continual ICRL evaluation \citep{wang2025cicrl}, history filtering and weak-data ICRL \citep{chen2025filtering,wang2026suboptimalicrl}, retrieval-augmented ICRL \citep{schmied2024radt,sridhar2024regent}, and non-stationary MDP or bandit stress tests \citep{besbes2014nonstationarybandit,cheung2020nonstationary,fei2020dynamic}.

A minimal protocol can be written as follows:
\begin{enumerate}
  \item \textbf{Choose a base environment and shift family.} Specify which component changes: reward, transition, observation, action semantics, constraint, task distribution, or behavior policy \citep{lecarpentier2019nonstationary,chandak2020changingaction,moeini2025safeicrl}.
  \item \textbf{Generate a lifetime.} Run a pre-shift phase under $m_0$, a post-shift phase under $m_1$, and optionally a recurrence phase where $m_0$ returns \citep{wang2025cicrl,steinparz2022reactive}.
  \item \textbf{Build paired contexts.} For the same post-shift query state, construct at least four contexts: no-context, valid-current context, valid-current plus stale context, and oracle-truncated context that removes pre-shift evidence.
  \item \textbf{Freeze adaptation channels.} Keep policy parameters fixed and vary only the context policy: append-all, last-$k$, retrieval, compression, learned filter, oracle filter, and stale adversary \citep{moeini2025surveyicrl,schmied2024radt,chen2025filtering}.
  \item \textbf{Measure lifecycle outcomes.} Report pre-shift return, immediate post-shift drop, recovery time, post-shift AUC, dynamic regret, stale-context sensitivity, and recurring-mode re-adaptation \citep{wang2025cicrl,chen2025dynamicregret}.
  \item \textbf{Attribute the failure.} Use ablations to determine whether the bottleneck is change detection, retrieval, compression, influence weighting, forgetting, or policy capacity.
\end{enumerate}

The most important design rule is to use paired evaluations.
The valid and contaminated contexts should be identical except for the intervention being tested.
Otherwise, a method can look robust simply because the contaminated run also has more useful post-shift evidence, or it can look fragile because the contaminated context changes budget, ordering, or formatting.
When possible, the injected stale item should be strong under its original regime: a high-return demonstration, a confident value estimate, a retrieved near-neighbor, or a concise verbal rule.
Weak stale evidence is too easy to ignore and underestimates the risk faced by long-context and retrieval-augmented agents.
This paired-design principle is the context analogue of oracle-change and controlled-ablation baselines in changepoint, non-stationary regret, and ICRL reporting \citep{adams2007changepoint,chen2025dynamicregret,moeini2025surveyicrl}.

\begin{table}[htbp]
\caption{Concrete context-window interventions for stale-context stress tests.}
\label{tab:stress-interventions}
\centering
\begingroup
\SurveyTableSetup
\begin{tabularx}{\linewidth}{@{}
  B{0.13\linewidth}
  L{0.3\linewidth}
  L{0.25\linewidth}
  Y@{}}
\toprule
\textbf{Target Failure} & \textbf{Context Intervention} & \textbf{Paired Control} & \textbf{Diagnostic Signal} \\
\midrule
Stale Imitation & Inject high-return trajectories or demonstrations from the pre-shift reward/dynamics regime into the post-shift prompt. Keep observations similar so the old behavior looks relevant \citep{laskin2023algorithm,chen2021decision,wang2026suboptimalicrl}. & Same post-shift context with the stale trajectories removed, replaced by neutral padding, or replaced by valid current-regime trajectories of matched length. & Large drop in post-shift AUC or higher dynamic regret indicates over-imitation of obsolete behavior \citep{wang2025cicrl,chen2025dynamicregret}. \\
Context Inertia & Vary the amount of pre-shift history while holding the amount of post-shift evidence fixed. Test short, medium, and long stale prefixes \citep{grigsby2024amago2,elawady2024relic,li2024longcontext}. & Oracle-truncated context that starts at the shift point, plus last-$k$ context with the same token budget. & Recovery time increasing with stale-prefix length indicates that old context dominates new evidence. \\
Retrieval Contamination & Seed the memory database with near-neighbor trajectories from the wrong reward, dynamics, or action-interface regime, then query after the shift \citep{goyal2022retrievalrl,schmied2024radt,sridhar2024regent}. & Validity-filtered retrieval, oracle-current-mode retrieval, and random retrieval with matched retrieval count. & Retrieval improves stationary return but hurts post-shift return, or selected items have low current-regime validity. \\
Compression Loss & Compress a mixed pre-shift/post-shift history so that the rare change evidence is easy to omit, such as one reward reversal or one transition anomaly \citep{zhong2023memorybank,packer2023memgpt,chen2025filtering}. & Raw context containing the decisive post-shift event, and an oracle summary that preserves the event. & Compressed memory performs well on average but misses the shift or recovers late. \\
Action-Token Mismatch & Keep old action tokens in context after permuting, masking, rescaling, or renaming actions. The old trajectory should be high return before remapping \citep{chandak2020changingaction,sinii2024variable}. & Same remapped environment with action descriptions, interface metadata, or oracle remapping supplied. & Correct high-level behavior but wrong low-level actions indicates stale action semantics. \\
Over-Forgetting & Use an $m_0 \rightarrow m_1 \rightarrow m_0$ schedule and remove or downweight the original $m_0$ evidence before recurrence \citep{khetarpal2022continual,steinparz2022reactive,wang2025cicrl}. & Memory-isolated or oracle-regime context that preserves old $m_0$ evidence for reuse. & Good first recovery but poor recurrence performance indicates excessive forgetting. \\
Verbal Stale Memory & Insert an old natural-language rule, reflection, or reward explanation that was true before a preference or tool-schema shift \citep{shinn2023reflexion,zhao2024expel,wang2023voyager,song2026reward}. & Same transcript with timestamped, contradicted, or retired memory; oracle-current-rule context. & The agent follows obsolete verbal advice despite contrary feedback. \\
\bottomrule
\end{tabularx}
\endgroup
\end{table}

These interventions also define a practical ``non-stationary ICRL-ready'' benchmark checklist.
A benchmark is ready for this survey's claims only if it can freeze the policy, expose the agent to within-lifetime shifts, keep old context available after the shift, inject or remove specific context items, control context budget, and report paired lifecycle metrics.
It should also support oracle variants: oracle-current-mode context, oracle-change-point truncation, oracle retrieval, and oracle summary.
Without these controls, a benchmark may still be useful for generalization or continual learning, but it cannot isolate whether a frozen policy adapts through context under stale evidence.

\begin{table}[htbp]
\caption{Minimum reporting checklist for non-stationary ICRL experiments.}
\label{tab:evaluation-checklist}
\centering
\begingroup
\SurveyTableSetup
\begin{tabularx}{\linewidth}{@{}B{0.18\linewidth}Y@{}}
\toprule
\textbf{Item} & \textbf{What to Report} \\
\midrule
Shift Specification & Which MDP component changes, when it changes, and whether the change is abrupt, gradual, or recurring \citep{adams2007changepoint,lecarpentier2019nonstationary}; regret and ICRL variants give complementary examples \citep{feng2023nonstationary,wang2025cicrl}. \\
Shift Observability & Whether the change is signaled, partially observable, latent, delayed, or noisy \citep{kaelbling1998pomdp,hallak2015contextualmdp,ghosh2021epistemic}; supervised and Bayesian ICRL variants make this observable through context \citep{lee2023supervised,berkes2026spice}. \\
Training Exposure & Whether similar changes appear during pretraining or only at test time \citep{xu2024psbl,wang2025cicrl}. \\
Context Policy & What is written, retrieved, compressed, forgotten, and reset \citep{schmied2024radt,chen2025filtering}. \\
Budget & Context length, retrieval count, memory size, latency budget, and computational budget \citep{grigsby2024amago2,sridhar2024regent}. \\
Baselines & No-context, recent-only, full-history, oracle-change-point, random-retrieval, and stale-retrieval baselines where applicable \citep{garivier2011ucb,moeini2025surveyicrl}. \\
Lifecycle Curves & Return or regret before shift, immediately after shift, during recovery, and on recurring modes \citep{chen2025dynamicregret,wang2025cicrl}. \\
Ablations & Remove post-shift evidence, inject stale context, permute context order, and vary pre-shift context length \citep{chen2025filtering,berkes2026spice}. \\
Uncertainty & Variance across shift seeds, task sequences, and change frequencies \citep{besbes2014nonstationarybandit,russac2019weighted,cheung2020nonstationary}; MDP and evaluation studies provide related variance sources \citep{fei2020dynamic,feng2023nonstationary,kim2020randomized}. \\
\bottomrule
\end{tabularx}
\endgroup
\end{table}

\subsection{Baselines That Matter}

The most informative baselines are often simple.
A recent-only context baseline tests whether long-term memory is helping or hurting.
A full-history baseline tests whether selective retrieval matters.
A no-context baseline tests whether performance is static generalization.
An oracle-change-point baseline estimates the value of knowing when to forget.
A stale-context adversary estimates how badly the model can be misled by old evidence.
For retrieval methods, random retrieval and similarity-only retrieval should be compared with any learned validity-aware retrieval method.
The minimum reporting items in Table~\ref{tab:evaluation-checklist} are intended to make these baselines and ablations comparable across papers.

\subsection{Distinguishing Adaptation from Memorization}

Non-stationary evaluation should rule out three shortcuts.
The first is memorization of task IDs or environment templates.
The second is recency-only adaptation, where the model ignores long context and simply follows the last few rewards.
The third is retrieval shortcutting, where a memory database contains near-duplicates of test tasks.
Countermeasures include held-out generators, changed task order, unseen shift schedules, adversarial stale-memory insertion, and explicit tests of recurring modes where recency alone is insufficient.

\subsection{A Practical Evaluation Stack}

A practical stack should include at least three layers.
The first layer is a controlled diagnostic layer.
AnyMDP \citep{wang2025anymdp}, MiniGrid \citep{chevalierboisvert2023minigrid}, BabyAI \citep{chevalierboisvert2019babyai}, and Alchemy-style tasks \citep{wang2021alchemy} are appropriate here because the evaluator can exactly control which part of the decision process changes.
This layer should be used for mechanism claims: reward switches, transition switches, action remapping, hidden rule changes, and context contamination.
The experiments should be small enough that oracle baselines are meaningful, including oracle-change-point, oracle-retrieval, and oracle-current-mode policies.
Generalization benchmarks should be used with the same caution: a held-out level or task family tests extrapolation, whereas non-stationary ICRL requires within-lifetime invalidation and recovery \citep{packer2018generalization,ghosh2021epistemic,kirk2023zeroshot}.

The second layer is a scalable procedural layer.
XLand-MiniGrid and XLand-100B \citep{nikulin2024xlandminigrid,nikulin2025xland}, Kinetix \citep{matthews2024kinetix}, PHYRE/I-PHYRE/DeepPHY \citep{bakhtin2019phyre,li2024iphyre,xu2025deepphy}, Procgen \citep{cobbe2020procgen}, Craftax \citep{matthews2024craftax}, MiniHack/NLE \citep{samvelyan2021minihack,kuttler2020nethack}, and Jumanji \citep{bonnet2024jumanji} are natural candidates because they support many task instances while remaining programmable enough to impose explicit non-stationary schedules.
This layer should test whether a method that works on clean diagnostics still works when observations are richer, horizons are longer, rewards are sparser, and the task distribution has many modes.
It should also test context efficiency: if a method only succeeds by appending all experience, it may not be viable once the environment family becomes large.

The third layer is a memory and continual-protocol layer.
POPGym \citep{morad2023popgym}, POPGym Arcade \citep{wang2025popgymarcade}, NS-Gym \citep{keplinger2025nsgym}, CRLMaze \citep{lomonaco2019crlmaze}, and CORA \citep{powers2021cora} are useful here because they foreground memory, partial observability, non-stationary streams, and continual-learning metrics.
This layer should measure whether the context policy can handle delayed evidence, old-observation poisoning, forward transfer, forgetting, and re-adaptation when old modes return.
Unlike the first two layers, this layer is less about task diversity and more about diagnosing the lifetime of information.

An optional extended layer can include interactive language environments such as ScienceWorld \citep{wang2022scienceworld}, WebShop \citep{yao2022webshop}, and ALFWorld \citep{shridhar2020alfworld} when the paper explicitly studies LLM reward-feedback agents.
These environments are useful for testing preference drift, natural-language feedback, tool-use histories, and changing user instructions.
However, they should remain extended cases unless the experiment contains repeated decisions, external feedback, and a clear mechanism by which future behavior improves through context.
Otherwise, the result is better described as conversational memory or prompting rather than ICRL.
For LLM-agent bridge cases, a non-stationary protocol should vary user preferences, reward feedback, hidden instructions, tool schemas, website distributions, or feedback reliability across repeated trials.
Baselines should include no-memory, last-$k$ context, full transcript, summary memory, retrieval memory, oracle-current-rule context, and stale-memory adversary variants.
Metrics should include post-shift success rate, correction latency, invalid tool-call rate, stale-memory sensitivity, and preference compliance after drift.

Across all layers, the stack should be reported as a lifecycle evaluation rather than as a single leaderboard.
The paper should state the shift source, temporal pattern, observability, context budget, and reset rule for every environment.
It should also state whether the environment is intrinsically non-stationary or merely a programmable substrate used to construct a non-stationary protocol.
This distinction is important for review: success on a broad procedural benchmark is not automatically success on non-stationary ICRL unless stale context is present and recovery must occur through context.

\subsection{Interpreting Evaluation Outcomes}

Evaluation should be diagnostic enough to explain why a method succeeds.
High return with high stale-context sensitivity suggests that the method is powerful but brittle.
Low stale-context sensitivity with poor recurring-mode performance suggests over-forgetting.
Good recovery with poor dynamic regret may indicate that the method eventually adapts but pays a large cost after every change.
Strong oracle-change-point performance and weak learned-change-point performance indicates that the main bottleneck is detection or context management rather than policy capacity.

The conclusion is that non-stationary ICRL evaluation should not end with a single return number.
A convincing result must identify which failure mode was solved: recognizing that the regime changed, selecting evidence from the current regime, retaining useful evidence from older recurring regimes, or acting well after the relevant context has been selected.
The strongest evidence is therefore a consistent pattern across return, dynamic regret, recovery time, stale-context sensitivity, and oracle gaps.
If a method improves return but remains sensitive to stale context, the claim should be limited to better exploitation under favorable histories.
If it closes the learned-to-oracle change-point gap, the claim can be about change detection or context filtering.
If it improves recurring-mode performance without hurting post-shift recovery, the claim can be about useful long-term memory.
This interpretation discipline keeps benchmark results tied to the central question of the survey: whether a frozen policy can use context to adapt under changing decision processes, not merely whether it performs well on a diverse task suite.

\clearpage
\section{Theory and Conceptual Links}
\label{sec:theory}

\subsection{Existing Theoretical Coverage}

Existing theory has begun to make ICRL more than an analogy between prompting and reinforcement learning.
Supervised pretraining analyses show that decision transformers can implement in-context policies under structured task distributions \citep{lee2023supervised,lin2023transformers}.
General ICL theory supplies the broader mechanisms: transformers can learn simple function classes in context \citep{garg2022transformers,akyurek2023whatlearning}, encode implicit models or gradient-descent-like computations inside the forward pass \citep{vonoswald2023gradient,akyurek2024contextlanguage}, and depend strongly on demonstration structure \citep{min2022rethinking,wies2023learnability}.
Recent work studies chain-of-thought or explicit iterative computation as a path toward in-context reinforcement learning \citep{xie2026cot}.
Other work shows that transformers can implement policy-improvement-like procedures in context \citep{liang2026policyimprovement}.
These results are important because they show that ICRL is not merely an empirical analogy: under certain assumptions, the architecture can represent update-like computations.
Adjacent non-stationary ICL theory \citep{qin2026beyondstationarity}, while not reinforcement learning, also supports the relevance of recency, adaptive filtering \citep{sayed2011adaptive}, gating \citep{yang2023gla,katsch2023gateloop}, retention-style sequence models \citep{sun2023retnet}, and recurrent attention mechanisms \citep{peng2024eagle} when the target function changes over time.
Long-context ICL studies provide a complementary caution: increasing the number of examples can help only when the model can keep the relevant evidence accessible and distinguish it from distractors \citep{li2024longcontext,bertsch2024longcontext}.

\subsection{Added Challenges Under Non-Stationarity}

Non-stationarity adds two theoretical problems that are easy to blur together.
The comparator problem asks what the agent should be compared against when the optimal policy changes over time.
Dynamic regret is the natural answer in many settings, including adversarial or bandit settings \citep{auer2002nonstochastic,garivier2011ucb,besbes2014nonstationarybandit}, adaptive-window or weighted bandits \citep{auer2019adaptively,russac2019weighted,li2019online}, non-stationary MDPs \citep{cheung2020nonstationary,fei2020dynamic,feng2023nonstationary}, refined MDP bounds \citep{wei2021nonstationary,mao2021nearoptimal,domingues2021kernel}, and transformer-based analyses \citep{chen2025dynamicregret}.
The validity problem asks which context items should influence the current policy.
This part is less developed theoretically.
Most in-context theory assumes that examples are sampled from a coherent latent task or distribution.
Non-stationary ICRL requires theory for contexts that are finite, ordered, partially invalid, and sometimes contradictory.
POMDP \citep{kaelbling1998pomdp}, CMDP \citep{hallak2015contextualmdp}, epistemic-POMDP \citep{ghosh2021epistemic}, task-inference \citep{humplik2019taskinference}, and HiP-MDP theory \citep{killian2017hipmdp,doshivelez2013hipmdp} suggest one route: treat the current regime as latent and context as evidence about that regime.
The missing piece is that an ICRL agent observes this evidence through a bounded and possibly learned context channel, so inference error and context-selection error become part of the control problem.
One useful decomposition is to separate environmental movement from context-management error:
\begin{equation}
  \begin{aligned}
  \mathrm{Reg}^{\mathrm{dyn}}_T(\policy,g)
  \leq{}&
  \underbrace{\mathcal{A}(T,B_T)}_{\mathrm{move}}
  +
  \underbrace{\sum_{t=1}^{T}\epsilon_{\mathrm{ctx}}(t;g)}_{\mathrm{ctx}}
  +
  \underbrace{\sum_{t=1}^{T}\epsilon_{\mathrm{model}}(t;\theta)}_{\mathrm{model}} .
  \end{aligned}
  \label{eq:regret-decomposition}
\end{equation}
Here $B_T$ is a variation budget: the movement term captures adaptation to environmental drift, the context term captures retrieval, compression, or stale-context mistakes, and the model term captures the approximation limits of the frozen decision model.
This is not a theorem as stated; it is a target form for future theory that would make context-management assumptions explicit.
The main gap is therefore an interface between ICRL theory, non-stationary RL regret analysis, and retrieval or memory theory: the target object is not only a changing MDP, but a changing MDP observed through a bounded context channel.

\subsection{Open Theoretical Questions}

Several questions would make the theory more precise:
\begin{itemize}
  \item Which variation budgets or change-point assumptions make fixed-parameter context adaptation competitive with online updating?
  \item How should regret bounds scale with context length, memory size, retrieval error, and shift observability?
  \item When is stale context harmless because attention can suppress it, and when does it induce unavoidable lower bounds?
  \item How do value-aware and Bayesian context mechanisms alter sample complexity under reward or dynamics shift?
  \item Can recurring-mode memory be given a formal advantage over recency-only adaptation?
\end{itemize}
Table~\ref{tab:theory} maps existing theory topics to the non-stationary pieces that remain missing.

A useful first step is to make the underlying assumptions explicit:
\begin{itemize}
  \item Under bounded variation in rewards or dynamics, theory could characterize how regret scales with context length.
  \item Under a finite set of recurring latent regimes, theory could identify when retrieval reduces re-adaptation latency.
  \item Under adversarial stale context, theory could characterize robustness guarantees for attention or retrieval policies.
\end{itemize}
These stylized settings would not capture the full complexity of deployment, but they would turn vague claims about ``using context'' into testable statements about when context helps and when it provably hurts.

\begin{table}[htbp]
\caption{Theory topics and their relevance to non-stationary ICRL.}
\label{tab:theory}
\centering
\begingroup
\SurveyTableSetup
\begin{tabularx}{\linewidth}{@{}
  B{0.16\linewidth}
  L{0.5\linewidth}
  Y@{}}
\toprule
\textbf{Theory Topic} & \textbf{Current Contribution} & \textbf{Missing Non-Stationary Piece} \\
\midrule
General ICL Theory & Explains function-class learning \citep{garg2022transformers,akyurek2023whatlearning}, implicit model fitting or forward-pass optimization \citep{vonoswald2023gradient,kirsch2022gpicl}, and learnability conditions \citep{wies2023learnability}. & Need sequential feedback, reward, and invalid-context extensions. \\
Supervised ICRL Theory & Explains when sequence models can map context to near-optimal decisions in supervised ICRL \citep{lee2023supervised,lin2023transformers}, TD-style ICRL \citep{wang2024tdicrl}, or provable-emergence analyses of in-context temporal-difference learning \citep{wang2025provableicrl}. & Context examples are often assumed coherent rather than stale or conflicting. \\
Policy Improvement in Context & Shows update-like computations can be represented by chain-of-thought ICRL \citep{xie2026cot}, policy-improvement transformers \citep{liang2026policyimprovement}, generalized decision transformers \citep{furuta2022gdt}, or agentic decision models \citep{liu2023agentic}. & Need finite-context and post-shift guarantees. \\
Dynamic Regret & Provides comparator for changing adversarial or bandit environments \citep{auer2002nonstochastic,garivier2011ucb,besbes2014nonstationarybandit}, weighted bandits \citep{russac2019weighted}, non-stationary MDPs \citep{cheung2020nonstationary,fei2020dynamic,feng2023nonstationary}, refined MDP bounds \citep{wei2021nonstationary}, and transformer analyses \citep{chen2025dynamicregret}. & Need bounds that include learned context construction and retrieval error. \\
Bayesian And Latent-Context Use & Frames context as evidence combined with POMDP or CMDP beliefs \citep{kaelbling1998pomdp,hallak2015contextualmdp}, HiP-MDP variables \citep{killian2017hipmdp,doshivelez2013hipmdp}, task-inference variables \citep{humplik2019taskinference}, latent-belief meta-RL \citep{rakelly2019pearl,zintgraf2020varibad}, or Bayesian value fusion \citep{berkes2026spice}. & Need misspecification and stale-evidence analysis. \\
Retrieval And Memory Theory & Studies language-agent memory \citep{zhong2023memorybank,wang2023longmem,packer2023memgpt}, retrieval-based RL \citep{goyal2022retrievalrl,humphreys2022largescale}, and retrieval-augmented ICRL \citep{schmied2024radt,sridhar2024regent}. & Need validity-aware retrieval under changing rewards, dynamics, and actions. \\
\bottomrule
\end{tabularx}
\endgroup
\end{table}

\clearpage
\section{Design Principles and Open Problems}
\label{sec:open-problems}

\subsection{Design Principles}

The preceding sections point to several design principles for non-stationary ICRL systems.

\begin{enumerate}
  \item \textbf{Treat context as a managed resource.} Longer context is useful only when the system can identify what remains valid, a lesson shared by long-memory LLM systems \citep{zhong2023memorybank,wang2023longmem,packer2023memgpt}, long-context ICL studies \citep{li2024longcontext,bertsch2024longcontext}, efficient sequence models \citep{lu2023s4icrl}, and long-context ICRL agents \citep{elawady2024relic}.
  \item \textbf{Train on invalidation, not only adaptation.} Training histories should include moments where old evidence becomes wrong, not only stationary improvement traces from algorithm distillation \citep{laskin2023algorithm}, noise or history filtering \citep{zisman2024noise,chen2025filtering}, continual or suboptimal ICRL \citep{wang2025cicrl,wang2026suboptimalicrl}, or environment design \citep{dennis2020ued,jiang2021plr}.
  \item \textbf{Separate recency from relevance.} Recency is a useful heuristic in sliding-window and discounted methods \citep{garivier2011ucb,russac2019weighted,trovo2020sliding}, but it fails under changepoints and recurring modes \citep{adams2007changepoint}.
  \item \textbf{Expose retrieval to stale-memory tests.} Retrieval should be evaluated by validity, not only similarity or return, in retrieval-based RL \citep{goyal2022retrievalrl,humphreys2022largescale} and retrieval-augmented ICRL \citep{schmied2024radt,sridhar2024regent}.
  \item \textbf{Report lifecycle curves.} A single average return hides post-shift failure and recovery in non-stationary MDPs \citep{cheung2020nonstationary,fei2020dynamic,wei2021nonstationary} and continual ICRL \citep{wang2025cicrl}.
  \item \textbf{Include oracle context baselines.} Oracle-change-point baselines \citep{adams2007changepoint} and general ICRL reporting baselines \citep{moeini2025surveyicrl} estimate the headroom available from better context management.
  \item \textbf{State the adaptation channel.} Papers should make clear whether adaptation occurs through prompt tokens, hidden state, external memory, retrieved trajectories, summaries, or online parameter updates.
\end{enumerate}

\subsection{Open Problems}

Not every item in the research agenda is open in the same sense.
Some topics already have direct starting points, such as continual ICRL benchmarks and metrics \citep{wang2025cicrl}, dynamic-regret analyses for non-stationary transformers \citep{chen2025dynamicregret}, history filtering \citep{chen2025filtering}, retrieval-augmented ICRL \citep{schmied2024radt,sridhar2024regent}, and safe ICRL mechanisms \citep{moeini2025safeicrl,kwon2026qbarrier}.
Other topics are seeded only by adjacent literatures, such as changepoint detection, continual learning, retrieval, or safe RL, and still need to be reformulated under the fixed-parameter context-adaptation constraint.
A third group is more genuinely open: the field lacks clear benchmarks, formal objectives, or accepted measurement protocols.
Table~\ref{tab:open-problems} therefore distinguishes \emph{seeded} directions from \emph{partly seeded} and \emph{truly open} ones.

\begingroup
\SurveyLongTableSetup
\begin{longtable}{@{}
    B{0.14\linewidth}
    L{0.44\linewidth}
    L{0.4\linewidth}@{}}
\caption{A status-aware roadmap for non-stationary ICRL.}
\label{tab:open-problems}\\
\toprule
\textbf{Problem} & \textbf{Current Status} & \textbf{Actionable Next Step} \\
\midrule
\endfirsthead
\caption[]{A status-aware roadmap for non-stationary ICRL (continued).}\\
\toprule
\textbf{Problem} & \textbf{Current Status} & \textbf{Actionable Next Step} \\
\midrule
\endhead
\bottomrule
\endlastfoot
Standardized Lifecycle Benchmarks & \textbf{Seeded.} Continual ICRL gives direct benchmark and metric examples, including post-shift recovery and lifecycle reporting \citep{wang2025cicrl}; broader benchmark reporting norms remain fragmented. & Consolidate a shared suite with declared shift source, temporal pattern, observability, context reset policy, stale-context ablations, and oracle-context headroom \citep{moeini2025surveyicrl,kirk2023zeroshot,wang2025cicrl}. \\
Change-Point-Aware Context Policies & \textbf{Seeded but unresolved.} Changepoint methods and continual ICRL show useful templates \citep{adams2007changepoint,wang2025cicrl}, while transformer dynamic-regret work studies related adaptation objectives \citep{chen2025dynamicregret}. & Learn write/retrieve/forget rules from weakly labeled or unlabeled shifts, then evaluate detection delay, context repair, and post-shift return together. \\
Validity-Aware Retrieval & \textbf{Partly seeded.} Retrieval-augmented ICRL and retrieval-based RL exist \citep{goyal2022retrievalrl,schmied2024radt,sridhar2024regent}, but most retrieval keys still do not estimate decision validity after shifts. & Retrieve by latent mode, uncertainty, reward fit, recency, and time metadata; report precision of valid retrieved items, not only downstream return. \\
Adaptive Forgetting And Memory Isolation & \textbf{Partly seeded.} Continual learning and lifelong memory offer replay, regularization, and isolation mechanisms \citep{khetarpal2022continual,kirkpatrick2017overcoming,zhong2023memorybank}, but context-only agents need different guarantees. & Partition memory by inferred regime and test both irreversible shifts and recurring modes, where aggressive forgetting and indiscriminate retention fail in opposite ways. \\
Finite-Context Theory Under Stale Evidence & \textbf{Truly open.} ICL theory, non-stationary regret, and chain-of-thought ICRL theory provide components \citep{garg2022transformers,cheung2020nonstationary,xie2026cot}, but not a theory of bounded context containing invalid examples. & Bound regret or sample complexity as a function of context length, retrieval error, stale-context mass, shift observability, and model approximation error. \\
Efficient Long-Memory Architectures & \textbf{Partly seeded.} S4, Mamba, and long-context agents address cost and memory length \citep{lu2023s4icrl,gu2023mamba,elawady2024relic}, but rarely under matched invalidation tests. & Compare compressed-memory models under identical abrupt, gradual, and recurring shifts, with oracle-reset and stale-state ablations. \\
Action-Semantics Shift & \textbf{Partly seeded.} Variable-action ICRL and changing-action RL cover action-set variation \citep{chandak2020changingaction,sinii2024variable}, but within-lifetime interface drift remains under-tested. & Test permutations, masks, rescaling, action-description changes, and interface remapping while old context still contains obsolete action tokens. \\
Context Poisoning & \textbf{Truly open for non-stationary ICRL.} Safety and Bayesian value-fusion work are relevant \citep{moeini2025safeicrl,berkes2026spice}, but adversarial stale memories are not yet a standard threat model. & Stress-test stale demonstrations, malicious retrievals, corrupted rewards, and misleading summaries; measure whether the agent can isolate harmful context without online parameter updates. \\
Verbal Memory Validity In LLM Agents & \textbf{Partly seeded.} Reflection and experience-library agents store natural-language lessons \citep{shinn2023reflexion,zhao2024expel,wang2023voyager}, but validity under preference, tool, or reward drift is rarely measured. & Timestamp, regime-tag, contradict, retire, or reactivate memories using reward and feedback evidence rather than semantic similarity alone. \\
Constraint And Safety Shifts & \textbf{Seeded.} Safe ICRL and barrier-style shielding provide mechanisms \citep{altman2021cmdp,moeini2025safeicrl,kwon2026qbarrier}, but shifting constraints require separate reward-validity and safety-validity tests. & Add safety filters that work without parameter updates and report reward, cost, and violation recovery after constraint changes. \\
Weak-Data Non-Stationarity & \textbf{Partly seeded.} Offline RL and value-aware ICRL address weak or suboptimal data \citep{fu2020d4rl,kumar2020cql,berkes2026spice}, but changing data quality is usually not isolated from changing tasks. & Factor benchmarks by behavior-policy quality, reward shift, support mismatch, and uncertainty, so failures are not attributed only to non-stationarity. \\
Exploration Under Changing Feedback & \textbf{Partly seeded.} In-context exploration and LLM bandit work study diagnostic action choice \citep{krishnamurthy2024explore,monea2024bandit,nie2025evolve}, but feedback drift adds a second inference problem. & Combine in-context exploration with shift-aware uncertainty summaries and report how quickly the agent collects disambiguating evidence after feedback changes. \\
\end{longtable}
\endgroup

Safety is a related but distinct pressure on this agenda.
Safe ICRL extends the adaptation problem to constrained decision processes \citep{altman2021cmdp,achiam2017cpo,satija2020cmdp}, where context must help the agent improve reward while respecting cost budgets \citep{moeini2025safeicrl}.
Recent shielding work makes the same point more explicit under OOD deployment shifts: Q-Barrier \citep{kwon2026qbarrier} suggests that a frozen safe-ICRL policy may need an action-level latent budget and cost filter, not only a longer history, to preserve a useful reward-safety tradeoff.
The non-stationary version is harder because a context item may be reward-valid but safety-invalid after a constraint or cost shift.
This reinforces the need to report validity with respect to rewards, dynamics, and constraints separately.
Classical safe exploration \citep{moldovan2012safe}, model-based safety guarantees \citep{berkenkamp2017safe}, safe policy improvement \citep{laroche2019safe}, and modern safe-RL benchmarks \citep{ji2023safetygymnasium} provide useful components, but they still need to be recast under the fixed-parameter, context-only adaptation constraint.

\subsection{A Research Agenda}

The next phase of the field should move beyond the broad question of whether context helps.
For non-stationary ICRL, the sharper questions are which context helps, when it helps, when it hurts, and how a frozen agent can know the difference.
The roadmap should proceed in stages.
The first stage is consolidation: seeded directions such as continual ICRL benchmarks \citep{wang2025cicrl}, safe ICRL \citep{moeini2025safeicrl,kwon2026qbarrier}, and retrieval-augmented ICRL \citep{schmied2024radt,sridhar2024regent} should be evaluated under a common lifecycle protocol rather than treated as isolated demonstrations.
The second stage is translation: adjacent tools from changepoint detection, non-stationary regret, continual learning, retrieval, and safe RL should be rewritten in context-channel terms, specifying what is stored, retrieved, suppressed, or exposed to the frozen model.
The third stage is genuinely open theory and stress testing: finite-context bounds, adversarial stale-context threat models, and validity-aware memory metrics should become standard targets.

Claims about non-stationary adaptation should specify the shift source, temporal pattern, observability regime, context policy, and training exposure.
They should report post-shift curves and stale-context ablations, compare against simple context policies before attributing gains to architecture scale, and state whether a result is solving a new open problem or extending an already seeded line.

The agenda also requires a cleaner separation between the pretrained decision model, the context manager, and the memory store.
Current papers often blur these components together.
In stationary settings, that may be tolerable because most useful context points in the same direction.
In non-stationary settings, the distinction is central.
A strong policy with a weak context manager can fail, while a modest policy with a good validity filter can recover quickly.

\subsection{Reporting Standards for a Mature Literature}

A mature non-stationary ICRL literature would make claims at the right level of specificity.
Instead of saying that a model handles non-stationarity, a paper would say that it handles reward/abrupt/partial shifts with a finite context budget, or dynamics/recurring/latent shifts with an external memory.
Instead of reporting only average return, it would show lifecycle curves, recovery time, stale-context sensitivity, and oracle-context headroom.
Instead of treating retrieval as a black-box improvement, it would report whether retrieved items are valid under the current regime.
Instead of presenting long context as inherently beneficial, it would show when additional context helps, when it hurts, and what mechanism controls its influence.

This reporting standard is demanding, but it is aligned with the central promise of ICRL.
If adaptation is supposed to happen without weight updates, then the evidence passed to the frozen model is the adaptation interface.
The field should therefore scrutinize that interface with the same care that reinforcement-learning papers apply to rewards, dynamics, and exploration.

\clearpage
\section{Conclusion}
\label{sec:conclusion}

This survey has argued that non-stationary \ICRL{} is not just ICRL evaluated on a larger or more diverse task distribution.
It is the problem of adapting through context while deployed policy parameters remain fixed, even when accumulated evidence becomes obsolete, misleading, or useful again after a regime returns.
The key difficulty is context validity: a fixed-parameter policy must infer which accumulated evidence still describes the current decision process and which evidence should be downweighted, retired, or reactivated.
This view connects black-box meta-RL \citep{duan2016rl2,wang2016learning}, decision sequence modeling \citep{chen2021decision,lee2023supervised}, algorithm distillation \citep{laskin2023algorithm}, retrieval-augmented agents \citep{schmied2024radt,sridhar2024regent}, and long-context adaptive systems \citep{elawady2024relic}, but it shifts the organizing question from ``does context help?'' to ``when should context be trusted?''.

From this perspective, the survey made five contributions to the non-stationary ICRL problem.
\textbf{First, }it defined the setting and separated it from adjacent notions such as stationary few-shot adaptation, domain randomization, online continual learning, standard offline RL, robotics-specific lifelong learning, and multi-agent strategic non-stationarity.
The defining constraint is that adaptation occurs through context while the deployed policy parameters remain fixed.
\textbf{Second, }it introduced a taxonomy that describes a non-stationary ICRL setting along three axes: what changes, how the change unfolds over time, and how observable the change is to the agent.
This source--pattern--observability vocabulary is intended to prevent overbroad claims: reward specifications, transition kernels, action interfaces, observation channels, constraint models, and data sources create different problems, and abrupt, gradual, recurring, expanding, or adversarial changes require different evidence-management behavior.
\textbf{Third, }the survey recast methods as context-management mechanisms rather than only as model families.
Raw history conditioning, context filtering, retrieval, compression, recurrent memory, value-aware inference, model-based prediction, reflection, and forgetting are all ways of deciding which past evidence reaches the frozen decision policy and how strongly it should influence action.
Under non-stationarity, these mechanisms are not engineering details; they are the adaptation channel.
\textbf{Fourth, }the survey translated that view into training and evaluation requirements.
Training data should contain invalidation, mixed pre-shift and post-shift histories, stale demonstrations, suboptimal or mismatched trajectories, recurring modes, and finite-context pressure \citep{chen2025filtering,wang2026suboptimalicrl}.
Evaluation should then report lifecycle curves, post-shift recovery, dynamic regret, stale-context sensitivity, retrieval utility, oracle-context gaps, recurring-mode reuse, and finite-context degradation rather than only average return.
Benchmarks such as procedural environments, continual RL suites, and memory-focused tasks become evidence for non-stationary ICRL only when they impose within-lifetime shifts and require recovery through context.
\textbf{Fifth, }the survey identified the theoretical gap created by stale and conflicting context.
Existing change-point methods \citep{adams2007changepoint}, non-stationary bandit regret \citep{besbes2014nonstationarybandit}, non-stationary MDP regret \citep{cheung2020nonstationary,fei2020dynamic}, and continual ICRL \citep{wang2025cicrl} provide useful pieces, but they do not yet explain how regret, sample efficiency, or robustness should scale with context length, retrieval error, memory size, shift observability, and context poisoning.
A mature theory of non-stationary ICRL should treat the context manager as part of the control system, because the agent acts through a bounded and imperfect evidence channel.

The practical conclusion is that future work should make its claim at the level of the shift it actually solves.
A method that improves return under clean histories is not yet a non-stationary ICRL solution if it fails under stale context.
A method that recovers after abrupt reward changes has not necessarily solved gradual dynamics drift or hidden action remapping.
A method that forgets quickly may adapt after one-way change but fail when old regimes recur.
Progress in non-stationary ICRL will come from systems that can detect change, select currently valid evidence, preserve reusable memories, reject obsolete context, and act well under a finite context budget.
The central benchmark for the field is therefore not larger context alone, but disciplined evidence use across the lifetime of a changing decision process.

\clearpage

\FloatBarrier
\bibliography{main}
\bibliographystyle{tmlr}

\end{document}